\documentclass{article}

\PassOptionsToPackage{numbers}{natbib}
\usepackage[preprint]{neurips_2023}

\usepackage[utf8]{inputenc} 
\usepackage[T1]{fontenc}    
\usepackage{hyperref}       
\usepackage{url}            
\usepackage{booktabs}       
\usepackage{amsfonts}       
\usepackage{nicefrac}       
\usepackage{microtype}      
\usepackage[dvipsnames]{xcolor}         
\usepackage{times}
\usepackage{latexsym}
\usepackage{CJKutf8}
\usepackage{amsmath}
\usepackage{verbatim}
\usepackage{booktabs}
\usepackage{bbm}
\usepackage{amssymb}
\usepackage{pifont}
\usepackage{stfloats}
\usepackage{multicol}
\usepackage{float}
\usepackage{graphicx}
\usepackage{subfigure}
\usepackage{multirow}
\usepackage{array}
\usepackage{enumerate}
\newcommand{\PreserveBackslash}[1]{\let\temp=\\#1\let\\=\temp}
\newcolumntype{C}[1]{>{\PreserveBackslash\centering}p{#1}}
\newcolumntype{R}[1]{>{\PreserveBackslash\raggedleft}p{#1}}
\newcolumntype{L}[1]{>{\PreserveBackslash\raggedright}p{#1}}
\usepackage{diagbox}
\usepackage{colortbl}
\usepackage{enumitem}
\usepackage{bm}
\usepackage{mathtools}
\usepackage{wrapfig}
\usepackage[skins]{tcolorbox}
\tcbuselibrary{breakable} 
\newtcolorbox[auto counter, number within=section, list type=subsubsection, list inside=toc]{sectionbox}[1]{
colback=white, colframe=black, 
colbacktitle=white!80!gray, coltitle=black, 
fonttitle=\bfseries, title={Comment \thetcbcounter}, list entry={Comment \thetcbcounter\quad}, 
breakable, 
before upper={\parindent10pt\noindent},  
left = 1mm, 
    right = 1mm,
    top = 1mm,
    bottom = 1mm,
}

\usepackage{CJKutf8}
\usepackage{tabularx}
\usepackage{amsmath}
\usepackage{ragged2e}
\usepackage{longtable}

\newcommand\icon{\raisebox{-3.7pt}{\includegraphics[width=1.1em]{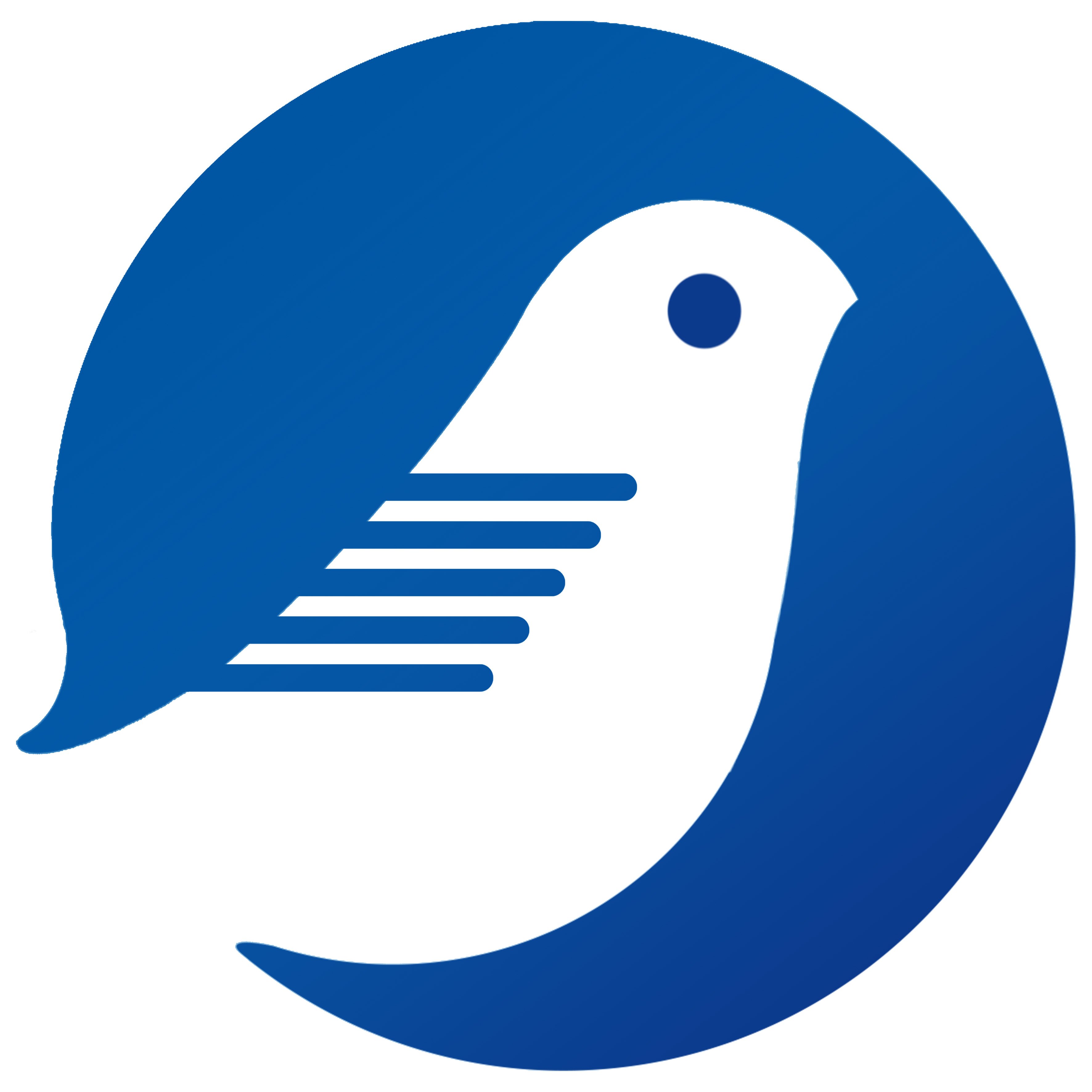}}}

\title{\icon~BayLing: Bridging Cross-lingual Alignment and Instruction Following through Interactive Translation for Large Language Models}

\author{
Shaolei Zhang\textsuperscript{\rm 1},\quad
Qingkai Fang\textsuperscript{\rm 1},\quad
Zhuocheng Zhang\textsuperscript{\rm 1},\quad
Zhengrui Ma\textsuperscript{\rm 1},\\
\textbf{Yan Zhou\textsuperscript{\rm 1}, \quad
Langlin Huang\textsuperscript{\rm 1},\quad
Mengyu Bu\textsuperscript{\rm 1},\quad
Shangtong Gui\textsuperscript{\rm 2},}\\
\textbf{Yunji Chen\textsuperscript{\rm 2},\quad
Xilin Chen\textsuperscript{\rm 1},\quad
Yang Feng}\textsuperscript{\rm 1}\thanks{Corresponding author: Yang Feng.} \\
\textsuperscript{\rm 1}{Key Laboratory of Intelligent Information Processing} \\ $\;\:$Institute of Computing Technology, Chinese Academy of Sciences (ICT/CAS)\\
\textsuperscript{\rm 2}{State Key Lab of Computer Architecture} \\ $\;\:$Institute of Computing Technology, Chinese Academy of Sciences (ICT/CAS)\\
$\;\:$\texttt{\{\href{mailto:zhangshaolei20z@ict.ac.cn}{zhangshaolei20z},\href{mailto:fengyang@ict.ac.cn}{fengyang}\}@ict.ac.cn}}

\begin{document}

\maketitle

\begin{abstract}
Large language models (LLMs) have demonstrated remarkable prowess in language understanding and generation. Advancing from foundation LLMs to instruction-following LLMs, instruction tuning plays a vital role in aligning LLMs to human preferences. However, the existing LLMs are usually focused on English, leading to inferior performance in non-English languages. In order to improve the performance for non-English languages, it is necessary to collect language-specific training data for foundation LLMs and construct language-specific instructions for instruction tuning, both of which are heavy loads. To minimize human workload, we propose to transfer the capabilities of language generation and instruction following from English to other languages through an interactive translation task. We have developed BayLing, an instruction-following LLM by utilizing LLaMA as the foundation LLM and automatically constructing interactive translation instructions for instructing tuning. Extensive assessments demonstrate that BayLing achieves comparable performance to GPT-3.5-turbo, despite utilizing a considerably smaller parameter size of only 13 billion. Experimental results on translation tasks show that BayLing achieves 95\% of single-turn translation capability compared to GPT-4 with automatic evaluation and 96\% of interactive translation capability compared to GPT-3.5-turbo with human evaluation. To estimate the performance on general tasks, we created a multi-turn instruction test set called BayLing-80. The experimental results on BayLing-80 indicate that BayLing achieves 89\% of performance compared to GPT-3.5-turbo. BayLing also demonstrates outstanding performance on knowledge assessment of Chinese GaoKao and English SAT, second only to GPT-3.5-turbo among a multitude of instruction-following LLMs. Demo\footnote{Demo: \href{http://nlp.ict.ac.cn/bayling/demo}{http://nlp.ict.ac.cn/bayling/demo}\qquad \textsuperscript{\rm 3}Homepage: \href{http://nlp.ict.ac.cn/bayling}{http://nlp.ict.ac.cn/bayling}}, homepage\footnotesize\footnotemark, code\footnote{Code: \href{https://github.com/ictnlp/BayLing}{https://github.com/ictnlp/BayLing}\qquad $\!\!$ \textsuperscript{\rm 5}Models: \href{https://huggingface.co/ICTNLP/bayling-7b-diff}{BayLing-7B}, \href{https://huggingface.co/ICTNLP/bayling-13b-diff}{BayLing-13B}} and models\footnotesize\footnotemark of BayLing are available.
\end{abstract}

\section{Introduction}
Large language models (LLMs), such as GPT-3 \cite{brown2020language}, PaLM \cite{chowdhery2022palm}, OPT \cite{zhang2022opt}, GLM \cite{du-etal-2022-glm}, BLOOM \cite{workshop2023bloom} and LLaMA \cite{touvron2023llama}, possess powerful language generation capabilities learning to predict the next token based on extensive training on a large corpus, even handle a large range of natural language processing (NLP) tasks with only a description or some instances of the task as the prompt. However, the training objective of generating the most likely words may not align well with human intent, which makes LLMs often generate unexpected responses. Hence,  instruction tuning \cite{NIPS2017_d5e2c0ad,NEURIPS2022_b1efde53} ranging across various NLP tasks is introduced to further train the foundation LLMs to behave in accordance with human intent \cite{ouyang2022training,mishra-etal-2022-cross}. In this way, we can get instruction-following LLMs, such as ChatGPT \cite{chatgpt}, GPT-4 \cite{openai2023gpt4} and Bard. 

In practice, foundation LLMs are often trained on corpora dominated by the English language \cite{brown2020language,touvron2023llama} as it is more challenging to collect equivalently sized corpora of languages other than English. Additionally, instructions are commonly constructed in English \cite{alpaca}, further emphasizing the English-centric nature of LLMs. Both lead to the inferior performance of LLMs on non-English languages. To improve the performance of LLMs in non-English languages, a larger corpus of training data specific to those languages needs to be collected to enhance generation capabilities, and meanwhile additional instructions in non-English languages across various NLP tasks should be constructed to strengthen the ability to follow instructions, both of which are difficult to fulfill.

Under these grounds, we explore to minimize human workload by aligning other languages to English and thereby transferring the capabilities of language generation and instruction following from English to those languages. In this scenario, we leverage the interactive translation task to enhance the two capabilities simultaneously, without introducing additional training data in non-English languages during the pre-training of the foundation LLM. 
It has been proven that compared to acquiring language generation capabilities for foundation LLMs, achieving semantic alignments among languages requires significantly less training data \cite{garcia2023unreasonable}, which implies that only a small number of interactive translation instructions need to be constructed for language alignment.
Furthermore, interactive translation typically employs multi-turn interaction to adjust translations based on human preferences and thus following the instructions of interactive translation entails aligning with human preferences through the process. Therefore, only a restricted number of interactive translation instructions involving English as part of the translation  are required to successfully transfer the abilities of language generation and instruction following from English to other languages participating in interactive translation.
Fortunately, translation is a well-defined task with clearly defined output, requiring the generated translations to convey the same meaning as the input, so finally with the help of interactive translation, we only need to generate interactive translation instructions automatically without the need of manual construction.

We have developed BayLing, an instruction-following LLM in the above method with LLaMA as its foundation. In order to assess the capabilities of BayLing, we conducted extensive experiments on translation tasks, general tasks and standardized tests. The results demonstrate that BayLing achieves comparable performance to GPT-3.5-turbo, despite utilizing a considerably smaller parameter size of only 13 billion. For single-turn translation tasks, the automatic evaluation results using standard translation benchmarks demonstrate that BayLing achieves a translation capability of 95\% compared to GPT-4 on Chinese$\Leftrightarrow$English and English$\Leftrightarrow$German translations. Furthermore, for multi-turn interactive translation tasks, human evaluation reveals that BayLing achieves a 96\% interactive capability of GPT-3.5-turbo. To evaluate BayLing's performance on general tasks, we extended the Vicuna-80 test set \cite{vicuna2023} to include multi-turn interactions, creating a multi-turn instruction test set called BayLing-80. Evaluation with GPT-4 on BayLing-80 demonstrates that BayLing achieves a performance level of 89\% compared to GPT-3.5-turbo. In knowledge assessment, BayLing showcases good performance on standardized tests such as Chinese GaoKao and English SAT, ranking second only to GPT-3.5-turbo. By further analyzing the experimental results, we get the following findings:
\begin{itemize}[,itemsep=0pt,topsep=0pt,leftmargin=12pt]
\setlength{\itemsep}{0pt}
\setlength{\parsep}{0pt}
\setlength{\parskip}{0pt}
    \item Introducing interactive translation instructions can enhance the generation capabilities of non-English languages involved in the interactive translation task, without negatively impacting the generation capabilities of the English language.
    \item Introducing interactive translation instructions can assist LLMs in comprehending human intent and accurately following user instructions.
    \item Introducing interactive translation instructions can refine the translation for both single-turn and multi-turn interactive translation across the languages included in the interactive translation.
    \item Introducing interactive translation instructions 
    can improve the performance on general tasks for the languages incorporated in interactive translation tasks.
    \item BayLing demonstrates proficiency in knowledge-related tasks, writing tasks and generic tasks, but requires further improvement in math, coding and reasoning-related tasks.
\end{itemize}

\section{Related Work}

Recent advances in natural language processing (NLP) have led to the development of powerful large language models such as GPT-3 \cite{brown2020language}, PaLM \cite{chowdhery2022palm}, OPT \cite{zhang2022opt}, GLM \cite{du-etal-2022-glm}, BLOOM \cite{workshop2023bloom} and LLaMA \cite{touvron2023llama}. Pre-training on a large corpus endows these foundation LLMs with extraordinary language understanding and generation capabilities. On this basis, the most important step for these foundation LLMs to evolve into powerful instruction-following LLMs such as ChatGPT \cite{chatgpt} and GPT-4 \cite{openai2023gpt4} is to understand human instructions and align with human preferences.

ChatGPT \cite{chatgpt} is a milestone in the field of LLMs, which endows LLMs with remarkable capabilities to understand human instructions and align with human preferences through the application of instruction tuning \cite{ouyang2022training} and reinforcement learning from human feedback (RLHF) \cite{NIPS2017_d5e2c0ad,NEURIPS2022_b1efde53}. Nevertheless, due to the substantial labor costs associated with the RLHF stage, researchers have made efforts to align LLMs with human preferences primarily through high-quality instruction tuning \cite{sun2023principledriven,köpf2023openassistant,zhou2023lima}, resulting in noteworthy advancements in this area.

In this study, we focus on how to transfer the language generation and instruction following capabilities of LLMs learned from English-dominated corpora to other languages. In particular, we aim to find an efficient way to simultaneously transfer these capabilities through instruction tuning. To this end, we develop BayLing to use multi-turn interactive translation tasks to simultaneously improve the language alignment, instruction following and multi-turn interactive capabilities of LLMs. The emergence of BayLing fills the gaps in the previous research on instruction tuning in cross-lingual alignment and multi-turn interaction.

\section{Construction of BayLing}
\begin{figure}
    \centering
    \includegraphics[width=0.9\textwidth]{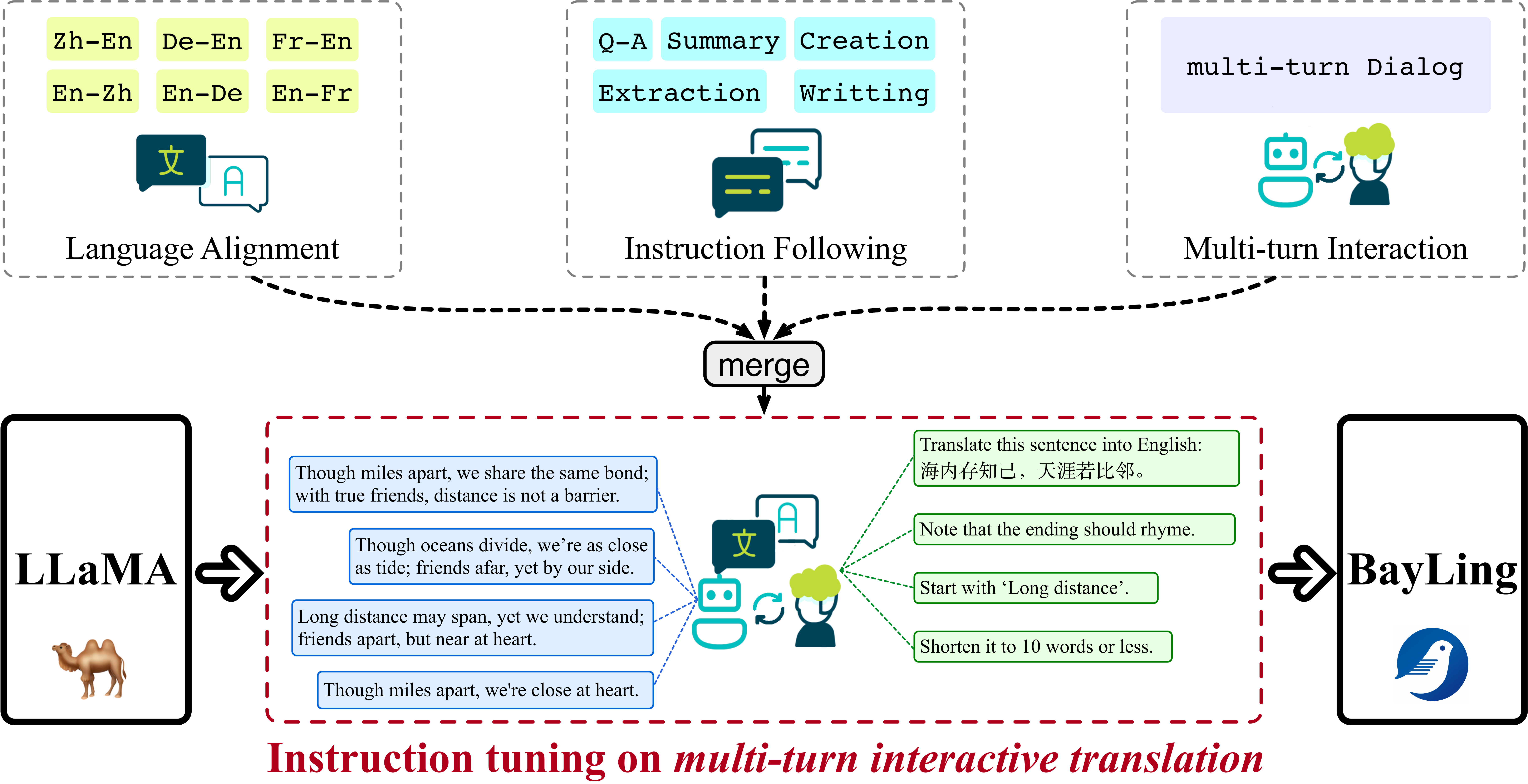}
    \caption{Overview of BayLing. BayLing is built upon LLaMA and fine-tuned with instruction data of interactive translation task and general tasks.}
    \label{fig:model}
\end{figure}

We develop BayLing, an instruction-following LLM, to enhance the language generation capabilities of LLM in non-English languages and meanwhile align LLM with human intent. We choose the interactive translation task to achieve the two goals simultaneously, which can transfer language generation capabilities from English to those languages through cross-lingual alignment and as well follow human feedback through multi-turn interaction.
As depicted in Figure~\ref{fig:model}, we build BayLing upon the LLaMA model \cite{touvron2023llama} and fine-tune it with interactive translation instructions and general instructions. In the following sections, we will give a detailed description of BayLing.

\subsection{Foundation Model}

Recently, LLaMA has demonstrated outstanding performance on a wide range of tasks in English, which is trained on approximately 1.4 trillion English-dominant tokens. Given their remarkable language understanding and generation capabilities, LLaMA-7B and LLaMA-13B are used as the foundation models of BayLing-7B and BayLing-13B, respectively.

\subsection{Alignment with Interactive Translation Task}
\label{sec:Instruction}

In interactive translation, users communicate with the LLM in multi-turn to polish the translation until satisfaction. An illustration of interactive translation is provided in Figure~\ref{fig:imt_case}. To fulfill the requirements of the users, the LLM needs to first comprehend the user's intention and then generate the translation in accordance with the intention. 

\begin{figure}[h]
\small
\begin{tcolorbox}[colback=SeaGreen!10!CornflowerBlue!10,colframe=RoyalPurple!55!Aquamarine!100!]

Below is a dialog consisting of instructions and responses. Write a response that completes the request.

\textbf{\#\#\# Instruction:}

Translate this sentence into English:\begin{CJK*}{UTF8}{gbsn} 海内存知己，天涯若比邻。\end{CJK*}

\textbf{\#\#\# Response:}

Though miles apart, we share the same bond; with true friends, distance is not a barrier.

\textbf{\#\#\# Instruction:}

Note that the ending should rhyme.

\textbf{\#\#\# Response:}

Though oceans divide, we're as close as tide; friends afar, yet by our side.

\textbf{\#\#\# Instruction:}

Start with `Long distance'.

\textbf{\#\#\# Response:}

Long distance may span, yet we understand; friends apart, but near at heart.

\textbf{\#\#\# Instruction:}

Shorten it to 10 words or less.

\textbf{\#\#\# Response:}

Though miles apart, we're close at heart.

\centering$\cdots$
\end{tcolorbox}
\caption{An illustration of interactive translation task}
\label{fig:imt_case}
\end{figure}

Due to the inherent challenges and intricacies involved, the effective execution of interactive translation requires LLMs to exhibit a high level of proficiency in language alignment, instruction following and contextual comprehension.
Furthermore, since the multi-turn interaction task is always human-centered and encourages the output of LLMs to meet the user's requirements and constraints, the alignment between LLMs and human preferences is naturally enhanced.
Therefore, the task of interactive translation presents an ideal opportunity for simultaneously enhancing the language alignment and instruction following capabilities of LLMs. Moreover, LLMs can leverage their strong generalization capability to transfer these proficiencies learned from interactive translation to other tasks.

To this end, we constructed an instruction dataset of 160K interactive translation instances, encompassing various types of interactions, such as vocabulary, grammar, style, creation, etc. In order to enhance the diversity of the instruction data for BayLing, we supplemented the instruction data with relevant public datasets, including \texttt{Alpaca} for single-turn instruction data and \texttt{ShareGPT} for multi-turn instruction data. Table~\ref{tab:instruction} provides the statistics of instruction data employed in BayLing.

\begin{table}[t]
\centering
\caption{Statistics of instruction data for BayLing}\label{tab:instruction}
\begin{tabular}{lcccc}\toprule
\textbf{Source}                                   & \textbf{Interactive}                            & \multicolumn{2}{c}{\textbf{Languages}}                                                       & \textbf{\#Instances} \\ \midrule
\textbf{Alpaca}                                   & Single-turn                                     & \multicolumn{2}{c}{English}                                                                  & 52K                  \\
\textbf{ShareGPT}                                 &Multi-turn                                      & \multicolumn{2}{c}{English-dominant}                                                         & 90K                  \\\midrule
\multirow{4}{*}{\begin{tabular}[l]{@{}l@{}}\textbf{Interactive}\\ \textbf{Translation}\end{tabular}} & \multicolumn{1}{c}{\multirow{4}{*}{Multi-turn}} & Instruction Languages     & Translation   Languages                                                            & \multicolumn{1}{c}{\multirow{4}{*}{160K}} \\ \cmidrule(lr){3-4}
                                         &                             & English, Chinese & \begin{tabular}[c]{@{}c@{}}English, Chinese\\ German, French\end{tabular} & \multicolumn{1}{l}{}                       \\\bottomrule
\end{tabular}
\end{table}

\subsection{Training Details}
By utilizing LLaMA-7B and LLaMA-13B as the foundational models, we fine-tuned BayLing-7B and BayLing-13B on the instruction data proposed in Section~\ref{sec:Instruction}. Our training configuration involved the learning rate of 2e-5 and weight decay of 0.0. Furthermore, we fine-tune BayLing for 3 epochs, using a batch size of 32. We limit the maximum text length to 1024, and only compute the loss on the output parts in the multi-turn interaction. During training, we employed DeepSpeed \cite{10.1145/3394486.3406703} and Gradient Checkpointing \cite{chen2016training} techniques to optimize memory consumption. During inference, BayLing-13B can be deployed on a GPU with 16G memory through 8-bit quantization.

\section{Experiments}

To verify whether multi-turn interactive translation instructions can enhance the language alignment, instruction following and multi-turn interactive capabilities of LLMs, ultimately aligning LLMs with human preferences, we evaluate the performance of BayLing on extensive tasks including multilingual translation, interactive translation, general tasks and standardized tests.

\subsection{Multilingual Translation}

We verify BayLing's capability of language alignment on the standard translation benchmarks. 

\subsubsection{Experimental Setup}

\textbf{Benchmarks}\quad We evaluate the multilingual translation \cite{johnson-etal-2017-googles} capability of BayLing on WMT22 benchmarks\footnote{\url{https://www.statmt.org/wmt22/translation-task.html}}, which encompass four translation directions: Chinese$\Leftrightarrow$English and German$\Leftrightarrow$English. Additionally, we explore BayLing's effectiveness in zero-shot translation by incorporating some other languages, such as Czech$\Leftrightarrow$English, Japanese$\Leftrightarrow$English, Russian$\Leftrightarrow$English, Ukrainian$\Leftrightarrow$English and German$\Leftrightarrow$French. All these language pairs were not involved in the instruction tuning stage, allowing us to evaluate BayLing's capability of translating between unseen language pairs.

\textbf{Baselines}\quad We compare BayLing-7B and BayLing-13B with state-of-the-art translation models, including both translation-specific large models and general instruction-following LLMs. For translation-specific large models, we employ \textbf{Google Translate}\footnote{\url{https://translate.google.com/}}, \textbf{NLLB-3.3B}\footnote{\url{https://huggingface.co/facebook/nllb-200-3.3B}} \cite{nllbteam2022language}, where Google Translate is the most widely used translation service and NLLB-3.3B is a multilingual translation model trained on a vast parallel corpus of over 200 languages. For general instruction-following LLMs, we employ \textbf{GPT-4}\footnote{We use GPT-4 API of version 0314} \cite{openai2023gpt4}, \textbf{GPT-3.5-turbo}\footnote{We use GPT-3.5-turbo API} \cite{chatgpt}, \textbf{ChatGLM-6B}\footnote{\url{https://huggingface.co/THUDM/chatglm-6b}}, \textbf{BLOOMZ-7B1-MT}\footnote{\url{https://huggingface.co/bigscience/bloomz-7b1-mt}} \cite{workshop2023bloom}, \textbf{Vicuna-13B}\footnote{\url{https://huggingface.co/lmsys/vicuna-13b-delta-v1.1}} \cite{vicuna2023}, \textbf{ParroT-7B}\footnote{\url{https://huggingface.co/wxjiao/ParroT-7b}} \cite{jiao2023parrot} and \textbf{Alpaca-7B}\footnote{\url{https://huggingface.co/tatsu-lab/alpaca-7b-wdiff}} \cite{alpaca}. Currently, GPT-4 and GPT-3.5-turbo are considered the most powerful LLM services. ChatGLM-6B and BLOOMZ-7B1-MT are instruction-tuned versions based on the bilingual foundation LLMs, GLM \cite{du-etal-2022-glm} and BLOOM \cite{workshop2023bloom}, respectively. Vicuna-13B, ParroT-7B and Alpaca-7B are all built upon the LLaMA \cite{touvron2023llama} and further fine-tuned using shareGPT, translation parallel corpus, and 52K instruction data, respectively. We provide the translation quality of all these LLMs on the WMT 22 benchmarks in Appendix~\ref{app:benchmark} and release their translation results as open source, which can serve as baselines in future translation research.

\textbf{Evaluation}\quad We apply two widely-used metrics, \texttt{SacreBLEU} \cite{post-2018-call} and \texttt{COMET} \cite{rei-etal-2022-comet}, to assess the quality of systems' translation. SacreBLEU score measures the statistical similarity based on n-gram accuracy. COMET score measures the semantic similarity using cross-lingual pre-trained models, which is currently regarded as the most human-aligned evaluation metric for translation tasks.

\subsubsection{Single-turn Translation}

\begin{figure}[t]
\centering
\subfigure[COMET scores on Chinese-to-English translation]{
\includegraphics[width=0.48\textwidth]{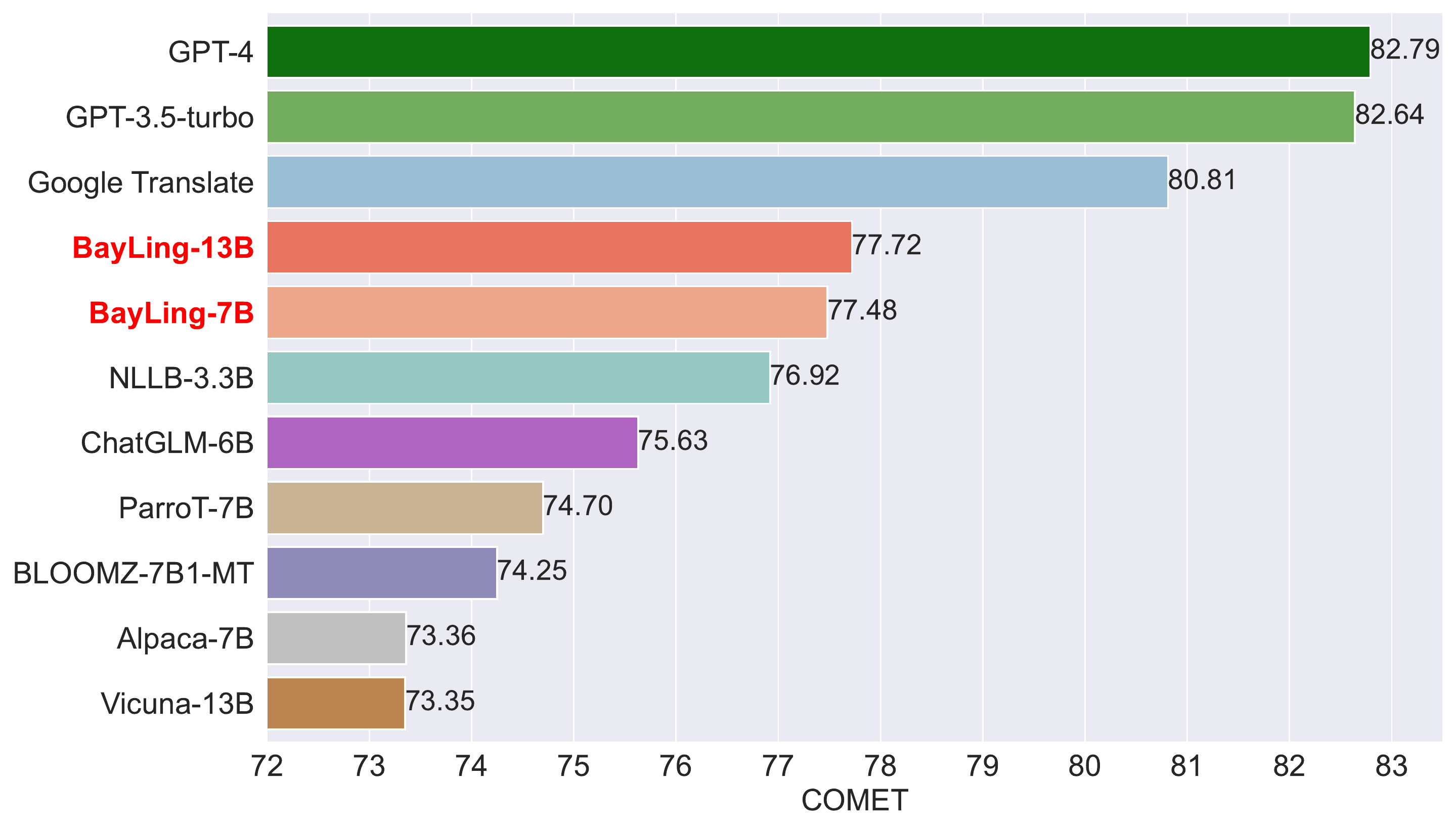} \label{fig:general_mt_zhen1}
}
\subfigure[COMET scores on English-to-Chinese translation]{
\includegraphics[width=0.48\textwidth]{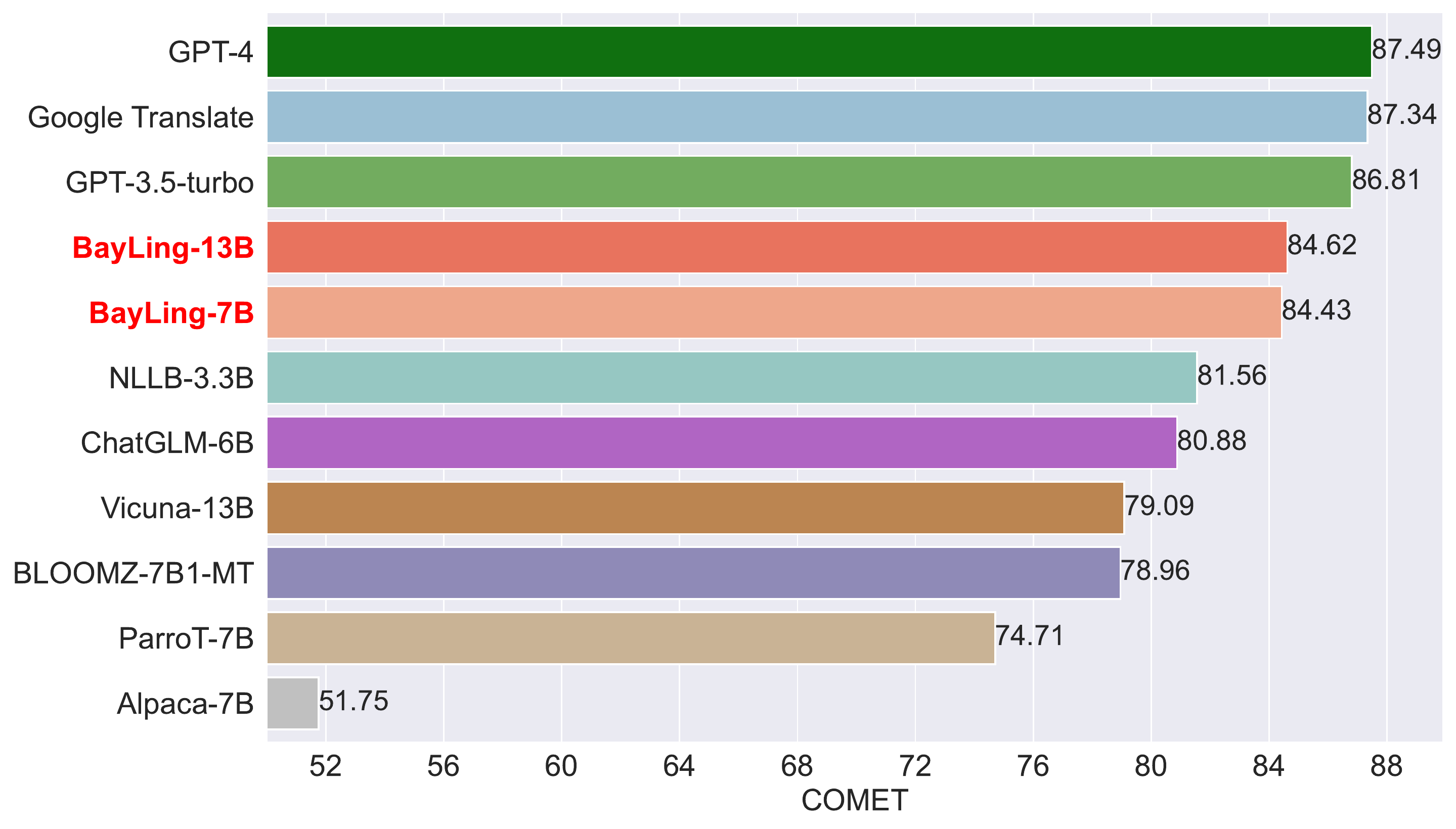} \label{fig:general_mt_zhen2}
}
\subfigure[BLEU scores on Chinese-to-English translation]{
\includegraphics[width=0.48\textwidth]{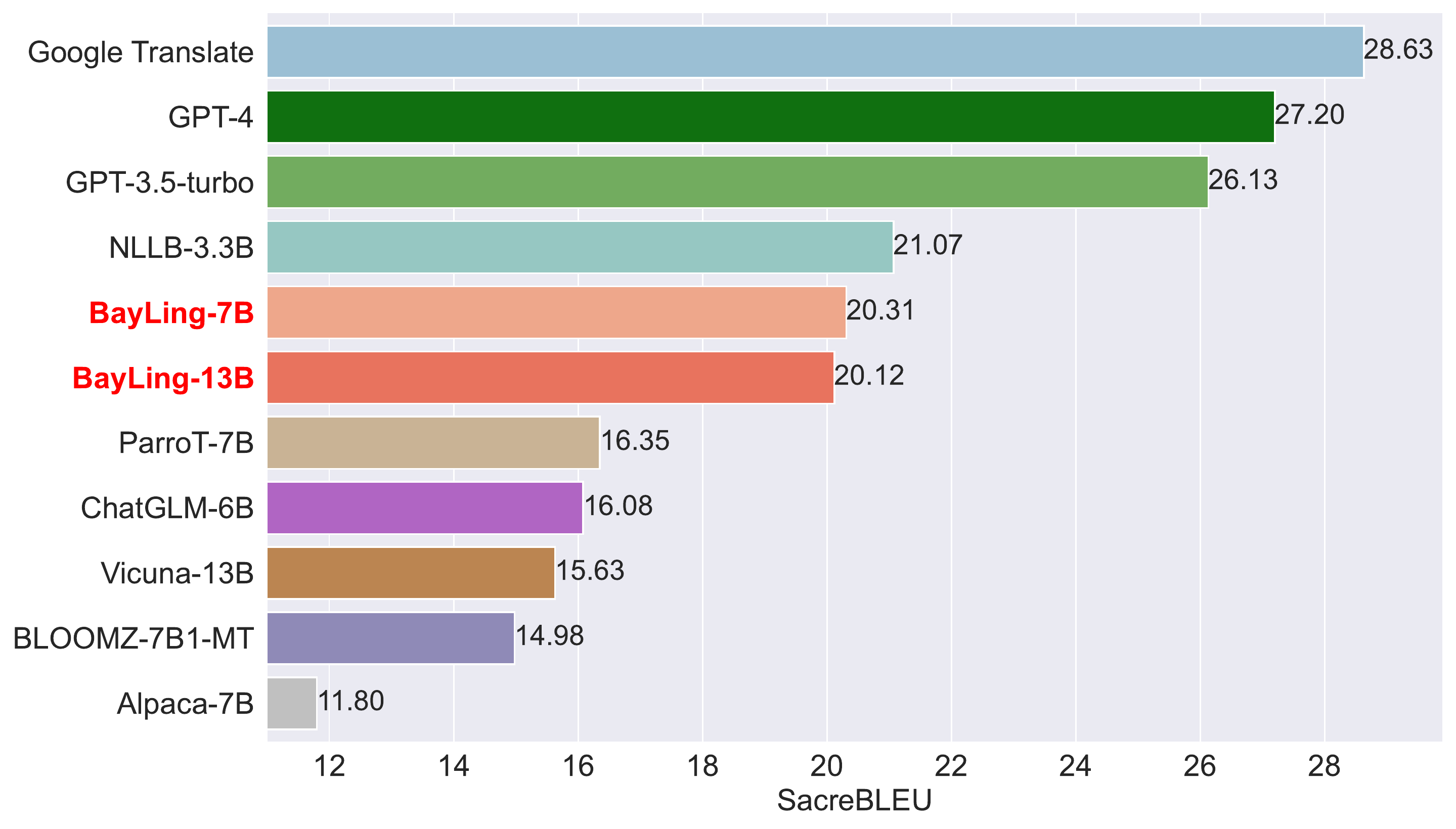} \label{fig:general_mt_bleu_zhen1}
}
\subfigure[BLEU scores on English-to-Chinese translation]{
\includegraphics[width=0.48\textwidth]{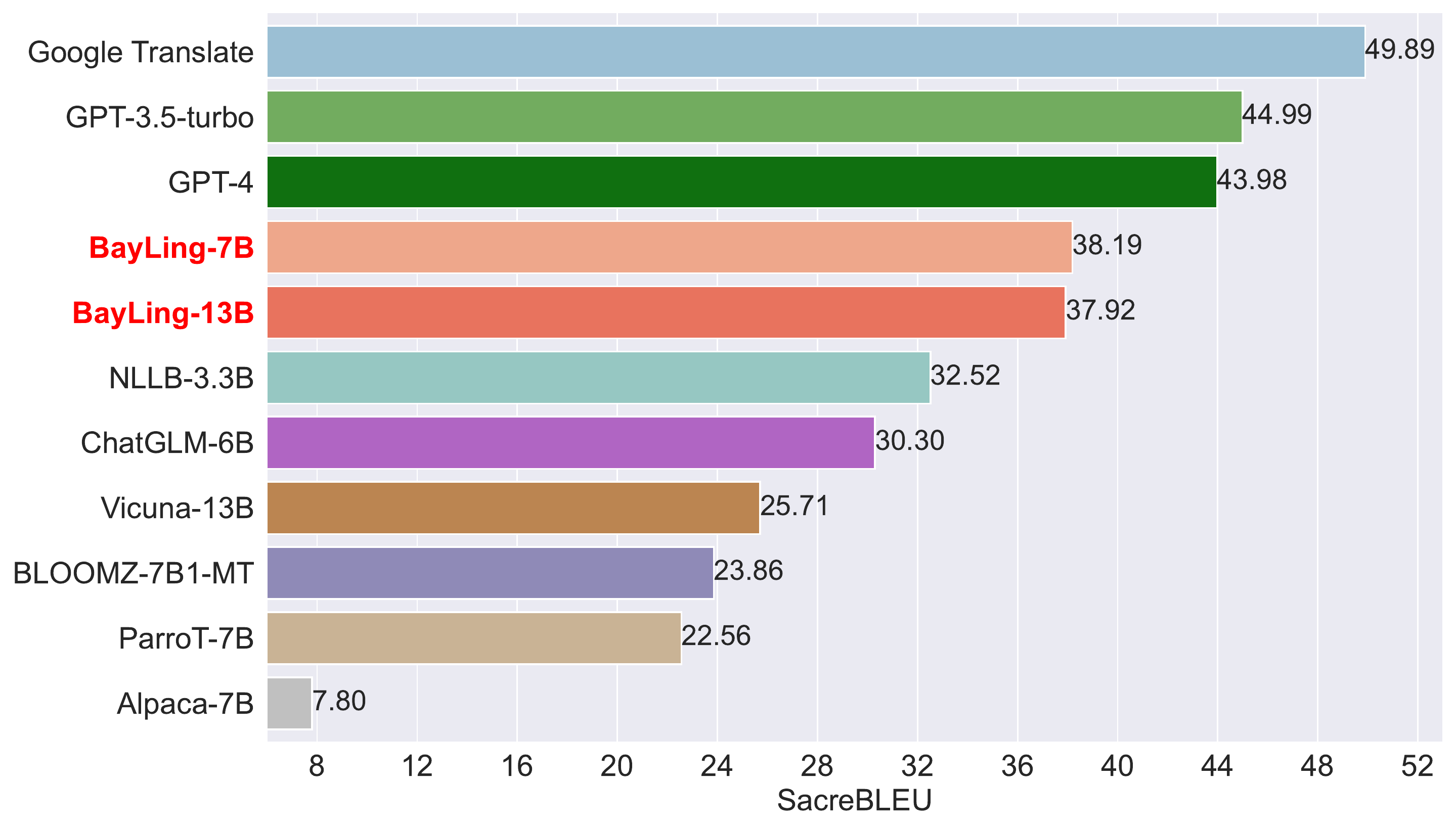} \label{fig:general_mt_bleu_zhen2}
}
\caption{Performance comparison on WMT22 Chinese$\Leftrightarrow$English translation task}
\label{fig:general_mt}
\end{figure}

\begin{figure}[t]
\centering
\subfigure[COMET score on German-to-English translation]{
\includegraphics[width=0.48\textwidth]{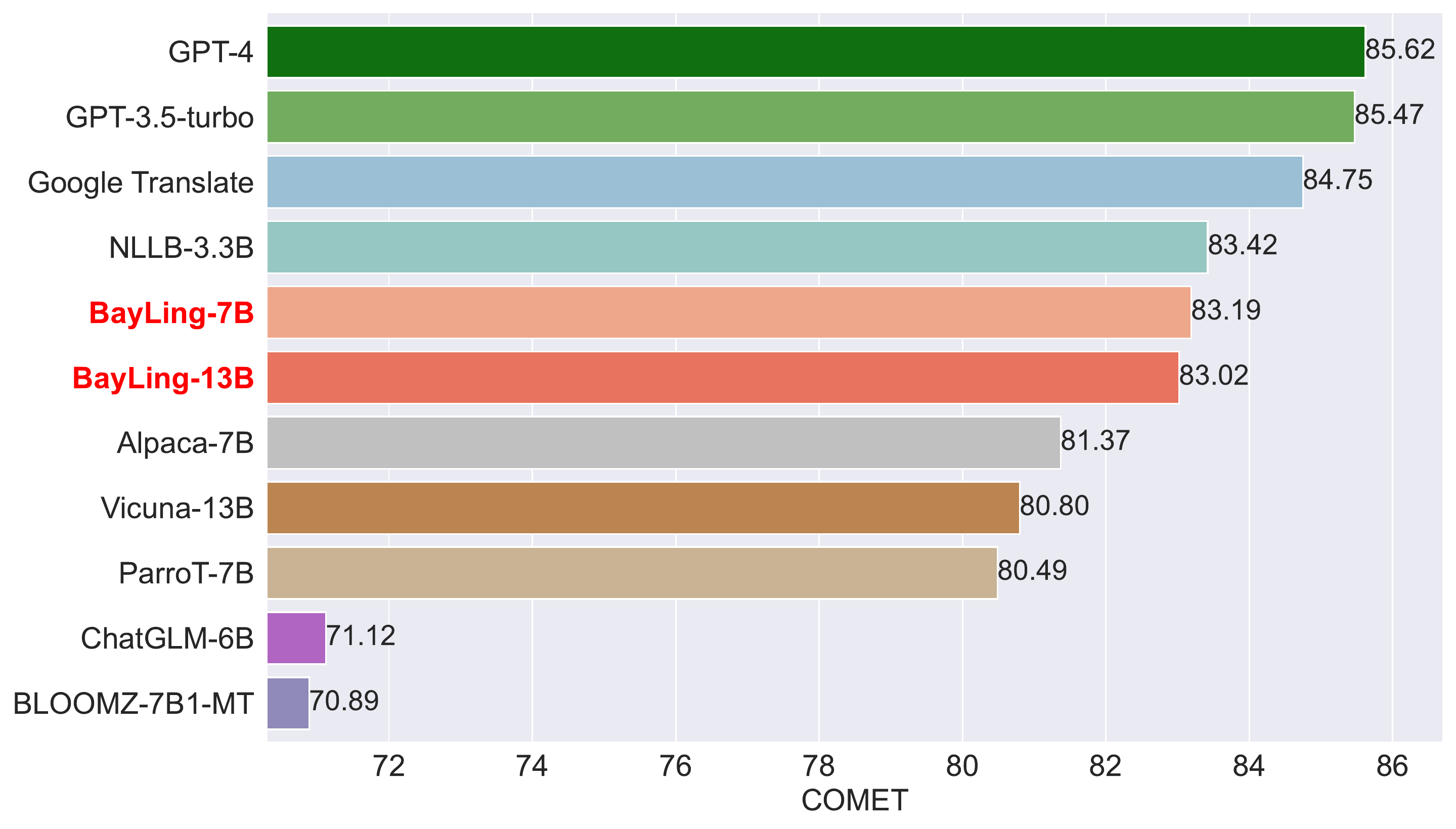} \label{fig:general_mt_deen1}
}
\subfigure[COMET score on English-to-German translation]{
\includegraphics[width=0.48\textwidth]{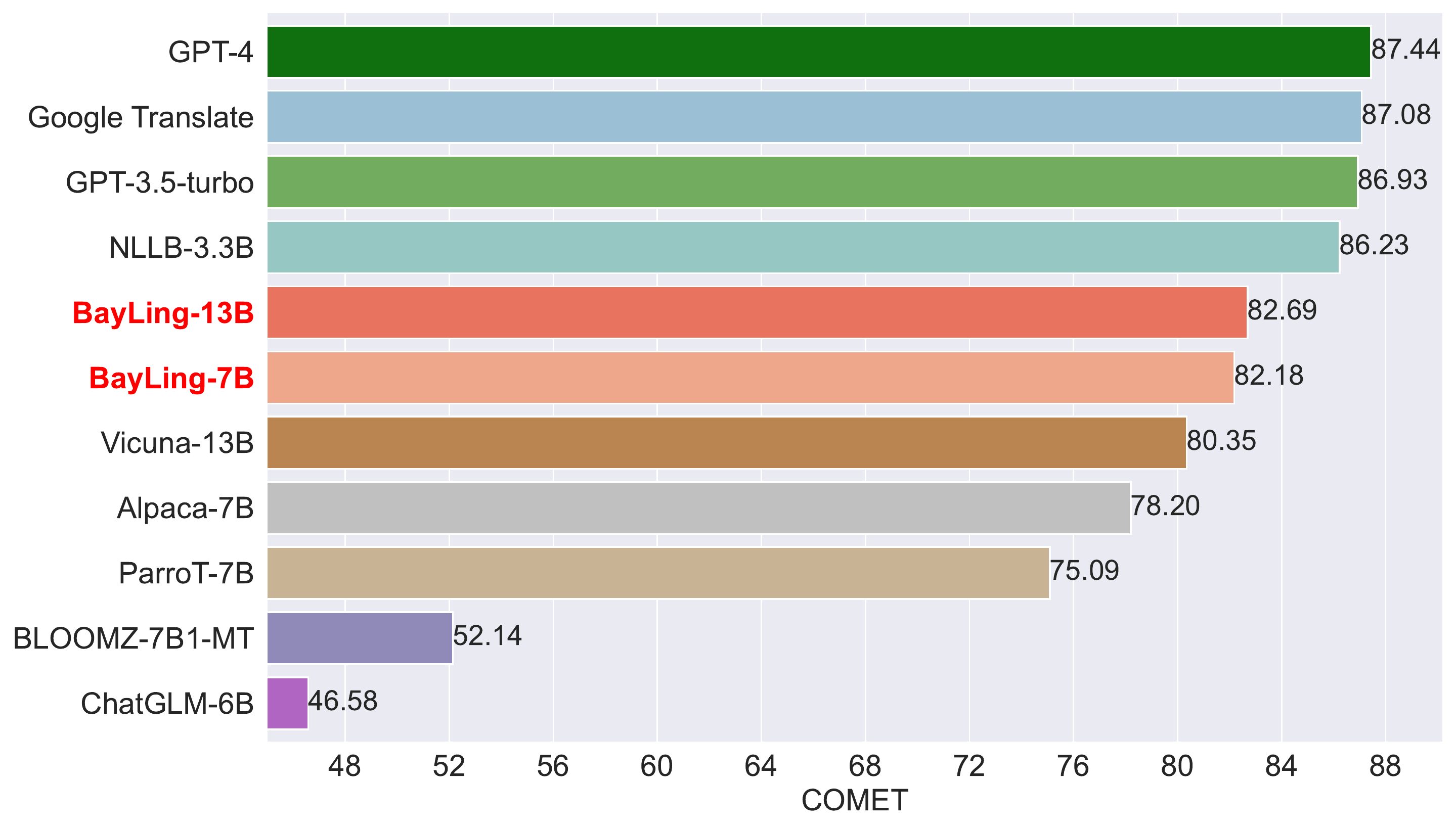} \label{fig:general_mt_deen2}
}
\subfigure[BLEU score on German-to-English translation]{
\includegraphics[width=0.48\textwidth]{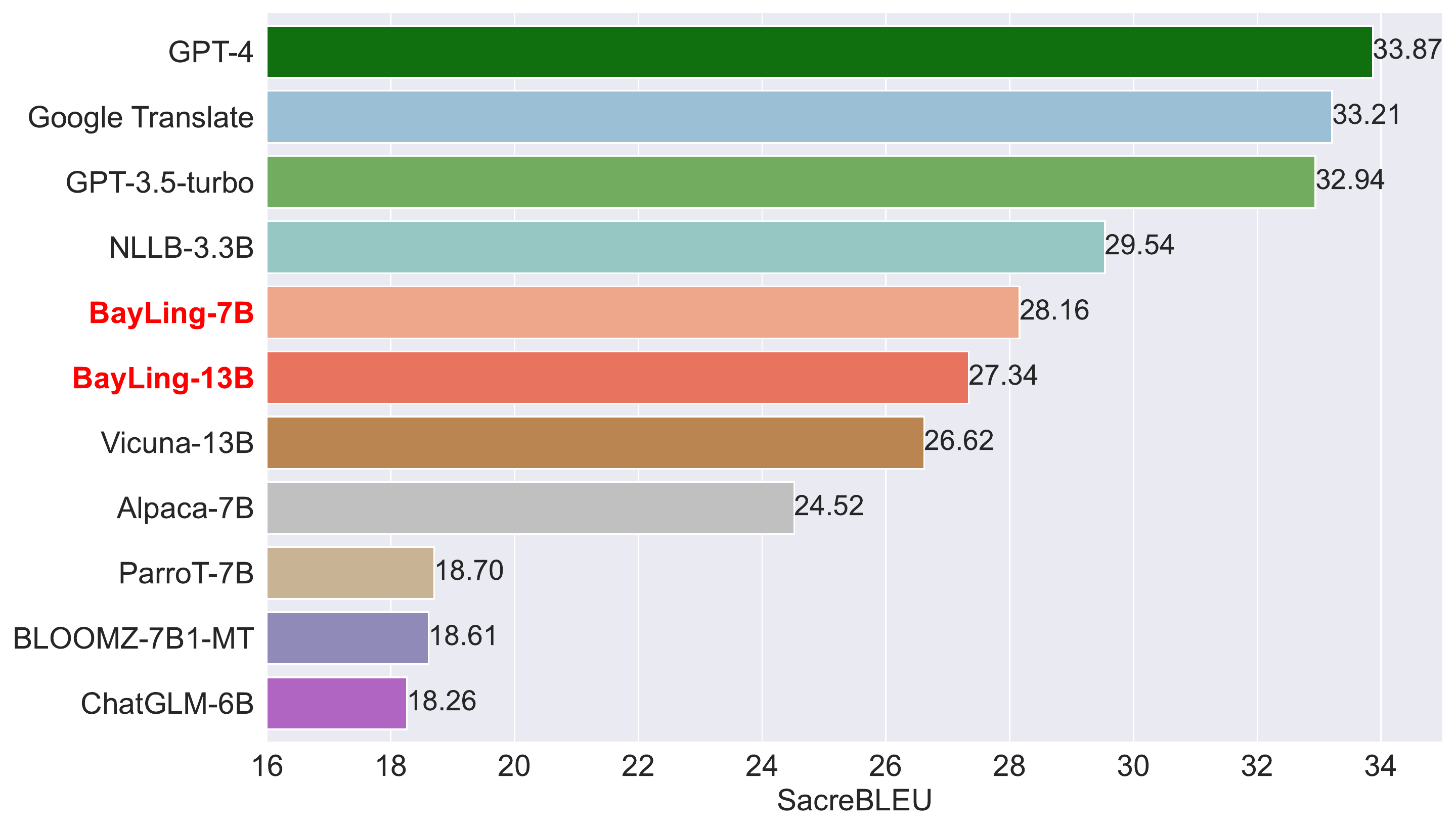} \label{fig:general_mt_bleu_deen1}
}
\subfigure[BLEU score on English-to-German translation]{
\includegraphics[width=0.48\textwidth]{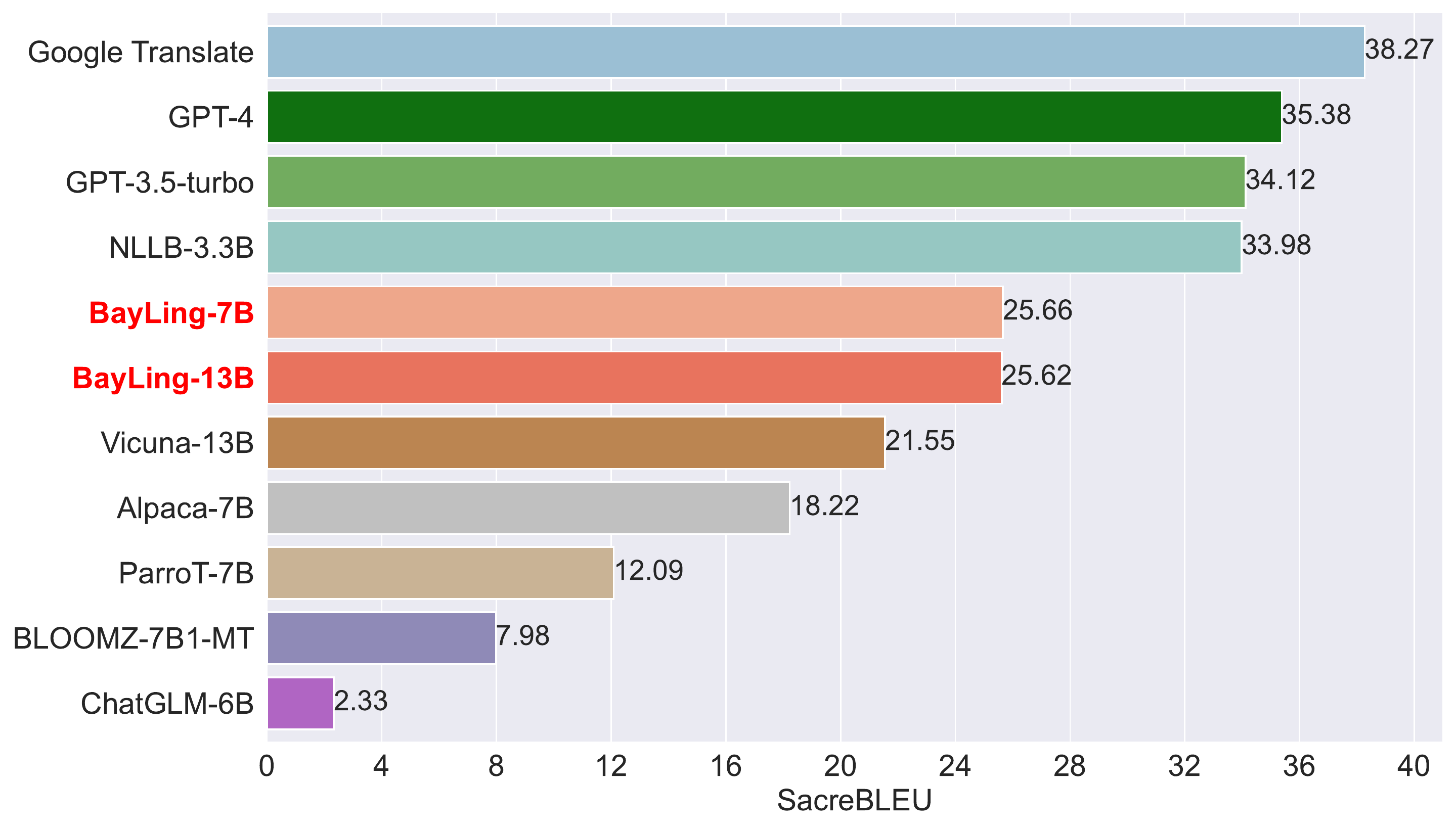} \label{fig:general_mt_bleu_deen2}
}
\caption{Performance comparison on WMT22 German$\Leftrightarrow$English translation task}
\label{fig:general_mt_bleu}
\end{figure}

Figures~\ref{fig:general_mt} and Figures~\ref{fig:general_mt_bleu} depict the performance of BayLing on Chinese$\Leftrightarrow$English and German$\Leftrightarrow$English translations, respectively. For Chinese$\Leftrightarrow$English translation, BayLing exhibits outstanding performance, ranking just behind GPT-4/GPT-3.5-turbo and Google Translate, and is significantly outperforming other open-source systems. Specifically, BayLing-13B achieved 94\% and 97\% of the translation capabilities (COMET score) compared to GPT-4 on Chinese-to-English and English-to-Chinese translation, respectively. In particular, due to pre-training on English-dominant corpus, LLaMA's multilingual ability is comparatively weaker than those foundation LLMs that are pre-trained on the language-balanced corpus, such as BLOOM \cite{workshop2023bloom} and GLM \cite{du-etal-2022-glm}. However, through instruction tuning on the interactive translation task, BayLing effectively enhances the language alignment between Chinese and English within LLaMA, resulting in superior translation performance compared to BLOOMZ-7B1-MT and ChatGLM-6B. For German$\Leftrightarrow$English translation, BayLing also outperforms most systems, achieving 97\% and 95\% of the translation capabilities (COMET score) compared to GPT-4 on German-to-English and English-to-German translation, respectively.

Overall, by fine-tuning on interactive translation instructions, BayLing successfully establishes a high-quality alignment across multiple languages, including Chinese$\Leftrightarrow$English and German$\Leftrightarrow$English, thereby acquiring better translation capabilities and achieving state-of-the-art translation performance among open-source LLMs. Furthermore, this high-quality language alignment sets the stage for extending LLaMA's English proficiency to other languages, such as Chinese. We will further evaluate BayLing's Chinese capability on other general tasks in Section~\ref{sec:general}.

\subsubsection{Zero-Shot Translation}

\begin{figure}
    \centering
    \includegraphics[width=\textwidth]{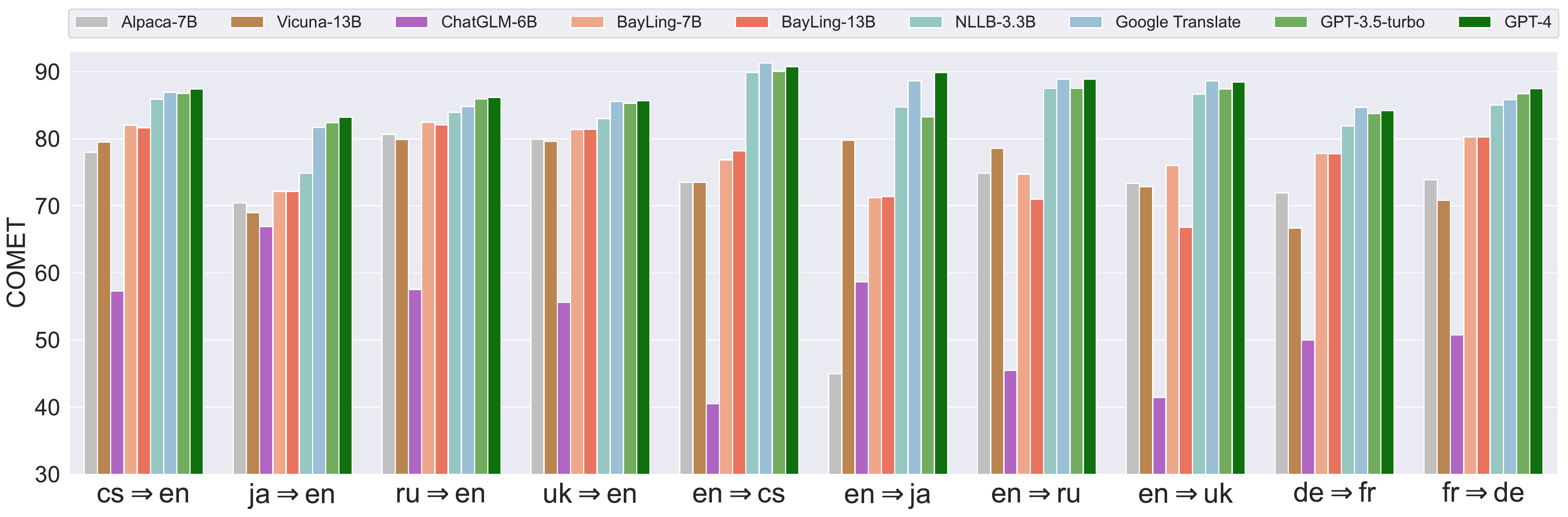}
    \caption{Zero-shot translation performance 
    of BayLing on WMT22 multilingual translation tasks. These translation directions do not appear in BayLing's fine-tuning instructions, but it is uncertain whether they are zero-shot in other models for comparison.}
\label{fig:multilingual_mt}
\end{figure}

In instruction tuning, we mainly use Chinese and English interactive translation data and a small amount of German and French data to equip BayLing with multilingual capabilities. To evaluate BayLing's performance in translation directions that are never seen previously, i.e., zero-shot multilingual capability on out-of-distribution directions, we conduct experiments on WMT22 multilingual translation benchmarks, encompassing 10 translation directions: Czech$\Leftrightarrow$English (cs$\Leftrightarrow$en), Japanese$\Leftrightarrow$English (ja$\Leftrightarrow$en), Russian$\Leftrightarrow$English (ru$\Leftrightarrow$en), Ukrainian$\Leftrightarrow$English (uk$\Leftrightarrow$en) and German$\Leftrightarrow$French (de$\Leftrightarrow$fr). 

As depicted in Figure~\ref{fig:multilingual_mt}, BayLing exhibits good zero-shot multilingual capability on those out-of-distribution translation directions. Compared to Vicuna-13B, BayLing achieves superior translation ability (on average) on these out-of-distribution languages solely through instruction tuning on Chinese and English interactive translation. This highlights that the alignment of other languages is strengthened as a by-product of aligning high-resource languages such as Chinese and English.
Moreover, BayLing exhibits better performance in X$\Rightarrow$English translations (93\% performance compared to GPT-3.5-turbo) compared to English$\Rightarrow$X translations (82\% performance compared to GPT-3.5-turbo), implying that the multilingual understanding capability is more easily transferred across languages than multilingual generation capability.

\subsubsection{Robustness in Instruction Language}

\begin{figure}[t]
\centering

\begin{small}

\begin{tcolorbox}[colback=Emerald!10,colframe=cyan!40!black,title=\textbf{The English Instruction of Translation Task}]
\small
Provide the \{\texttt{target\_language}\} translation of this sentence: \{\texttt{sentence}\}
\end{tcolorbox}

\begin{tcolorbox}[colback=Emerald!10,colframe=cyan!40!black,title=\textbf{The Chinese Instruction of Translation Task}]
\small
\begin{CJK*}{UTF8}{gbsn}提供这句话的\{\texttt{target\_language}\}翻译：\end{CJK*}\{\texttt{sentence}\}
\end{tcolorbox}
\end{small}

\subfigure[Chinese-to-English translation]{
\includegraphics[width=0.48\textwidth]{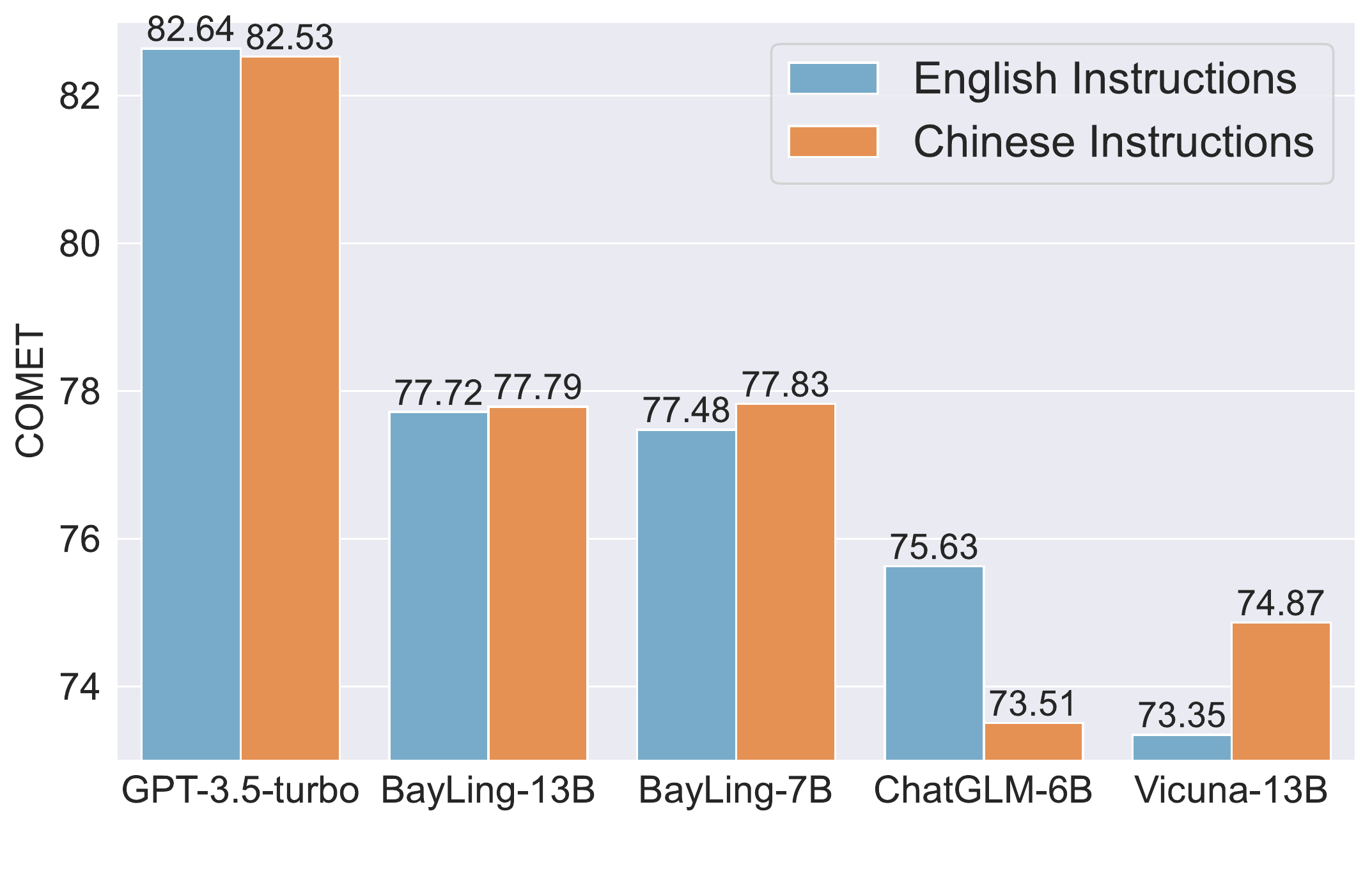} \label{fig:general_mt_instruction_zhen1}
}
\subfigure[English-to-Chinese translation]{
\includegraphics[width=0.48\textwidth]{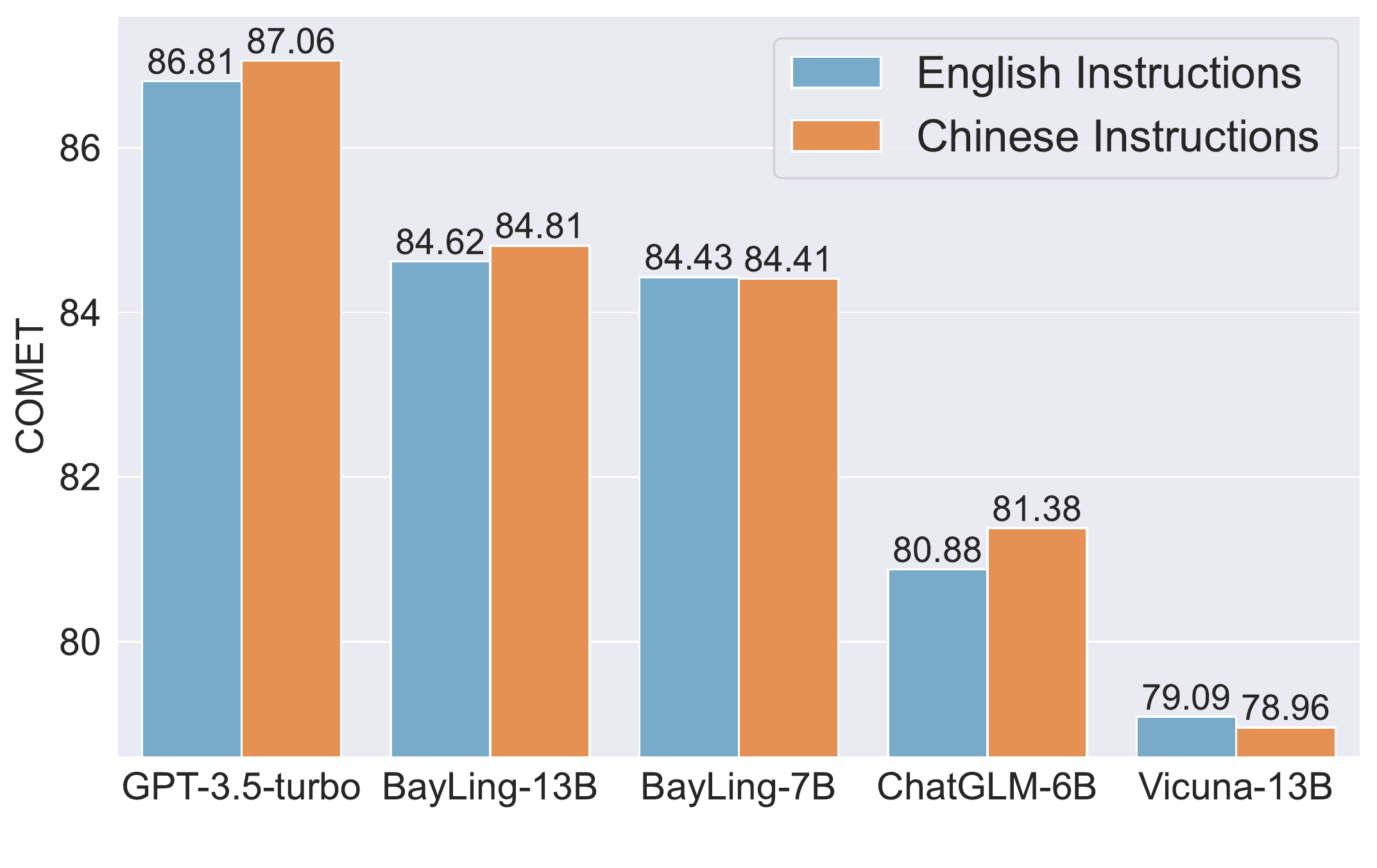} \label{fig:general_mt_instruction_zhen2}
}
\caption{Robustness testing between English and Chinese instructions on Chinese$\Leftrightarrow$English translation tasks}
\label{fig:general_mt_instruction_zhen}
\end{figure}

When applying instruction-following LLMs for translation tasks, previous works often assume that the translation instructions are in English \cite{jiao2023parrot,zhu2023multilingual}. However, when interacting with user, the instruction language may not necessarily be English, especially when the target language for translation is English (i.e., when the user is not a native English speaker). To evaluate the robustness of LLMs given translation instructions in different languages, we show the translation quality of LLMs using both English and Chinese instructions in Figure~\ref{fig:general_mt_instruction_zhen}. 

It is evident that the translation quality of ChatGLM-6B and Vicuna-13B is obviously influenced by the instruction language. ChatGLM-6B performs better when the instruction language corresponds to the target language, while Vicuna-13B prefers the instruction language consistent with the source language. In contrast, GPT-3.5-turbo and BayLing are not sensitive to the language used for translation instructions, achieving similar performance regardless of the instruction language. This characteristic contributes to better alignment with human preferences in interaction using various instruction languages.

\subsubsection{Lexically Constrained Translation}

\begin{figure}[t]
\centering
\begin{small}
\begin{tcolorbox}[colback=Emerald!10,colframe=cyan!40!black,title=\textbf{Instruction of Lexically Constrained Translation Task}]
\small

Provide the \{\texttt{target\_language}\} translation of this sentence: \{\texttt{sentence}\}

Make sure to include these words: \{\texttt{constraint\_1}\}, \{\texttt{constraint\_2}\} $\cdots$

\end{tcolorbox}
\end{small}
\subfigure[Chinese-to-English translation]{
\includegraphics[width=\textwidth]{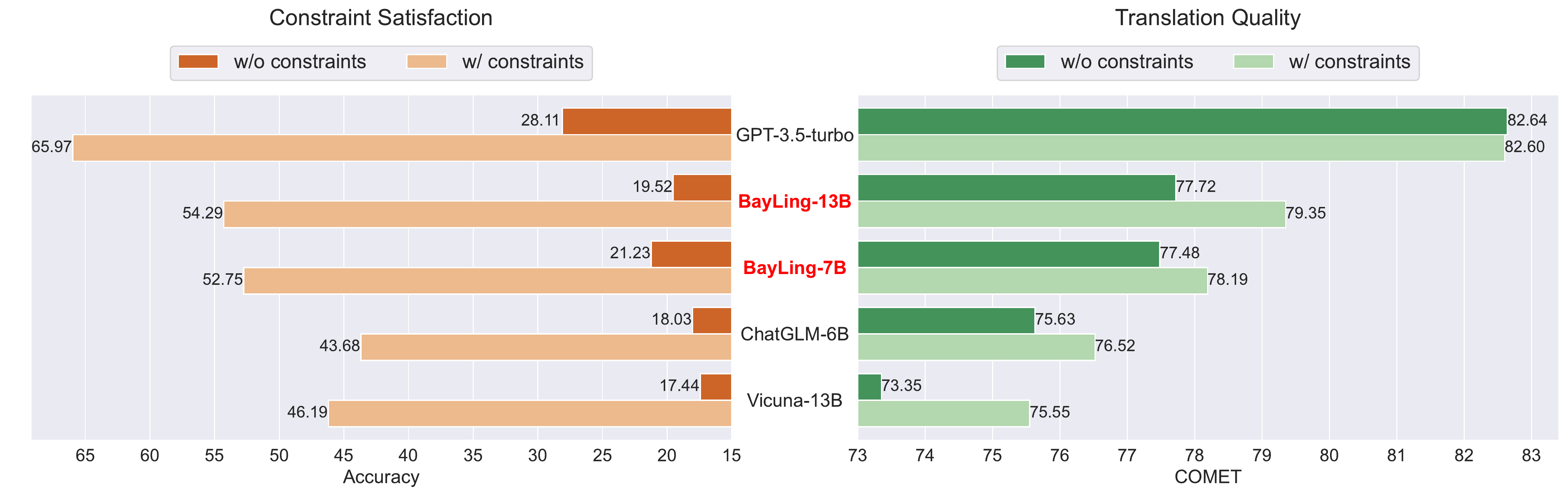} \label{fig:constrained_mt1}
}
\subfigure[English-to-Chinese translation]{
\includegraphics[width=\textwidth]{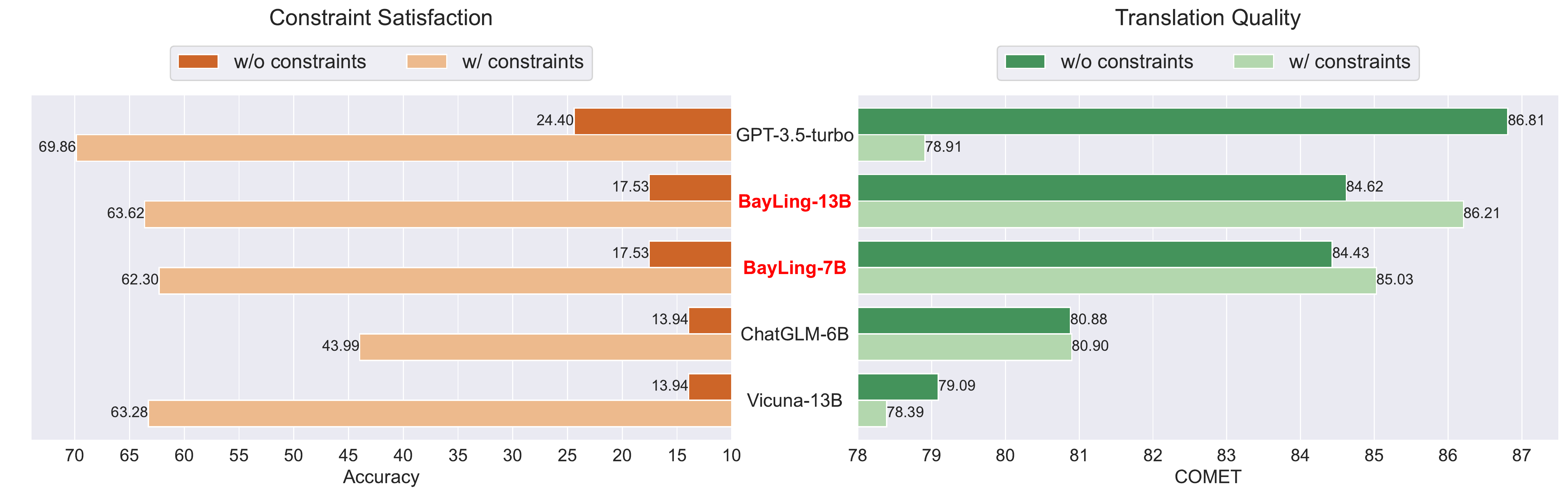} \label{fig:constrained_mt2}
}
\caption{Performance of LLMs on lexically constrained translation task. Constraint satisfaction reveals the ability of LLMs to meet the additional lexical requirements provided by users, while translation quality assesses the capability of LLMs to produce high-quality translations while adhering to the given lexical constraints.}
\label{fig:constrained_mt}
\end{figure}

Compared to translation-specific large models such as Google Translate and NLLB-3.3B, LLMs offer a significant advantage in translation tasks by allowing users to provide additional instructions to constrain the translation results. To evaluate BayLing's ability to satisfy user's additional instructions on translation task, we assess its performance on the task of lexically constrained translation \cite{hu-etal-2019-improved,ijcai2020p0496,jon-etal-2021-end}. In lexically constrained translation, the user not only asks for the translation of the source sentences, but also requires some specific words or phrases that must be included in the translation results. An example of instruction for lexically constrained translation is shown below. As seen, lexically constrained translation task demands LLMs to possess both multilingual and instruction following capabilities to generate translation results that satisfy the specified constraints. 

Figure~\ref{fig:constrained_mt} demonstrates the translation capability of BayLing when subjected to additional constraints, as well as its accuracy in satisfying these constraints. BayLing exhibits a strong following capability of constraint instructions, enabling it to produce constraint-compliant translations in 55\% of the cases. Remarkably, BayLing manages to maintain its translation quality while satisfying the constraints. Conversely, both GPT-3.5-turbo and Vicuna-13B occasionally exhibit a tendency to merely replicate the given constraints, which leads to a noticeable decline in translation quality when tackling English-to-Chinese translation tasks involving constraints.

\subsection{Interactive Translation}
To evaluate how well BayLing aligns with human preferences in terms of the language alignment and multi-turn interactive capabilities, we examine the performance of BayLing with baseline systems on the interactive translation task by human evaluation.

\subsubsection{Experimental Setup}
For human evaluation, we first select 60 sentences, comprising 30 Chinese and 30 English sentences, as source sentences for translation. Then we invite five English-major annotators\footnote{Human annotators are required to demonstrate their English proficiency by passing the Test for English Majors-Band 8 (TEM-8).} to interact with the five systems BayLing-13B, BayLing-7B, ChatGPT\footnote{In human evaluation, we use the web version of ChatGPT, which is more powerful than GPT-3.5-turbo API.}, Vicuna-13B and ChatGLM-6B on these 60 translated sentences for four turns. Note that the system identities are concealed and randomly shuffled, ensuring that annotators remained unaware of the system they were interacting with. During the interactive process, we impose a requirement for annotators to use Chinese for half of the cases and English for the other half, ensuring the diversity of instruction languages. Furthermore, we design five distinct interaction categories across the 60 cases: vocabulary, grammar, style, suggestion and creation (i.e., each category consists of 12 cases), enabling the analyses of the system's performance when confronted with various types of requirements.

\textbf{Scoring \& Ranking}\quad For each case, we request the annotators to simultaneously assess the interaction process of the five systems and assign scores ranging from 1 to 10 from three aspects of translation quality, instruction following, and multi-turn interactive performance (one score for each aspect). Furthermore, we request the annotators to give the rank of five systems from 1 to 5 from these three aspects respectively. Finally, we averaged the scores and rankings of the five annotators as the final score and ranking of the five systems.

\subsubsection{Results from Human Evaluation}

\begin{figure}[t]
\centering
\subfigure[Translation]{
\includegraphics[width=0.3\textwidth]{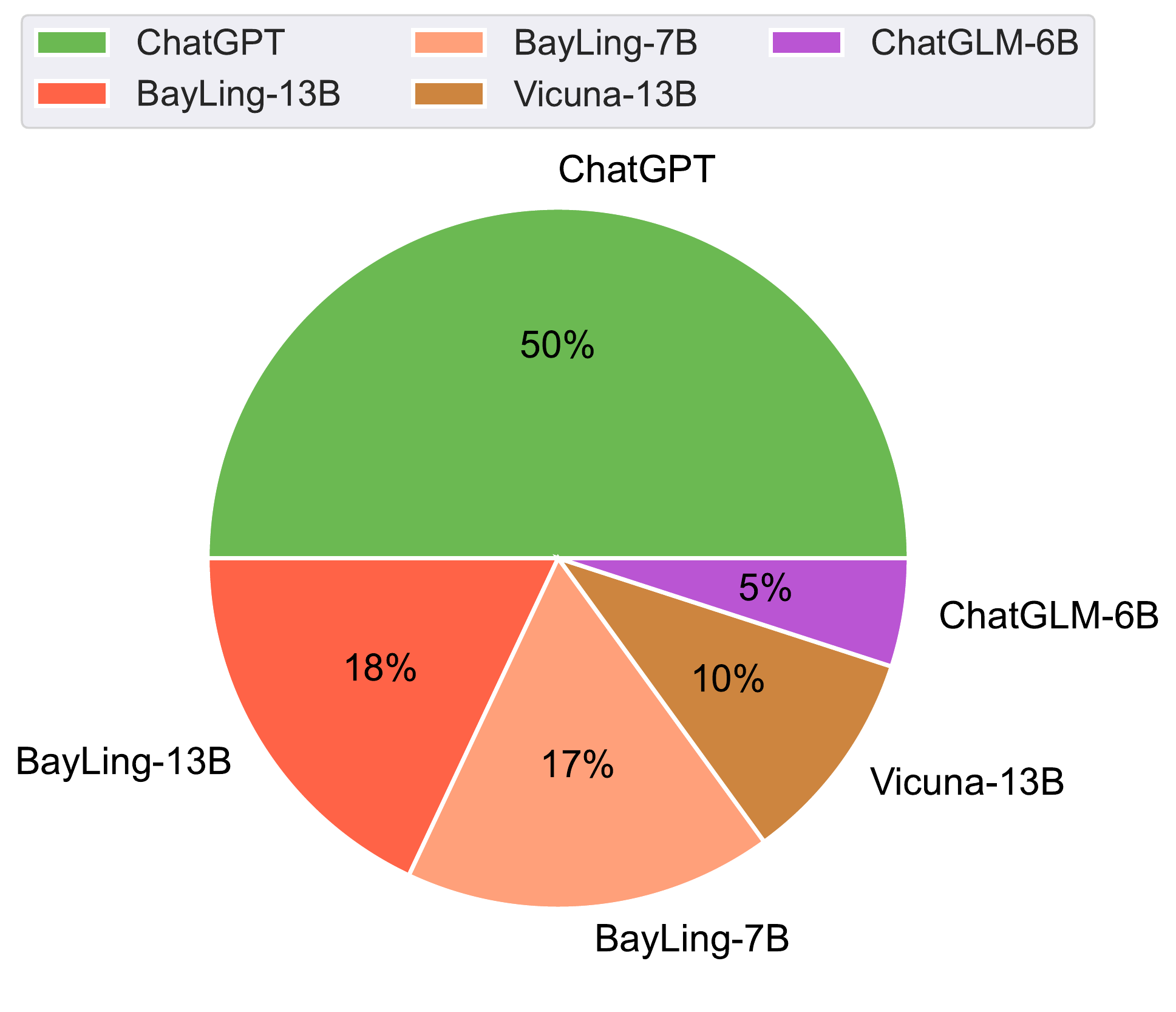} \label{fig:imt_rank1}
}
\subfigure[Instruction following]{
\includegraphics[width=0.31\textwidth]{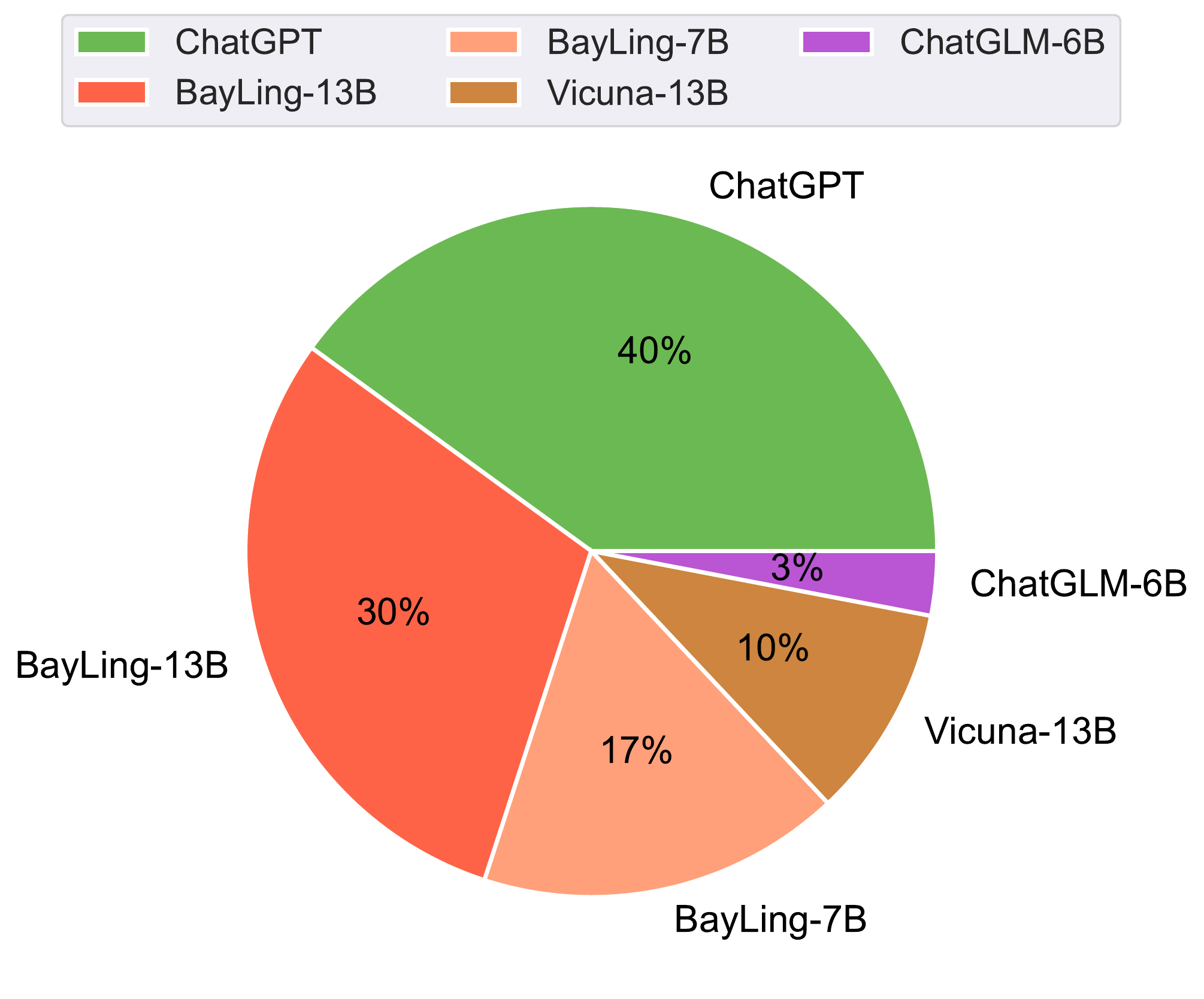} \label{fig:imt_rank2}
}
\subfigure[Multi-turn interaction]{
\includegraphics[width=0.303\textwidth]{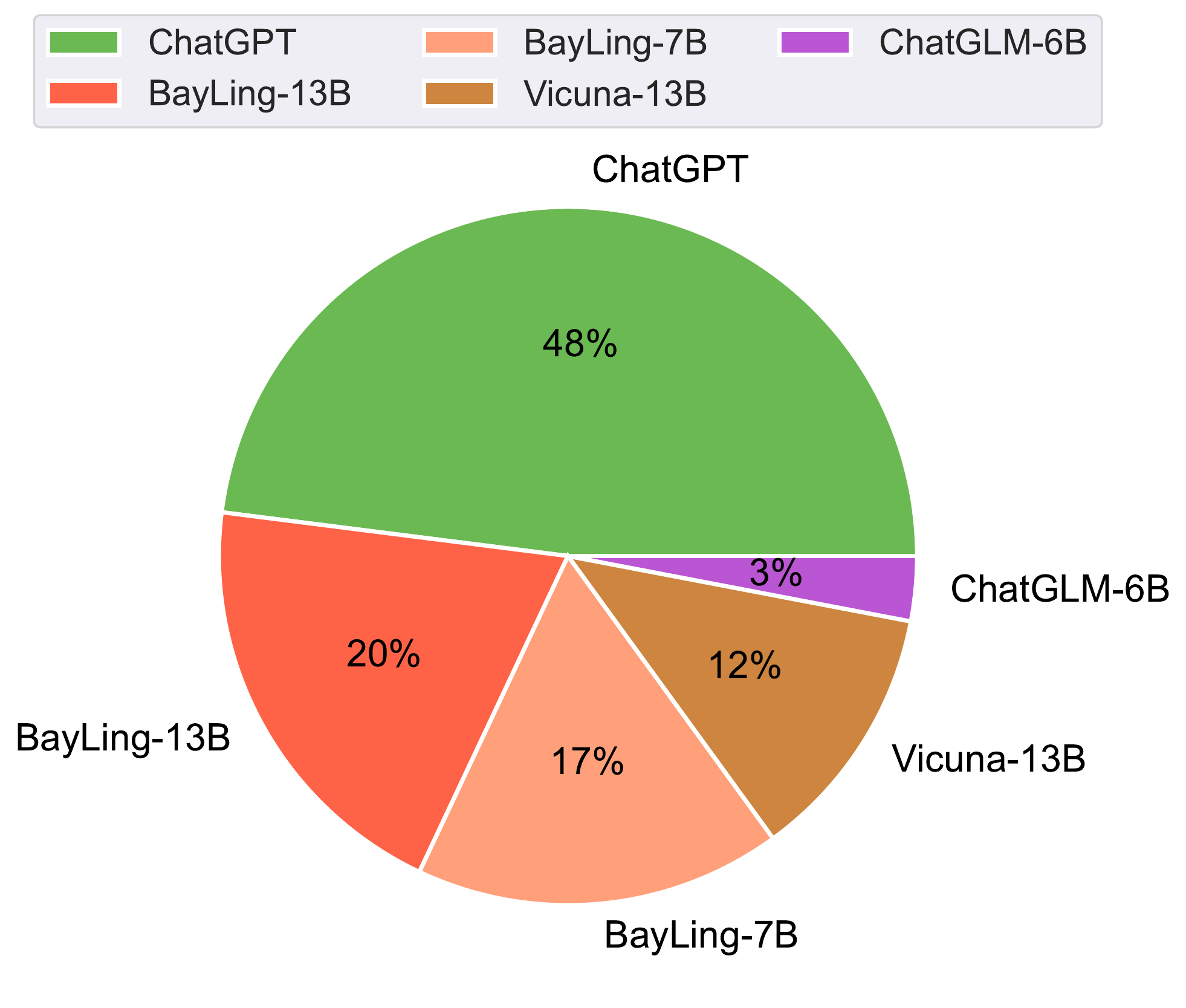} \label{fig:imt_rank3}
}
\caption{Proportions by which each of the five systems wins out in human evaluation of interactive translation task. Human evaluation is given separately from three aspects of capability, including translation, instruction following and multi-turn interaction. (a) Translation: assesses the overall translation proficiency of LLMs. (b) Instruction following: evaluates the extent to whether LLMs can successfully fulfill instructions and satisfy requirements from human. (c) Multi-turn interaction: measures the LLMs' capability to comprehend and maintain coherence across multiple turns of dialogue, testing their contextual understanding and consistency.}
\label{fig:imt_rank}
\end{figure}

\begin{figure}[t]
\centering
\subfigure[Translation]{
\includegraphics[width=0.28\textwidth]{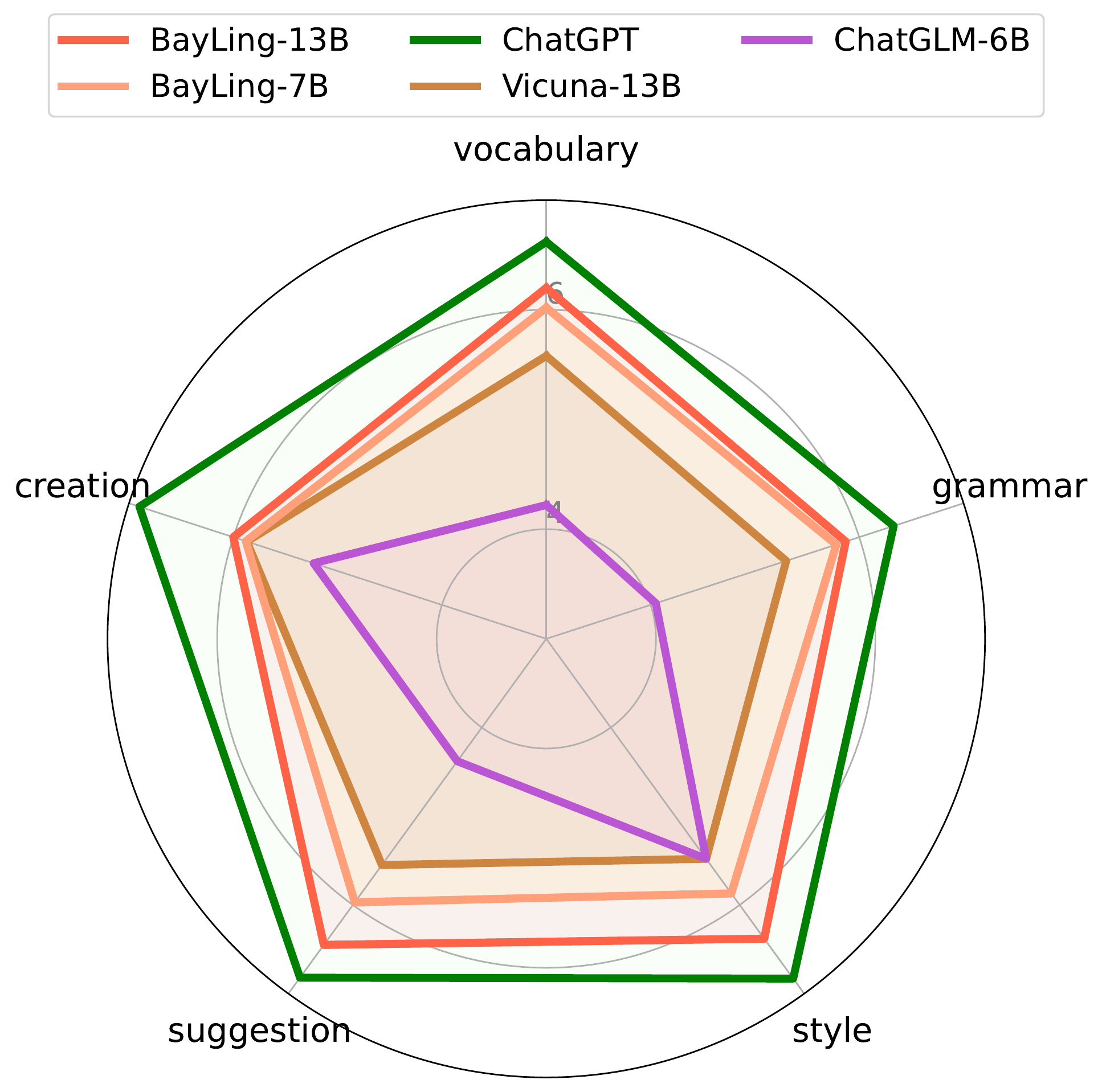} \label{fig:imt1}
}\quad
\subfigure[Instruction following]{
\includegraphics[width=0.28\textwidth]{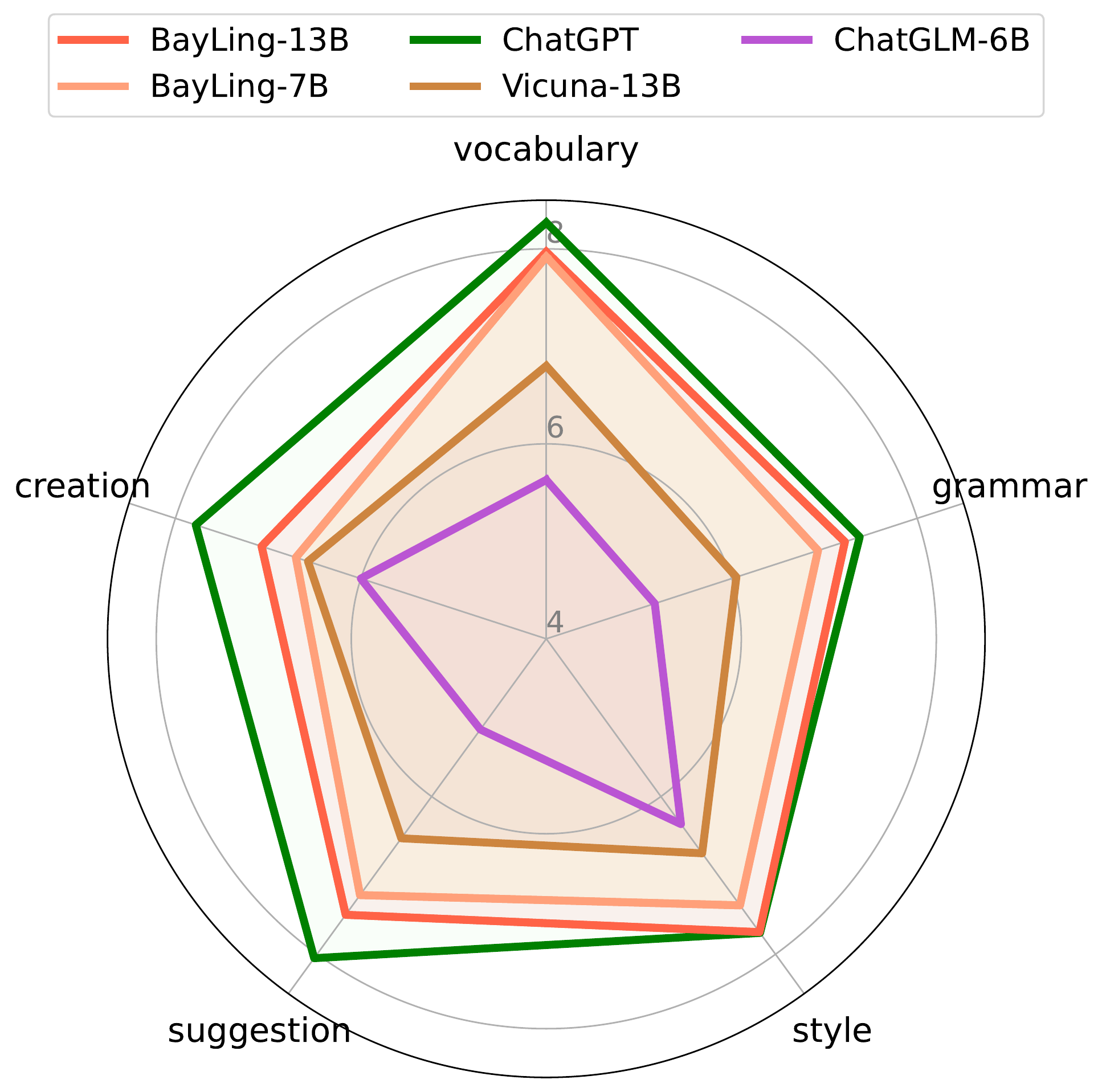} \label{fig:imt2}
}\quad
\subfigure[Multi-turn interaction]{
\includegraphics[width=0.28\textwidth]{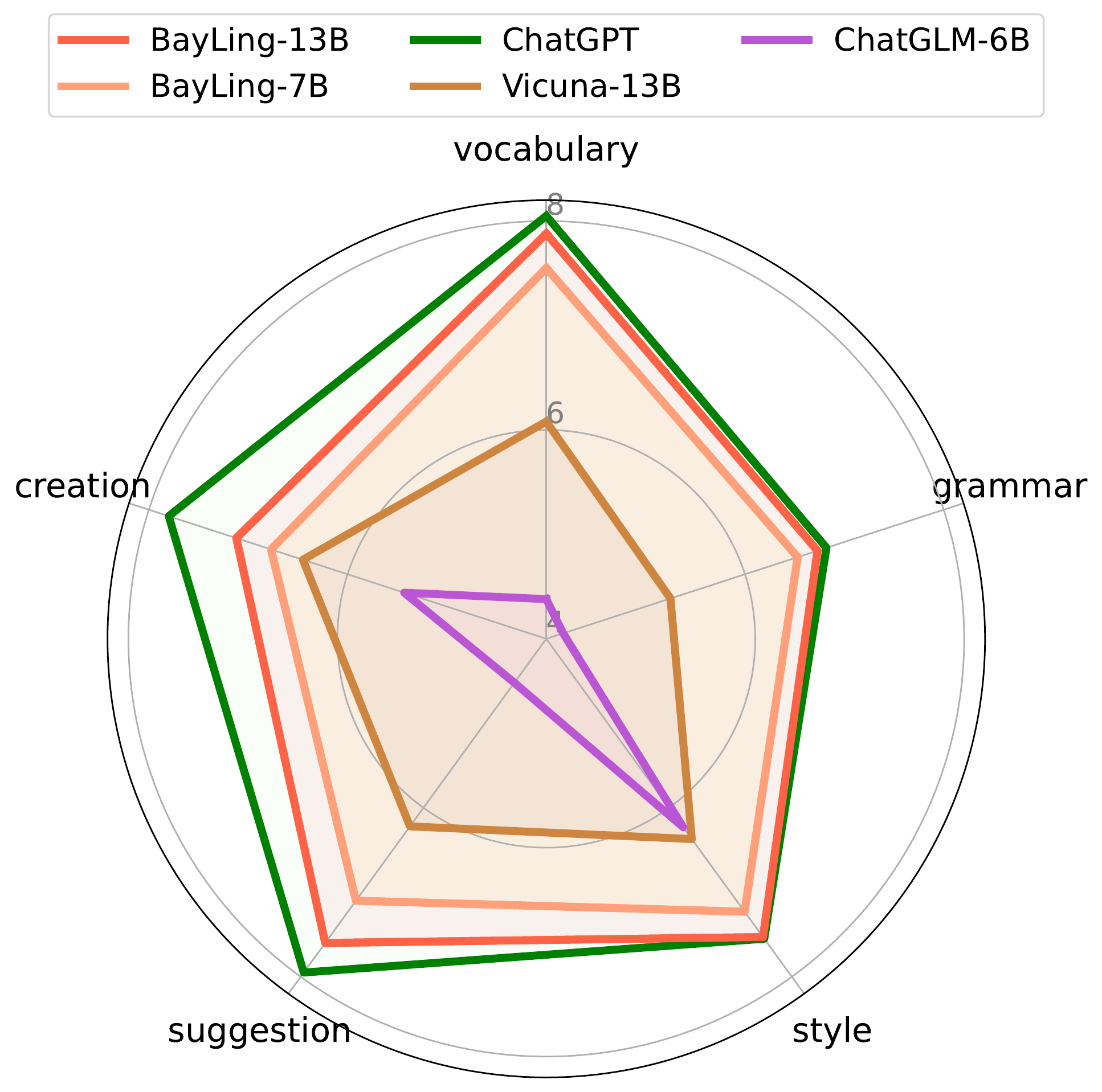} \label{fig:imt3}
}
\caption{Average scores for different models on five interactive categories (vocabulary, grammar, style, suggestion and creation) in human evaluation of interactive translation task.}
\label{fig:imt}
\end{figure}

Figure~\ref{fig:imt_rank} presents the proportion of achieving Rank\#1 among five systems in human evaluation. In terms of evaluating the capability of translation, instruction following and multi-turn interaction, BayLing-13B receives the highest rating from humans in 18\%, 30\% and 20\% of the cases respectively, placing second only to ChatGPT. In particular, favorable performance regarding instruction following and multi-turn interactive capabilities in human evaluation demonstrates that BayLing successfully integrates language alignment, instruction following and multi-turn interactive capabilities through the instruction tuning on composite task, thereby achieving better alignment with human preferences.

Furthermore, we refine BayLing's average scores on five different types of interactions in Figure~\ref{fig:imt} to investigate its specific interactive capabilities. As depicted, for the instruction following and multi-turn interaction, BayLing-13B exhibits comparable performance to ChatGPT in terms of those specific instructions, such as vocabulary, grammar and style requirements. However, when faced with instructions with more abstract meanings, such as creative requirements and suggestions from humans, BayLing-13B still shows room for improvement compared to ChatGPT.

\subsection{General Tasks}
\label{sec:general}

\begin{figure}[t]
\centering
\subfigure[Comparison on single-turn English instructions]{
\includegraphics[width=0.48\textwidth]{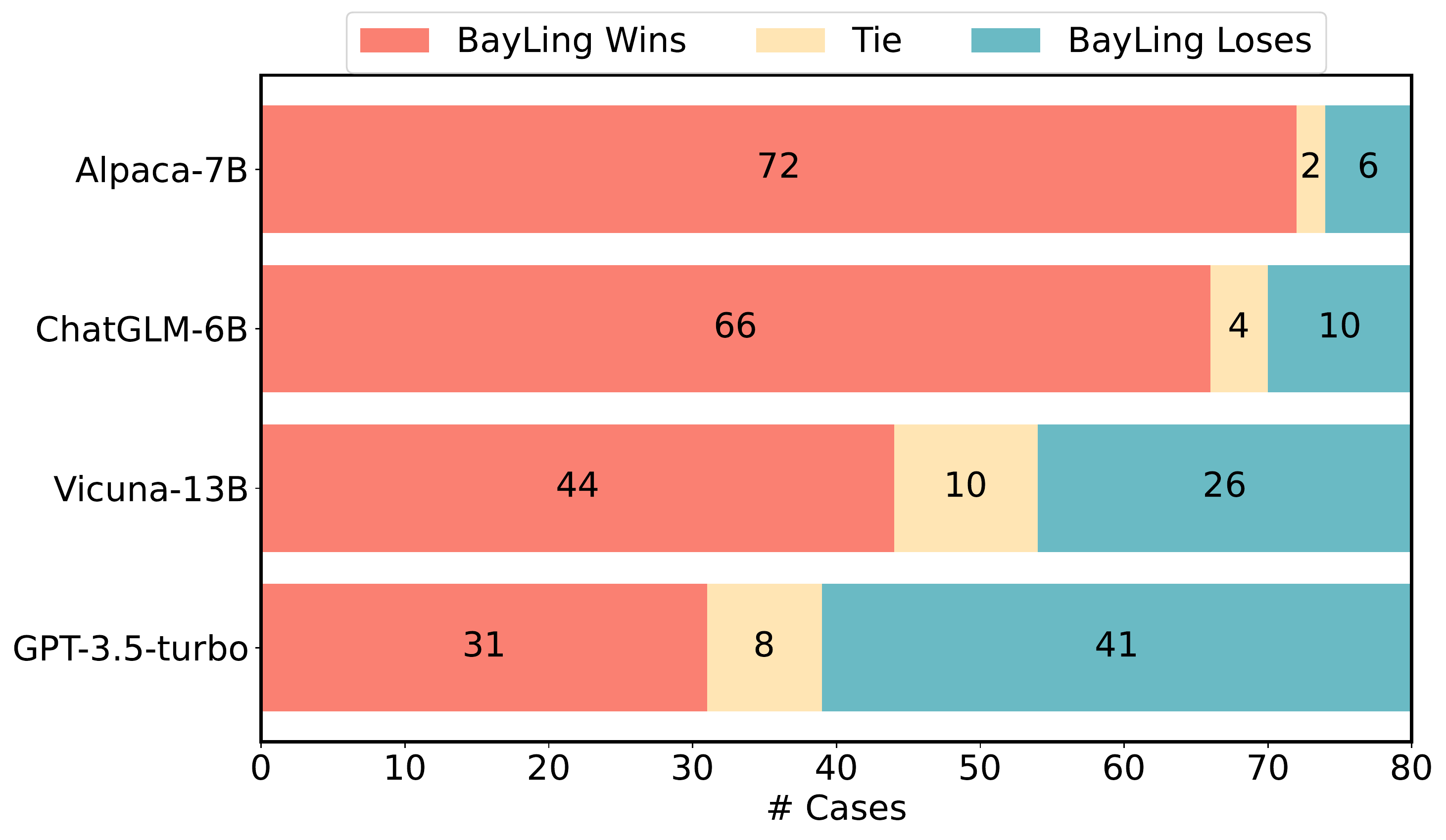} \label{fig:general_task1}
}
\subfigure[Comparison on single-turn Chinese instructions]{
\includegraphics[width=0.48\textwidth]{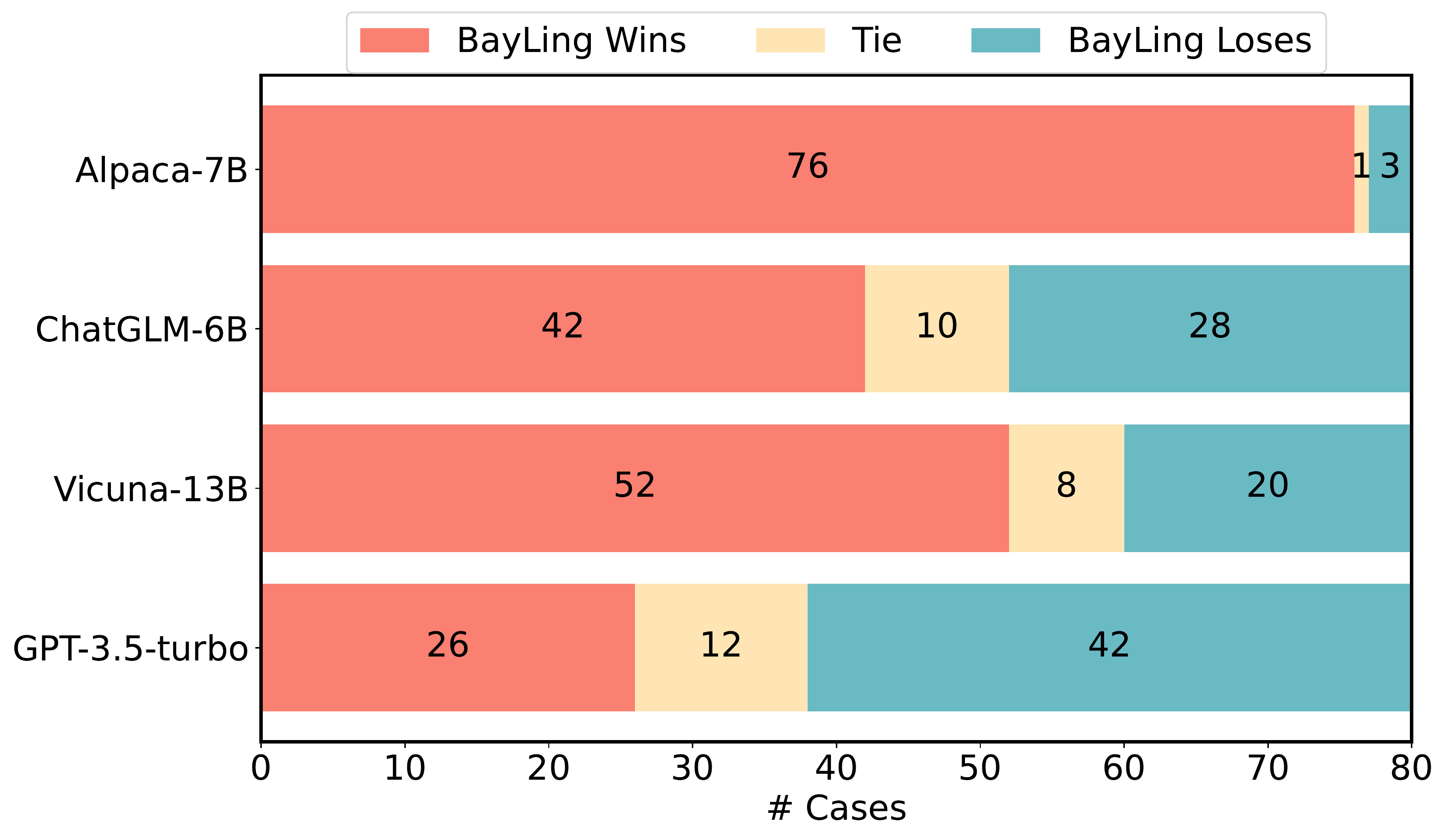} \label{fig:general_task2}
}
\subfigure[Comparison on multi-turn English instructions]{
\includegraphics[width=0.48\textwidth]{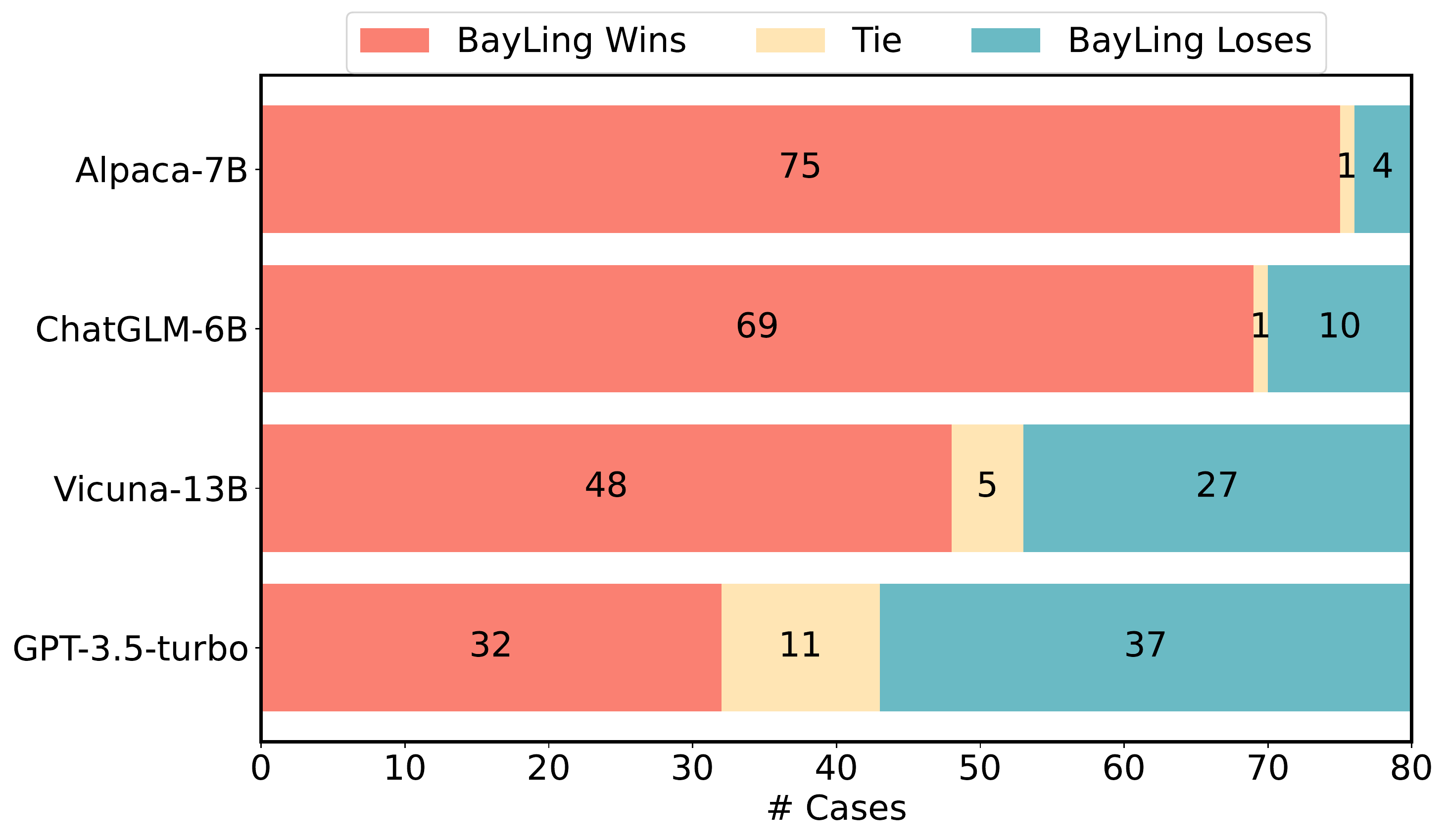} \label{fig:general_task_multiturn1}
}
\subfigure[Comparison on multi-turn Chinese instructions]{
\includegraphics[width=0.48\textwidth]{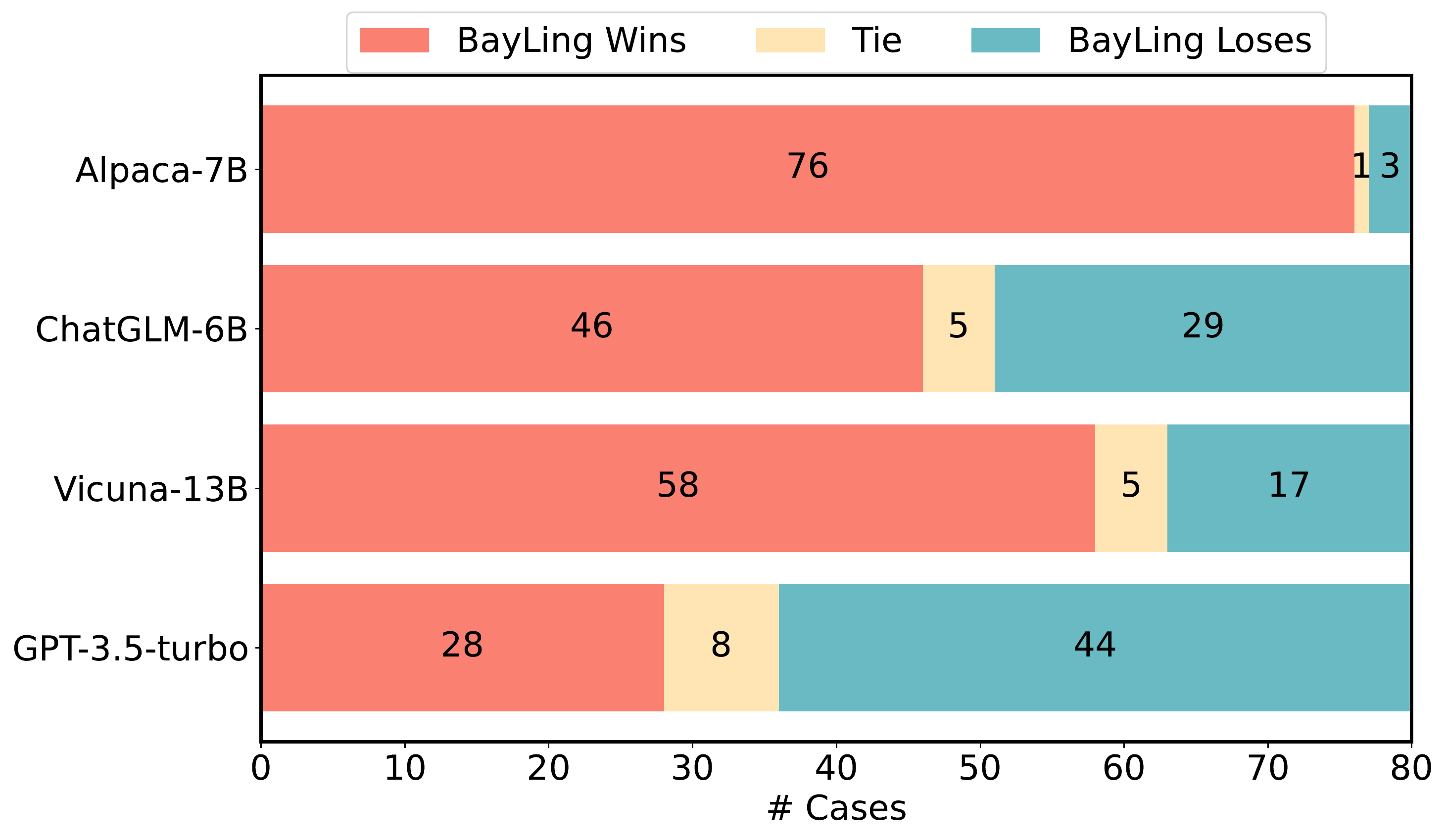} \label{fig:general_task_multiturn2}
}
\caption{GPT-4 evaluation results on general tasks. The responses of BayLing are paired with the responses of GPT-3.5-turbo/Vicuna-13B/ChatGLM-6B/Alpaca-7B to be evaluated by GPT-4.}
\label{fig:general_task}
\end{figure}

In addition to evaluating BayLing's multilingual and multi-turn interactive capabilities on various translation tasks, we further assess these capabilities of BayLing on general tasks.

\subsubsection{BayLing-80 Test Set}
To evaluate the performance on general tasks, we employ the widely-used 80-sentence general task test set introduced by Vicuna\footnote{\url{https://github.com/lm-sys/FastChat}} \cite{vicuna2023}, which comprises 80 single-turn instructions in English. However, in order to assess the multilingual and multi-turn interactive capabilities of LLMs, we extend this test set from language and interaction turns, thus creating BayLing-80 test set. To achieve this, we first extend the 80 single-turn instructions to two-turn instructions by annotating the additional second-turn instruction, where the second-turn instruction closely relied on the first-turn instruction and encompassed hidden references or logical relations. Then, we translate these English instructions into Chinese. Finally, we create the BayLing-80\footnote{BayLing-80 test set is released at: \url{https://github.com/ictnlp/BayLing/tree/main/data/BayLing-80}} test set, which encompasses four sets of single-turn/multi-turn instructions in both English and Chinese. By adopting BayLing-80 test set, we aimed to comprehensively evaluate the multilingual and interactive capabilities of LLMs. Refer to Appendix~\ref{app:BayLing-80} for the specific construction process and examples of BayLing-80 test set.

\subsubsection{Results from GPT-4 Evaluation}
Following \citet{liu2023geval}, we use GPT-4 as an evaluator to measure the alignment of LLMs and human preferences.
Specifically, we send the responses of systems on the BayLing-80 test set to GPT-4 in pairs, and ask GPT-4 to review and score the responses of two comparison systems from 1 to 10, thereby selecting the system with the higher score as the winner. The specific templates provided to GPT-4 for scoring are shown in Appendix~\ref{app:template}.

\textbf{BayLing-13B v.s. Vicuna-13B/ChatGLM-6B/Alpaca-7B}\quad We show the comparison results between BayLing-13B and baseline systems in Figure~\ref{fig:general_task}. Compared with Alpaca-7B, BayLing has a significant advantage and achieves higher scores in almost all cases. When compared to ChatGLM-6B, BayLing-13B demonstrates a clear advantage in English instruction, showcasing superior performance in 80\% of cases. Regarding Chinese instruction, although the foundation model LLaMA does not exhibit the same level of Chinese proficiency as GLM, BayLing-13B enhances its Chinese capability by aligning Chinese to English based on Chinese$\Leftrightarrow$English interactive translation data, thereby outperforming ChatGLM-6B in 50\% of cases with Chinese instructions. Compared to Vicuna-13B, which is also based on LLaMA, BayLing-13B also exhibits significant advantages, especially in Chinese instructions. This demonstrates that BayLing enhances its language alignment and successfully transfers LLaMA's powerful English capabilities to Chinese, thereby achieving superior Chinese proficiency. Furthermore, when it comes to multi-turn instructions, BayLing-13B's performance superiority to other systems becomes more pronounced compared to single-turn interactive tasks, highlighting that interactive translation aids BayLing to get improved multi-turn interactive capability. 

\begin{figure}[t]
\centering
\subfigure[Single-turn, English]{
\includegraphics[width=0.232\textwidth]{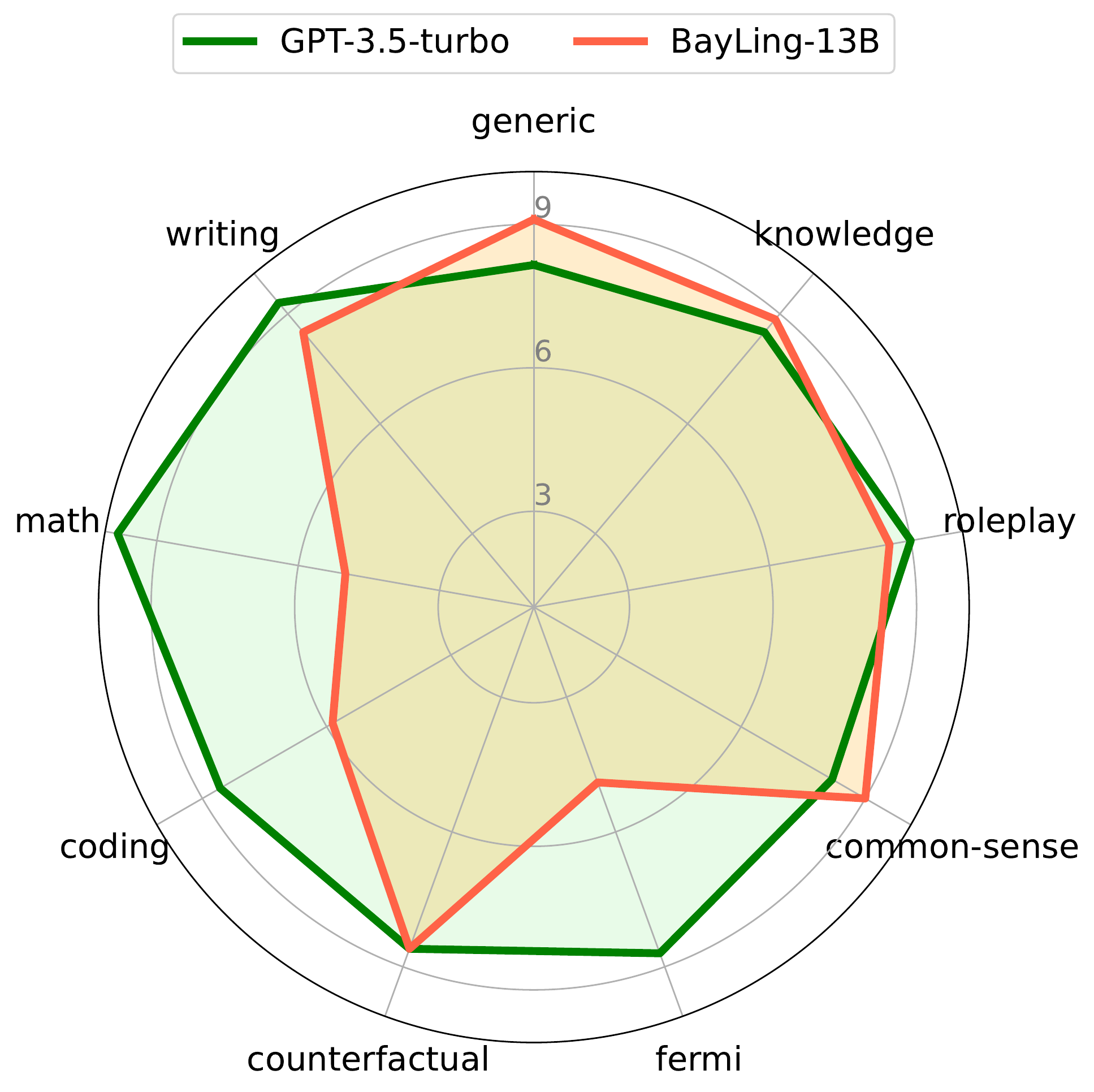} \label{fig:cap1}
}
\subfigure[Single-turn, Chinese]{
\includegraphics[width=0.232\textwidth]{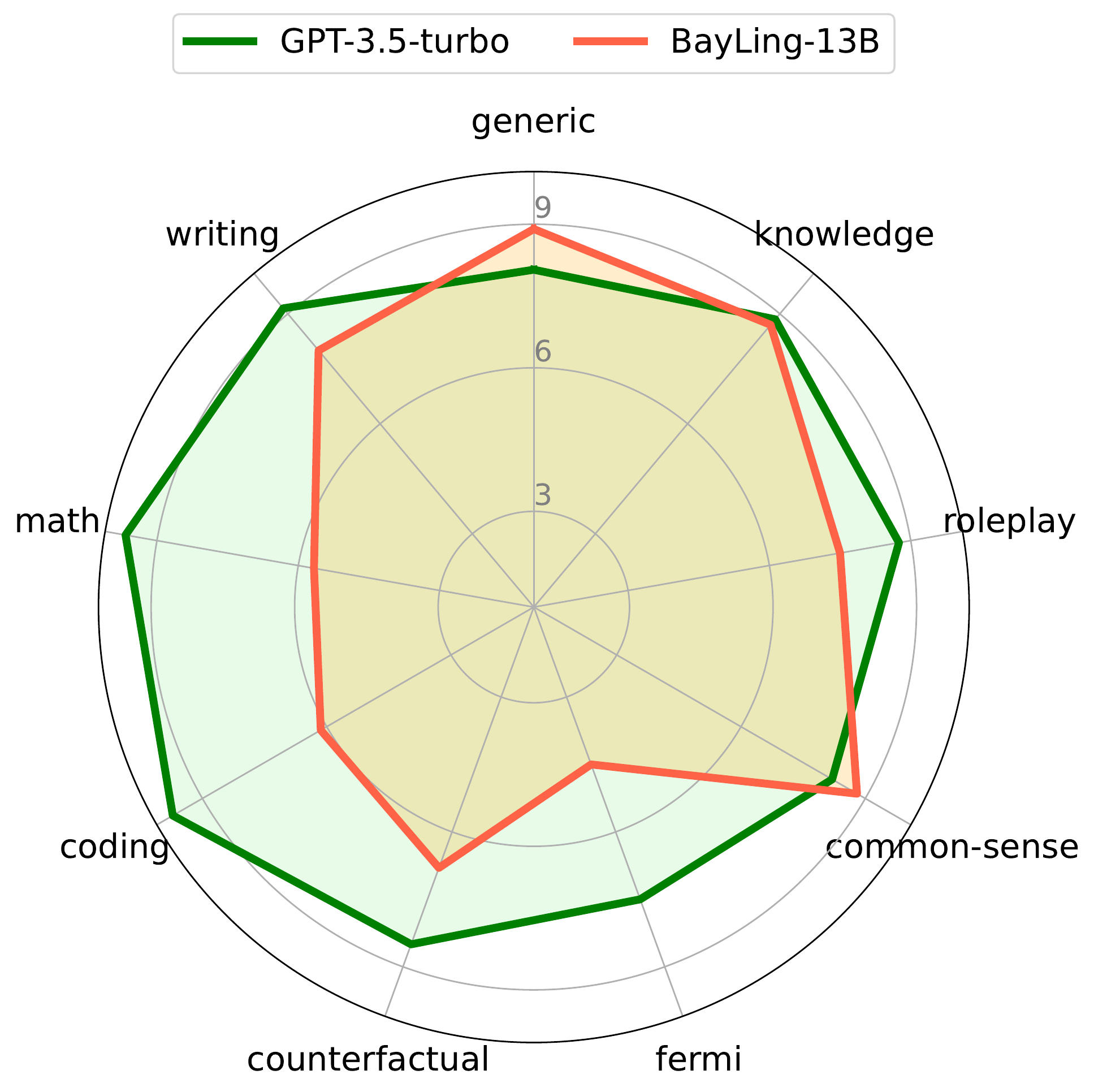} \label{fig:cap2}
}
\subfigure[Multi-turn, English]{
\includegraphics[width=0.232\textwidth]{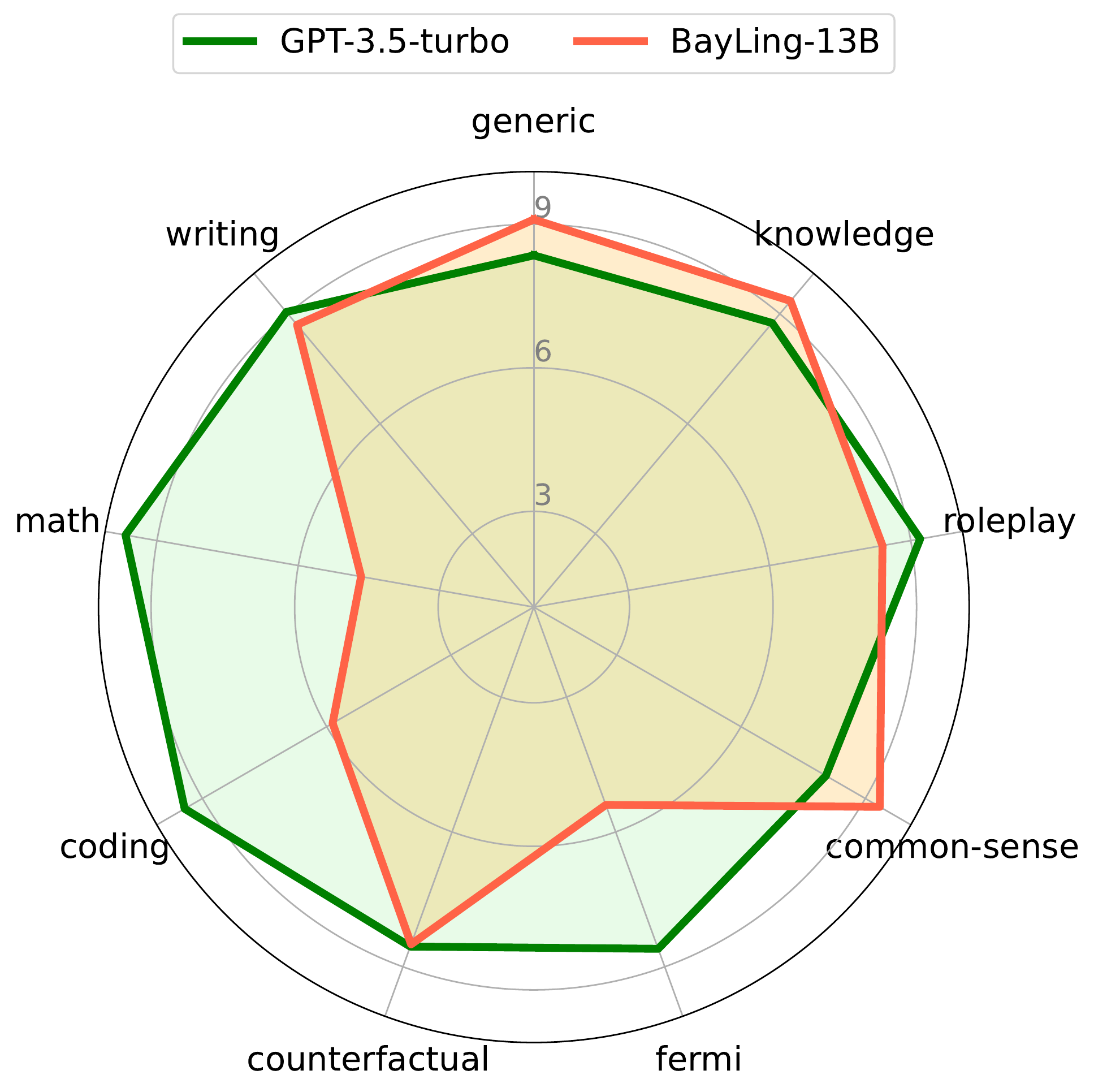} \label{fig:cap3}
}
\subfigure[Multi-turn, Chinese]{
\includegraphics[width=0.232\textwidth]{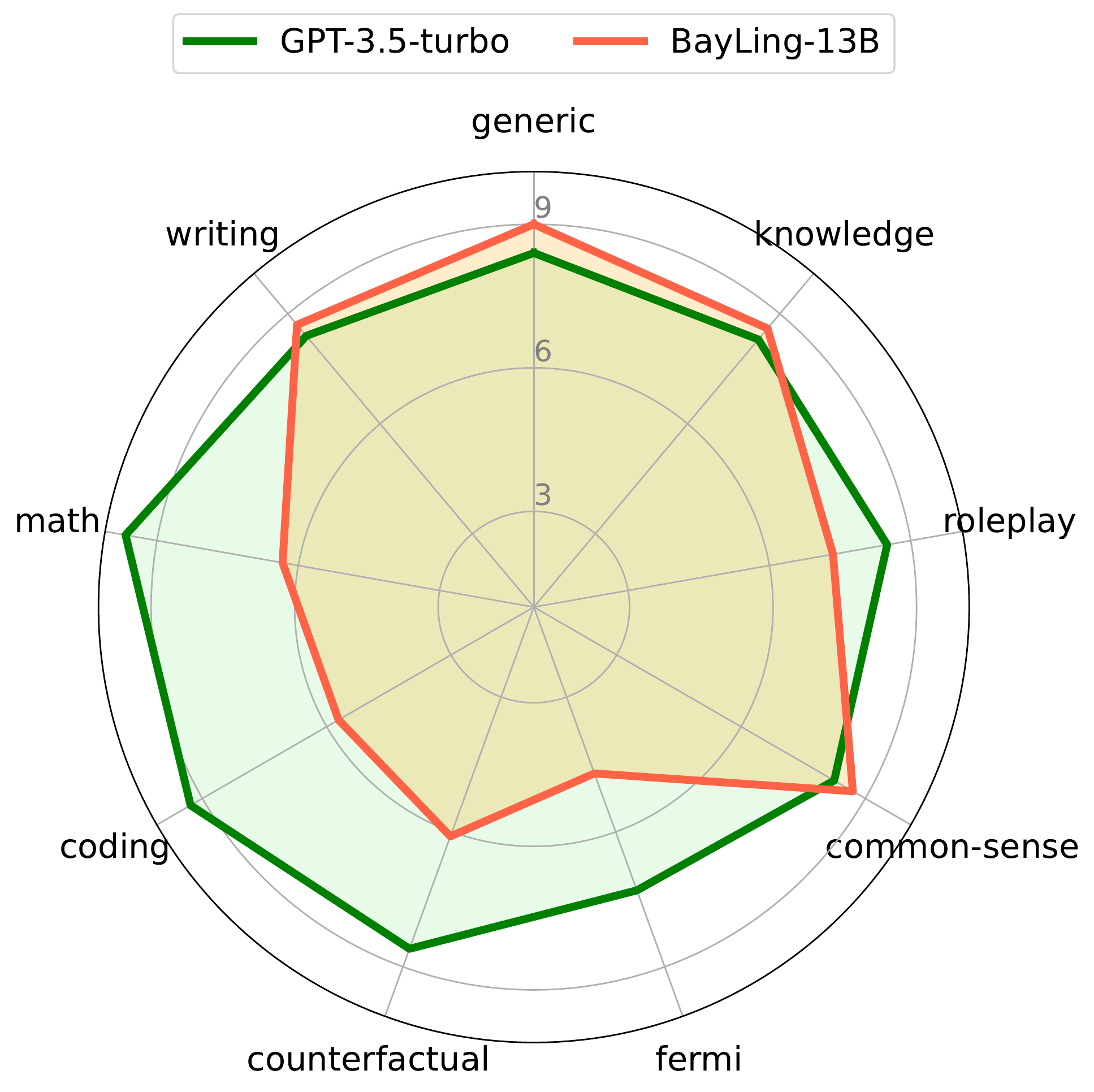} \label{fig:cap4}
}
\caption{Average scores of BayLing-13B against GPT-3.5-turbo on 9 categories of general tasks}
\label{fig:cap}
\end{figure}

\begin{table}
\caption{Overall scores of BayLing-13B compared to GPT-3.5-turbo on general tasks}
\label{table:total}
\advance\leftskip+1mm
\small
\centering
\begin{tabular}{llccC{1.5cm}}\toprule
\multicolumn{2}{c}{Instruction}                  &GPT-3.5-turbo&BayLing-13B&Ratio \\ \cmidrule(lr){1-2}\cmidrule(lr){3-4}\cmidrule(lr){5-5}
\multirow{2}{*}{Single-turn}&English&694.0        &631.0      &91\% \\
                            &Chinese&687.0        &592.0      &86\% \\
\multirow{2}{*}{Multi-turn} &English&700.5        &643.0      &92\% \\
                            &Chinese&671.5        &590.5      &88\% \\\cmidrule(lr){1-2}\cmidrule(lr){3-4}\cmidrule(lr){5-5}
\multicolumn{2}{c}{Average}               &688.3        &614.1      &89\% \\\bottomrule
\end{tabular}
\end{table}

\textbf{BayLing-13B v.s. GPT-3.5-turbo}\quad Compared to GPT-3.5-turbo, BayLing-13B outperforms GPT-3.5-turbo in 35\% of cases when evaluated by GPT-4, and is not worse than GPT-3.5-turbo in 50\% of cases. To further analyze the various capabilities of BayLing-13B, we conducted a detailed comparison of scores between BayLing-13B and GPT-3.5-turbo on various types of general tasks. The test set consisted of 80 cases, which can be categorized into 9 groups: generic (10), knowledge (10), roleplay (10), common-sense (10), fermi (10), counterfactual (10), coding (7), math (3) and writing (10). Figure~\ref{fig:cap} presents the average scores achieved by BayLing-13B and GPT-3.5-turbo across these 9 categories. BayLing-13B demonstrates comparable performance to GPT-3.5-turbo on generic, knowledge, common-sense and writing tasks. In particular, the exceptional performance of BayLing-13B across Chinese generic, knowledge, common-sense, and writing tasks provides compelling evidence that BayLing effectively transfers LLaMA's English knowledge and capabilities to Chinese, and also achieves a commendable level of Chinese generative ability. However, on tasks that require logical reasoning, such as math, coding and fermi tasks, BayLing-13B still exhibits a notable gap compared to GPT-3.5-turbo. Additionally, since LLaMA is pre-trained on an English-dominant corpus, further improvements are necessary for BayLing-13B to excel in Chinese counterfactual tasks. Overall, Table~\ref{table:total} reports the total score obtained by BayLing-13B compared to GPT-3.5-turbo in 80 cases, where BayLing-13B achieves 89\% performance of GPT-3.5-turbo. Considering the development cost of BayLing and GPT-3.5-turbo, we believe that BayLing demonstrates promising performance in general tasks and shows significant potential. 

\subsection{Standardized Tests}
Enabling LLMs to incorporate knowledge is a focal point of related research. Foundation LLMs have been infused with a large amount of textual knowledge through pre-training on the large-scale corpus. However, this knowledge is typically limited to the language in which the foundation LLM was trained. For instance, the knowledge acquired by LLaMA from the English-dominant corpus is predominantly stored in English \cite{touvron2023llama}, which allows English tasks to benefit from such knowledge but poses a challenge for Chinese tasks to fully leverage this knowledge. Introducing an additional pre-training stage specifically for Chinese knowledge would be inefficient, as Chinese and English knowledge share the same essence and differ only in language.

In our work, BayLing presents an efficient solution to enhance the Chinese ability of LLMs based on knowledge transfer. By employing instruction tuning on interactive translation, BayLing successfully aligns multiple languages within LLMs, which paves the way for transferring the knowledge stored in English form to other languages. To verify the benefit of enhanced language alignment for knowledge transfer across languages, we evaluate the knowledge implicit in BayLing through the Chinese and English standardized tests that rely on professional knowledge.

\subsubsection{Experimental Setup}
We use \texttt{AGIEval} \cite{zhong2023agieval} as benchmarks for standardized tests, including Chinese and English tests. The Chinese tests consist of GaoKao (i.e., Nationwide Unified Examination for Admissions to General Universities and Colleges), including nine subjects of Chinese, English, mathqa, physics, chemistry, biology, history, geography and mathcloze. The English tests consist of SAT, LSAT, Civil Service Examination, GRE and GMAT. The tests consist of Q-A and cloze questions, and we take the percentage of correct answers as the final score of each subject. For all LLMs, we adopt a zero-shot setting, i.e., directly give the questions to the LLMs and ask them to generate the answers.

\begin{table}[t]
\centering
\caption{Test scores on the Chinese Gaokao standardized tests. BayLing-13B achieved an average score of 32.13 across nine subjects.}
\label{table:gaokao}
\scriptsize
\begin{tabular}{lC{0.5cm}C{0.77cm}C{0.77cm}C{0.77cm}C{0.77cm}C{0.77cm}C{0.77cm}C{0.77cm}C{0.77cm}C{0.77cm}} \toprule
\multirow{2}{*}{\textbf{Systems}} & \multirow{2}{*}{\textbf{Avg.}}&\multicolumn{9}{c}{\textbf{GaoKao} (\%)}                                                                                                                                                          \\\cmidrule(lr){3-11}
                                  &  &\textbf{chinese} & \textbf{english} & \textbf{mathqa} & \textbf{physics} & $\!\!$\textbf{chemistry} & \textbf{biology} & \textbf{history} & $\!\!\!$\textbf{geography} & $\!\!\!$\textbf{mathcloze} \\ \midrule
\textbf{GPT-3.5-turbo}            & 43.87         & 42.68            & 86.27            & 30.48           & 21.00            & 44.44              & 46.19            & 59.57            & 63.32              & 0.85               \\
\textbf{BayLing -13B}             & 32.13         & 29.27            & 69.28            & 29.34           & 21.50            & 36.71              & 30.00            & 34.04            & 38.19              & 0.85               \\
\textbf{BayLing-7B}               & 28.20         & 27.64            & 55.56            & 26.78           & 24.50            & 29.95              & 29.05            & 33.19            & 27.14              & 0.00               \\
\textbf{ChatGLM-6B}               & 31.83         & 31.71            & 52.29            & 26.50           & 16.00            & 27.54              & 28.10            & 54.04            & 47.74              & 2.54               \\
\textbf{Vicuna-13B}               & 29.36         & 21.14            & 71.24            & 21.94            & 23.00            & 31.88              & 27.14            & 33.19            & 34.67              & 0.00               \\
\textbf{Alpaca-7B}                & 20.03         & 24.80            & 36.27            & 17.95           & 6.00             & 20.77              & 20.95            & 24.68            & 27.14              & 1.69           \\\bottomrule   
\end{tabular}
\end{table}

\begin{table}[t]
\centering
\caption{Test scores on the English standardized tests. BayLing-13B achieved an average score of 35.31 points across nine subjects.}
\label{table:English_test}
\scriptsize
\begin{tabular} {lC{0.5cm}C{0.77cm}C{0.77cm}C{0.77cm}C{0.77cm}C{0.77cm}C{0.77cm}C{0.77cm}C{0.77cm}C{0.77cm}}\toprule
\multirow{2}{*}{\textbf{Systems}} & \multirow{2}{*}{\textbf{Avg.}} & \multicolumn{3}{c}{\textbf{SAT}}                                                                                                                                                                  & \textbf{\begin{tabular}[c]{@{}c@{}}GRE/\\ GMAT\end{tabular}} & \multicolumn{3}{c}{\textbf{\begin{tabular}[c]{@{}c@{}}LSAT\\ Analytics / Logic / Reading\end{tabular}}}                                                                                 & \multicolumn{2}{c}{\textbf{\begin{tabular}[c]{@{}c@{}}Civil Service\\ Examination\end{tabular}}}                              \\ \cmidrule(lr){3-5}\cmidrule(lr){6-6}\cmidrule(lr){7-9}\cmidrule(lr){10-11}
                                  &                                & \textbf{\begin{tabular}[c]{@{}c@{}}sat\\ -math\end{tabular}} & \textbf{\begin{tabular}[c]{@{}c@{}}sat\\ -en\end{tabular}} & \textbf{\begin{tabular}[c]{@{}c@{}}$\!\!\!$sat-en w/o\\ $\!\!\!$passage\end{tabular}} & \textbf{\begin{tabular}[c]{@{}c@{}}aqua\\ -rat\end{tabular}} & \textbf{\begin{tabular}[c]{@{}c@{}}lsat\\ -ar\end{tabular}} & \textbf{\begin{tabular}[c]{@{}c@{}}lsat\\ -lr\end{tabular}} & \textbf{\begin{tabular}[c]{@{}c@{}}lsat\\ -rc\end{tabular}} & \textbf{\begin{tabular}[c]{@{}c@{}}logiqa\\ -en\end{tabular}} & \textbf{\begin{tabular}[c]{@{}c@{}}logiqa\\ -zh\end{tabular}} \\\midrule
\textbf{GPT-3.5-turbo}            & 49.30                          & 42.27                                                        & 82.04                                                      & 55.83                                                                 & 30.31                                                        & 28.70                                                       & 54.51                                                       & 66.17                                                       & 42.70                                                         & 41.17                                                         \\
\textbf{BayLing -13B}             & 35.31                          & 27.27                                                        & 55.34                                                      & 38.35                                                                 & 22.83                                                        & 22.61                                                       & 38.04                                                       & 42.38                                                       & 35.64                                                         & 31.80                                                         \\
\textbf{BayLing-7B}               & 28.60                          & 25.45                                                        & 42.72                                                      & 29.61                                                                 & 21.26                                                        & 19.13                                                       & 26.86                                                       & 33.83                                                       & 29.95                                                         & 23.81                                                         \\
\textbf{ChatGLM-6B}               & 32.79                          & 27.73                                                        & 56.31                                                      & 37.86                                                                 & 16.54                                                        & 19.57                                                       & 38.04                                                       & 33.09                                                       & 33.18                                                         & 30.57                                                         \\
\textbf{Vicuna-13B}               & 35.97                          & 27.73                                                        & 62.14                                                      & 36.89                                                                 & 20.47                                                        & 20.43                                                       & 41.18                                                       & 45.72                                                       & 33.18                                                         & 28.88                                                         \\
\textbf{Alpaca-7B}                & 24.03                          & 21.36                                                        & 28.16                                                      & 29.13                                                                 & 18.11                                                        & 19.13                                                       & 22.35                                                       & 26.02                                                       & 27.96                                                         & 21.51                                                      \\\bottomrule
\end{tabular}
\end{table}

\subsubsection{Results from Automatic Evaluation}
Tables~\ref{table:gaokao} and \ref{table:English_test} present the performance of LLMs on the Chinese GaoKao and English examinations. BayLing-13B achieves an average score of 32.13 points and 35.31 points on the Chinese GaoKao and English examinations, respectively. 

\textbf{BayLing-13B v.s. Vicuna-13B}\quad BayLing-13B and Vicuna-13B are both built upon LLaMA. Compared to Vicuna-13B, which is solely fine-tuned based on conversation data, BayLing-13B demonstrates an average improvement of 3 points in the Chinese GaoKao through instruction tuning on interactive translation instructions. Specifically, in `Chinese' subject which directly relies on Chinese capability, BayLing-13B outperforms Vicuna-13B by 8 points, demonstrating that BayLing significantly enhances the Chinese capabilities of LLaMA. Besides, subjects like physics, chemistry, biology, mathematics, history and geography always rely on domain knowledge, while this knowledge is typically stored in LLaMA in English form. By enhancing language alignment, BayLing-13B successfully transfers the knowledge in LLaMA from English to Chinese, surpassing Vicuna-13B in most Chinese subjects. While enhancing Chinese knowledge via language alignment, BayLing-13B also achieves comparable performance to Vicuna-13B in English tests.

\textbf{BayLing-13B v.s. ChatGLM-6B}\quad Furthermore, We compare BayLing-13B and ChatGLM-6B, which is built upon GLM that is pre-trained on large-scale Chinese corpus, to analyze the characteristics of knowledge transfer brought by BayLing. BayLing-13B outperforms ChatGLM-6B in subjects such as physics, chemistry and biology, where the domain knowledge is language-independent. This demonstrates that BayLing successfully transfers knowledge in LLaMA from English to Chinese, achieving similar and even better effects as ChatGLM, which directly injects Chinese knowledge into Foundation GLM via the Chinese pre-training. On the contrary, in subjects such as history and geography, which may be closely related to language (e.g., GaoKao may rely on some knowledge about ancient Chinese history and Chinese geography), BayLing-13B's score is slightly lower than ChatGLM-6B. This is because BayLing-13B can transfer the knowledge stored in the foundation LLMs to other languages, but these specific pieces of knowledge may not have initially existed in LLaMA and thus BayLing cannot transfer knowledge out of thin air.

Overall, BayLing is equipped with a high-quality language alignment by instruction tuning on interactive translation. As a by-product, this language alignment enables BayLing to effectively transfer the English capabilities and English knowledge of foundation model into Chinese, achieving enhanced Chinese proficiency and a wealth of Chinese knowledge. BayLing offers a feasible solution for transferring and leveraging the capabilities of foundation LLMs across languages.

\section{Conclusion}

In this study, we develop BayLing, an instruction-following LLM endowed with enhanced language alignment, instruction following and multi-turn interactive capability. Through the comprehensive assessments across various tasks such as multilingual translation, interactive translation, general tasks and standardized tests, BayLing showcases exceptional performance compared to existing open-source LLMs. The performance that closely approaches the state-of-the-art GPT-3.5-turbo highlights the substantial potential of BayLing and the proposed method for instruction tuning.

Moving forward, the contribution of BayLing can be reflected in the following aspects:
\begin{itemize}[itemsep=0pt,topsep=0pt,leftmargin=12pt]
\setlength{\itemsep}{0pt}
\setlength{\parsep}{0pt}
\setlength{\parskip}{0pt}
    \item BayLing offers a viable solution for transferring the proficiency and knowledge of foundation models acquired from the English corpus to other languages.
    \item BayLing showcases the feasibility of simultaneously enhancing multiple capabilities of LLMs by incorporating composite task in instruction tuning.
    \item BayLing can serve as a powerful translation baseline for future research on machine translation as BayLing achieves state-of-the-art translation performance among existing open-source LLMs.
    \item With the integration of powerful language alignment and multi-turn interactive capabilities, BayLing can serve as a good candidate to build upon for the development of follow-up LLMs.
\end{itemize}

\section*{Acknowledgements}

We would like to express our gratitude to all those who have contributed to BayLing. We extend special thanks to Ms. Xiaohong Wang for her valuable comments and suggestions on the use of InforSuperBahn MLOps, and for her organizational and resource support in providing computing resources and showcasing BayLing. We also acknowledge Xiaodong Liu for his pivotal role in the construction of the distributed system and overall coordination of the demo deployment. Furthermore, we appreciate the contribution of the development team from the Nanjing Institute of InforSuperBahn in maintaining the computing resources and creating the display interface for BayLing's webpage and demo.

\bibliographystyle{unsrtnat}
\bibliography{customs}

\begin{thebibliography}{32}
\providecommand{\natexlab}[1]{#1}
\providecommand{\url}[1]{\texttt{#1}}
\expandafter\ifx\csname urlstyle\endcsname\relax
  \providecommand{\doi}[1]{doi: #1}\else
  \providecommand{\doi}{doi: \begingroup \urlstyle{rm}\Url}\fi

\bibitem[Brown et~al.(2020)Brown, Mann, Ryder, Subbiah, Kaplan, Dhariwal,
  Neelakantan, Shyam, Sastry, Askell, Agarwal, Herbert-Voss, Krueger, Henighan,
  Child, Ramesh, Ziegler, Wu, Winter, Hesse, Chen, Sigler, Litwin, Gray, Chess,
  Clark, Berner, McCandlish, Radford, Sutskever, and Amodei]{brown2020language}
Tom~B. Brown, Benjamin Mann, Nick Ryder, Melanie Subbiah, Jared Kaplan,
  Prafulla Dhariwal, Arvind Neelakantan, Pranav Shyam, Girish Sastry, Amanda
  Askell, Sandhini Agarwal, Ariel Herbert-Voss, Gretchen Krueger, Tom Henighan,
  Rewon Child, Aditya Ramesh, Daniel~M. Ziegler, Jeffrey Wu, Clemens Winter,
  Christopher Hesse, Mark Chen, Eric Sigler, Mateusz Litwin, Scott Gray,
  Benjamin Chess, Jack Clark, Christopher Berner, Sam McCandlish, Alec Radford,
  Ilya Sutskever, and Dario Amodei.
\newblock Language models are few-shot learners, 2020.

\bibitem[Chowdhery et~al.(2022)Chowdhery, Narang, Devlin, Bosma, Mishra,
  Roberts, Barham, Chung, Sutton, Gehrmann, Schuh, Shi, Tsvyashchenko, Maynez,
  Rao, Barnes, Tay, Shazeer, Prabhakaran, Reif, Du, Hutchinson, Pope, Bradbury,
  Austin, Isard, Gur-Ari, Yin, Duke, Levskaya, Ghemawat, Dev, Michalewski,
  Garcia, Misra, Robinson, Fedus, Zhou, Ippolito, Luan, Lim, Zoph, Spiridonov,
  Sepassi, Dohan, Agrawal, Omernick, Dai, Pillai, Pellat, Lewkowycz, Moreira,
  Child, Polozov, Lee, Zhou, Wang, Saeta, Diaz, Firat, Catasta, Wei,
  Meier-Hellstern, Eck, Dean, Petrov, and Fiedel]{chowdhery2022palm}
Aakanksha Chowdhery, Sharan Narang, Jacob Devlin, Maarten Bosma, Gaurav Mishra,
  Adam Roberts, Paul Barham, Hyung~Won Chung, Charles Sutton, Sebastian
  Gehrmann, Parker Schuh, Kensen Shi, Sasha Tsvyashchenko, Joshua Maynez,
  Abhishek Rao, Parker Barnes, Yi~Tay, Noam Shazeer, Vinodkumar Prabhakaran,
  Emily Reif, Nan Du, Ben Hutchinson, Reiner Pope, James Bradbury, Jacob
  Austin, Michael Isard, Guy Gur-Ari, Pengcheng Yin, Toju Duke, Anselm
  Levskaya, Sanjay Ghemawat, Sunipa Dev, Henryk Michalewski, Xavier Garcia,
  Vedant Misra, Kevin Robinson, Liam Fedus, Denny Zhou, Daphne Ippolito, David
  Luan, Hyeontaek Lim, Barret Zoph, Alexander Spiridonov, Ryan Sepassi, David
  Dohan, Shivani Agrawal, Mark Omernick, Andrew~M. Dai,
  Thanumalayan~Sankaranarayana Pillai, Marie Pellat, Aitor Lewkowycz, Erica
  Moreira, Rewon Child, Oleksandr Polozov, Katherine Lee, Zongwei Zhou, Xuezhi
  Wang, Brennan Saeta, Mark Diaz, Orhan Firat, Michele Catasta, Jason Wei,
  Kathy Meier-Hellstern, Douglas Eck, Jeff Dean, Slav Petrov, and Noah Fiedel.
\newblock Palm: Scaling language modeling with pathways, 2022.

\bibitem[Zhang et~al.(2022)Zhang, Roller, Goyal, Artetxe, Chen, Chen, Dewan,
  Diab, Li, Lin, Mihaylov, Ott, Shleifer, Shuster, Simig, Koura, Sridhar, Wang,
  and Zettlemoyer]{zhang2022opt}
Susan Zhang, Stephen Roller, Naman Goyal, Mikel Artetxe, Moya Chen, Shuohui
  Chen, Christopher Dewan, Mona Diab, Xian Li, Xi~Victoria Lin, Todor Mihaylov,
  Myle Ott, Sam Shleifer, Kurt Shuster, Daniel Simig, Punit~Singh Koura, Anjali
  Sridhar, Tianlu Wang, and Luke Zettlemoyer.
\newblock Opt: Open pre-trained transformer language models, 2022.

\bibitem[Du et~al.(2022)Du, Qian, Liu, Ding, Qiu, Yang, and
  Tang]{du-etal-2022-glm}
Zhengxiao Du, Yujie Qian, Xiao Liu, Ming Ding, Jiezhong Qiu, Zhilin Yang, and
  Jie Tang.
\newblock {GLM}: General language model pretraining with autoregressive blank
  infilling.
\newblock In \emph{Proceedings of the 60th Annual Meeting of the Association
  for Computational Linguistics (Volume 1: Long Papers)}, pages 320--335,
  Dublin, Ireland, May 2022. Association for Computational Linguistics.
\newblock \doi{10.18653/v1/2022.acl-long.26}.
\newblock URL \url{https://aclanthology.org/2022.acl-long.26}.

\bibitem[Workshop et~al.(2023)Workshop, :, Scao, Fan, Akiki, Pavlick, Ilić,
  Hesslow, Castagné, Luccioni, Yvon, Gallé, Tow, Rush, Biderman, Webson,
  Ammanamanchi, Wang, Sagot, Muennighoff, del Moral, Ruwase, Bawden, Bekman,
  McMillan-Major, Beltagy, Nguyen, Saulnier, Tan, Suarez, Sanh, Laurençon,
  Jernite, Launay, Mitchell, Raffel, Gokaslan, Simhi, Soroa, Aji, Alfassy,
  Rogers, Nitzav, Xu, Mou, Emezue, Klamm, Leong, van Strien, Adelani, Radev,
  Ponferrada, Levkovizh, Kim, Natan, Toni, Dupont, Kruszewski, Pistilli,
  Elsahar, Benyamina, Tran, Yu, Abdulmumin, Johnson, Gonzalez-Dios, de~la Rosa,
  Chim, Dodge, Zhu, Chang, Frohberg, Tobing, Bhattacharjee, Almubarak, Chen,
  Lo, Werra, Weber, Phan, allal, Tanguy, Dey, Muñoz, Masoud, Grandury,
  Šaško, Huang, Coavoux, Singh, Jiang, Vu, Jauhar, Ghaleb, Subramani,
  Kassner, Khamis, Nguyen, Espejel, de~Gibert, Villegas, Henderson, Colombo,
  Amuok, Lhoest, Harliman, Bommasani, López, Ribeiro, Osei, Pyysalo, Nagel,
  Bose, Muhammad, Sharma, Longpre, Nikpoor, Silberberg, Pai, Zink, Torrent,
  Schick, Thrush, Danchev, Nikoulina, Laippala, Lepercq, Prabhu, Alyafeai,
  Talat, Raja, Heinzerling, Si, Taşar, Salesky, Mielke, Lee, Sharma, Santilli,
  Chaffin, Stiegler, Datta, Szczechla, Chhablani, Wang, Pandey, Strobelt,
  Fries, Rozen, Gao, Sutawika, Bari, Al-shaibani, Manica, Nayak, Teehan,
  Albanie, Shen, Ben-David, Bach, Kim, Bers, Fevry, Neeraj, Thakker, Raunak,
  Tang, Yong, Sun, Brody, Uri, Tojarieh, Roberts, Chung, Tae, Phang, Press, Li,
  Narayanan, Bourfoune, Casper, Rasley, Ryabinin, Mishra, Zhang, Shoeybi,
  Peyrounette, Patry, Tazi, Sanseviero, von Platen, Cornette, Lavallée,
  Lacroix, Rajbhandari, Gandhi, Smith, Requena, Patil, Dettmers, Baruwa, Singh,
  Cheveleva, Ligozat, Subramonian, Névéol, Lovering, Garrette, Tunuguntla,
  Reiter, Taktasheva, Voloshina, Bogdanov, Winata, Schoelkopf, Kalo, Novikova,
  Forde, Clive, Kasai, Kawamura, Hazan, Carpuat, Clinciu, Kim, Cheng, Serikov,
  Antverg, van~der Wal, Zhang, Zhang, Gehrmann, Mirkin, Pais, Shavrina,
  Scialom, Yun, Limisiewicz, Rieser, Protasov, Mikhailov, Pruksachatkun,
  Belinkov, Bamberger, Kasner, Rueda, Pestana, Feizpour, Khan, Faranak, Santos,
  Hevia, Unldreaj, Aghagol, Abdollahi, Tammour, HajiHosseini, Behroozi,
  Ajibade, Saxena, Ferrandis, Contractor, Lansky, David, Kiela, Nguyen, Tan,
  Baylor, Ozoani, Mirza, Ononiwu, Rezanejad, Jones, Bhattacharya, Solaiman,
  Sedenko, Nejadgholi, Passmore, Seltzer, Sanz, Dutra, Samagaio, Elbadri,
  Mieskes, Gerchick, Akinlolu, McKenna, Qiu, Ghauri, Burynok, Abrar, Rajani,
  Elkott, Fahmy, Samuel, An, Kromann, Hao, Alizadeh, Shubber, Wang, Roy,
  Viguier, Le, Oyebade, Le, Yang, Nguyen, Kashyap, Palasciano, Callahan,
  Shukla, Miranda-Escalada, Singh, Beilharz, Wang, Brito, Zhou, Jain, Xu,
  Fourrier, Periñán, Molano, Yu, Manjavacas, Barth, Fuhrimann, Altay, Bayrak,
  Burns, Vrabec, Bello, Dash, Kang, Giorgi, Golde, Posada, Sivaraman,
  Bulchandani, Liu, Shinzato, de~Bykhovetz, Takeuchi, Pàmies, Castillo,
  Nezhurina, Sänger, Samwald, Cullan, Weinberg, Wolf, Mihaljcic, Liu,
  Freidank, Kang, Seelam, Dahlberg, Broad, Muellner, Fung, Haller,
  Chandrasekhar, Eisenberg, Martin, Canalli, Su, Su, Cahyawijaya, Garda,
  Deshmukh, Mishra, Kiblawi, Ott, Sang-aroonsiri, Kumar, Schweter, Bharati,
  Laud, Gigant, Kainuma, Kusa, Labrak, Bajaj, Venkatraman, Xu, Xu, Xu, Tan,
  Xie, Ye, Bras, Belkada, and Wolf]{workshop2023bloom}
BigScience Workshop, :, Teven~Le Scao, Angela Fan, Christopher Akiki, Ellie
  Pavlick, Suzana Ilić, Daniel Hesslow, Roman Castagné, Alexandra~Sasha
  Luccioni, François Yvon, Matthias Gallé, Jonathan Tow, Alexander~M. Rush,
  Stella Biderman, Albert Webson, Pawan~Sasanka Ammanamanchi, Thomas Wang,
  Benoît Sagot, Niklas Muennighoff, Albert~Villanova del Moral, Olatunji
  Ruwase, Rachel Bawden, Stas Bekman, Angelina McMillan-Major, Iz~Beltagy, Huu
  Nguyen, Lucile Saulnier, Samson Tan, Pedro~Ortiz Suarez, Victor Sanh, Hugo
  Laurençon, Yacine Jernite, Julien Launay, Margaret Mitchell, Colin Raffel,
  Aaron Gokaslan, Adi Simhi, Aitor Soroa, Alham~Fikri Aji, Amit Alfassy, Anna
  Rogers, Ariel~Kreisberg Nitzav, Canwen Xu, Chenghao Mou, Chris Emezue,
  Christopher Klamm, Colin Leong, Daniel van Strien, David~Ifeoluwa Adelani,
  Dragomir Radev, Eduardo~González Ponferrada, Efrat Levkovizh, Ethan Kim,
  Eyal~Bar Natan, Francesco~De Toni, Gérard Dupont, Germán Kruszewski, Giada
  Pistilli, Hady Elsahar, Hamza Benyamina, Hieu Tran, Ian Yu, Idris Abdulmumin,
  Isaac Johnson, Itziar Gonzalez-Dios, Javier de~la Rosa, Jenny Chim, Jesse
  Dodge, Jian Zhu, Jonathan Chang, Jörg Frohberg, Joseph Tobing, Joydeep
  Bhattacharjee, Khalid Almubarak, Kimbo Chen, Kyle Lo, Leandro~Von Werra, Leon
  Weber, Long Phan, Loubna~Ben allal, Ludovic Tanguy, Manan Dey, Manuel~Romero
  Muñoz, Maraim Masoud, María Grandury, Mario Šaško, Max Huang, Maximin
  Coavoux, Mayank Singh, Mike Tian-Jian Jiang, Minh~Chien Vu, Mohammad~A.
  Jauhar, Mustafa Ghaleb, Nishant Subramani, Nora Kassner, Nurulaqilla Khamis,
  Olivier Nguyen, Omar Espejel, Ona de~Gibert, Paulo Villegas, Peter Henderson,
  Pierre Colombo, Priscilla Amuok, Quentin Lhoest, Rheza Harliman, Rishi
  Bommasani, Roberto~Luis López, Rui Ribeiro, Salomey Osei, Sampo Pyysalo,
  Sebastian Nagel, Shamik Bose, Shamsuddeen~Hassan Muhammad, Shanya Sharma,
  Shayne Longpre, Somaieh Nikpoor, Stanislav Silberberg, Suhas Pai, Sydney
  Zink, Tiago~Timponi Torrent, Timo Schick, Tristan Thrush, Valentin Danchev,
  Vassilina Nikoulina, Veronika Laippala, Violette Lepercq, Vrinda Prabhu, Zaid
  Alyafeai, Zeerak Talat, Arun Raja, Benjamin Heinzerling, Chenglei Si,
  Davut~Emre Taşar, Elizabeth Salesky, Sabrina~J. Mielke, Wilson~Y. Lee,
  Abheesht Sharma, Andrea Santilli, Antoine Chaffin, Arnaud Stiegler, Debajyoti
  Datta, Eliza Szczechla, Gunjan Chhablani, Han Wang, Harshit Pandey, Hendrik
  Strobelt, Jason~Alan Fries, Jos Rozen, Leo Gao, Lintang Sutawika, M~Saiful
  Bari, Maged~S. Al-shaibani, Matteo Manica, Nihal Nayak, Ryan Teehan, Samuel
  Albanie, Sheng Shen, Srulik Ben-David, Stephen~H. Bach, Taewoon Kim, Tali
  Bers, Thibault Fevry, Trishala Neeraj, Urmish Thakker, Vikas Raunak, Xiangru
  Tang, Zheng-Xin Yong, Zhiqing Sun, Shaked Brody, Yallow Uri, Hadar Tojarieh,
  Adam Roberts, Hyung~Won Chung, Jaesung Tae, Jason Phang, Ofir Press, Conglong
  Li, Deepak Narayanan, Hatim Bourfoune, Jared Casper, Jeff Rasley, Max
  Ryabinin, Mayank Mishra, Minjia Zhang, Mohammad Shoeybi, Myriam Peyrounette,
  Nicolas Patry, Nouamane Tazi, Omar Sanseviero, Patrick von Platen, Pierre
  Cornette, Pierre~François Lavallée, Rémi Lacroix, Samyam Rajbhandari,
  Sanchit Gandhi, Shaden Smith, Stéphane Requena, Suraj Patil, Tim Dettmers,
  Ahmed Baruwa, Amanpreet Singh, Anastasia Cheveleva, Anne-Laure Ligozat, Arjun
  Subramonian, Aurélie Névéol, Charles Lovering, Dan Garrette, Deepak
  Tunuguntla, Ehud Reiter, Ekaterina Taktasheva, Ekaterina Voloshina, Eli
  Bogdanov, Genta~Indra Winata, Hailey Schoelkopf, Jan-Christoph Kalo,
  Jekaterina Novikova, Jessica~Zosa Forde, Jordan Clive, Jungo Kasai, Ken
  Kawamura, Liam Hazan, Marine Carpuat, Miruna Clinciu, Najoung Kim, Newton
  Cheng, Oleg Serikov, Omer Antverg, Oskar van~der Wal, Rui Zhang, Ruochen
  Zhang, Sebastian Gehrmann, Shachar Mirkin, Shani Pais, Tatiana Shavrina,
  Thomas Scialom, Tian Yun, Tomasz Limisiewicz, Verena Rieser, Vitaly Protasov,
  Vladislav Mikhailov, Yada Pruksachatkun, Yonatan Belinkov, Zachary Bamberger,
  Zdeněk Kasner, Alice Rueda, Amanda Pestana, Amir Feizpour, Ammar Khan, Amy
  Faranak, Ana Santos, Anthony Hevia, Antigona Unldreaj, Arash Aghagol, Arezoo
  Abdollahi, Aycha Tammour, Azadeh HajiHosseini, Bahareh Behroozi, Benjamin
  Ajibade, Bharat Saxena, Carlos~Muñoz Ferrandis, Danish Contractor, David
  Lansky, Davis David, Douwe Kiela, Duong~A. Nguyen, Edward Tan, Emi Baylor,
  Ezinwanne Ozoani, Fatima Mirza, Frankline Ononiwu, Habib Rezanejad, Hessie
  Jones, Indrani Bhattacharya, Irene Solaiman, Irina Sedenko, Isar Nejadgholi,
  Jesse Passmore, Josh Seltzer, Julio~Bonis Sanz, Livia Dutra, Mairon Samagaio,
  Maraim Elbadri, Margot Mieskes, Marissa Gerchick, Martha Akinlolu, Michael
  McKenna, Mike Qiu, Muhammed Ghauri, Mykola Burynok, Nafis Abrar, Nazneen
  Rajani, Nour Elkott, Nour Fahmy, Olanrewaju Samuel, Ran An, Rasmus Kromann,
  Ryan Hao, Samira Alizadeh, Sarmad Shubber, Silas Wang, Sourav Roy, Sylvain
  Viguier, Thanh Le, Tobi Oyebade, Trieu Le, Yoyo Yang, Zach Nguyen,
  Abhinav~Ramesh Kashyap, Alfredo Palasciano, Alison Callahan, Anima Shukla,
  Antonio Miranda-Escalada, Ayush Singh, Benjamin Beilharz, Bo~Wang, Caio
  Brito, Chenxi Zhou, Chirag Jain, Chuxin Xu, Clémentine Fourrier,
  Daniel~León Periñán, Daniel Molano, Dian Yu, Enrique Manjavacas, Fabio
  Barth, Florian Fuhrimann, Gabriel Altay, Giyaseddin Bayrak, Gully Burns,
  Helena~U. Vrabec, Imane Bello, Ishani Dash, Jihyun Kang, John Giorgi, Jonas
  Golde, Jose~David Posada, Karthik~Rangasai Sivaraman, Lokesh Bulchandani,
  Lu~Liu, Luisa Shinzato, Madeleine~Hahn de~Bykhovetz, Maiko Takeuchi, Marc
  Pàmies, Maria~A Castillo, Marianna Nezhurina, Mario Sänger, Matthias
  Samwald, Michael Cullan, Michael Weinberg, Michiel~De Wolf, Mina Mihaljcic,
  Minna Liu, Moritz Freidank, Myungsun Kang, Natasha Seelam, Nathan Dahlberg,
  Nicholas~Michio Broad, Nikolaus Muellner, Pascale Fung, Patrick Haller, Ramya
  Chandrasekhar, Renata Eisenberg, Robert Martin, Rodrigo Canalli, Rosaline Su,
  Ruisi Su, Samuel Cahyawijaya, Samuele Garda, Shlok~S Deshmukh, Shubhanshu
  Mishra, Sid Kiblawi, Simon Ott, Sinee Sang-aroonsiri, Srishti Kumar, Stefan
  Schweter, Sushil Bharati, Tanmay Laud, Théo Gigant, Tomoya Kainuma, Wojciech
  Kusa, Yanis Labrak, Yash~Shailesh Bajaj, Yash Venkatraman, Yifan Xu, Yingxin
  Xu, Yu~Xu, Zhe Tan, Zhongli Xie, Zifan Ye, Mathilde Bras, Younes Belkada, and
  Thomas Wolf.
\newblock Bloom: A 176b-parameter open-access multilingual language model,
  2023.

\bibitem[Touvron et~al.(2023)Touvron, Lavril, Izacard, Martinet, Lachaux,
  Lacroix, Rozière, Goyal, Hambro, Azhar, Rodriguez, Joulin, Grave, and
  Lample]{touvron2023llama}
Hugo Touvron, Thibaut Lavril, Gautier Izacard, Xavier Martinet, Marie-Anne
  Lachaux, Timothée Lacroix, Baptiste Rozière, Naman Goyal, Eric Hambro,
  Faisal Azhar, Aurelien Rodriguez, Armand Joulin, Edouard Grave, and Guillaume
  Lample.
\newblock Llama: Open and efficient foundation language models, 2023.

\bibitem[Christiano et~al.(2017)Christiano, Leike, Brown, Martic, Legg, and
  Amodei]{NIPS2017_d5e2c0ad}
Paul~F Christiano, Jan Leike, Tom Brown, Miljan Martic, Shane Legg, and Dario
  Amodei.
\newblock Deep reinforcement learning from human preferences.
\newblock In I.~Guyon, U.~Von Luxburg, S.~Bengio, H.~Wallach, R.~Fergus,
  S.~Vishwanathan, and R.~Garnett, editors, \emph{Advances in Neural
  Information Processing Systems}, volume~30. Curran Associates, Inc., 2017.
\newblock URL
  \url{https://proceedings.neurips.cc/paper_files/paper/2017/file/d5e2c0adad503c91f91df240d0cd4e49-Paper.pdf}.

\bibitem[Ouyang et~al.(2022{\natexlab{a}})Ouyang, Wu, Jiang, Almeida,
  Wainwright, Mishkin, Zhang, Agarwal, Slama, Ray, Schulman, Hilton, Kelton,
  Miller, Simens, Askell, Welinder, Christiano, Leike, and
  Lowe]{NEURIPS2022_b1efde53}
Long Ouyang, Jeffrey Wu, Xu~Jiang, Diogo Almeida, Carroll Wainwright, Pamela
  Mishkin, Chong Zhang, Sandhini Agarwal, Katarina Slama, Alex Ray, John
  Schulman, Jacob Hilton, Fraser Kelton, Luke Miller, Maddie Simens, Amanda
  Askell, Peter Welinder, Paul~F Christiano, Jan Leike, and Ryan Lowe.
\newblock Training language models to follow instructions with human feedback.
\newblock In S.~Koyejo, S.~Mohamed, A.~Agarwal, D.~Belgrave, K.~Cho, and A.~Oh,
  editors, \emph{Advances in Neural Information Processing Systems}, volume~35,
  pages 27730--27744. Curran Associates, Inc., 2022{\natexlab{a}}.
\newblock URL
  \url{https://proceedings.neurips.cc/paper_files/paper/2022/file/b1efde53be364a73914f58805a001731-Paper-Conference.pdf}.

\bibitem[Ouyang et~al.(2022{\natexlab{b}})Ouyang, Wu, Jiang, Almeida,
  Wainwright, Mishkin, Zhang, Agarwal, Slama, Ray, Schulman, Hilton, Kelton,
  Miller, Simens, Askell, Welinder, Christiano, Leike, and
  Lowe]{ouyang2022training}
Long Ouyang, Jeff Wu, Xu~Jiang, Diogo Almeida, Carroll~L. Wainwright, Pamela
  Mishkin, Chong Zhang, Sandhini Agarwal, Katarina Slama, Alex Ray, John
  Schulman, Jacob Hilton, Fraser Kelton, Luke Miller, Maddie Simens, Amanda
  Askell, Peter Welinder, Paul Christiano, Jan Leike, and Ryan Lowe.
\newblock Training language models to follow instructions with human feedback,
  2022{\natexlab{b}}.

\bibitem[Mishra et~al.(2022)Mishra, Khashabi, Baral, and
  Hajishirzi]{mishra-etal-2022-cross}
Swaroop Mishra, Daniel Khashabi, Chitta Baral, and Hannaneh Hajishirzi.
\newblock Cross-task generalization via natural language crowdsourcing
  instructions.
\newblock In \emph{Proceedings of the 60th Annual Meeting of the Association
  for Computational Linguistics (Volume 1: Long Papers)}, pages 3470--3487,
  Dublin, Ireland, May 2022. Association for Computational Linguistics.
\newblock \doi{10.18653/v1/2022.acl-long.244}.
\newblock URL \url{https://aclanthology.org/2022.acl-long.244}.

\bibitem[OpenAI(2022)]{chatgpt}
OpenAI.
\newblock Introducing chatgpt, 2022.
\newblock URL \url{https://openai.com/blog/chatgpt}.

\bibitem[OpenAI(2023)]{openai2023gpt4}
OpenAI.
\newblock Gpt-4 technical report, 2023.

\bibitem[Taori et~al.(2023)Taori, Gulrajani, Zhang, Dubois, Li, Guestrin,
  Liang, and Hashimoto]{alpaca}
Rohan Taori, Ishaan Gulrajani, Tianyi Zhang, Yann Dubois, Xuechen Li, Carlos
  Guestrin, Percy Liang, and Tatsunori~B. Hashimoto.
\newblock Stanford alpaca: An instruction-following llama model.
\newblock \url{https://github.com/tatsu-lab/stanford_alpaca}, 2023.

\bibitem[Garcia et~al.(2023)Garcia, Bansal, Cherry, Foster, Krikun, Feng,
  Johnson, and Firat]{garcia2023unreasonable}
Xavier Garcia, Yamini Bansal, Colin Cherry, George Foster, Maxim Krikun,
  Fangxiaoyu Feng, Melvin Johnson, and Orhan Firat.
\newblock The unreasonable effectiveness of few-shot learning for machine
  translation, 2023.

\bibitem[Chiang et~al.(2023)Chiang, Li, Lin, Sheng, Wu, Zhang, Zheng, Zhuang,
  Zhuang, Gonzalez, Stoica, and Xing]{vicuna2023}
Wei-Lin Chiang, Zhuohan Li, Zi~Lin, Ying Sheng, Zhanghao Wu, Hao Zhang, Lianmin
  Zheng, Siyuan Zhuang, Yonghao Zhuang, Joseph~E. Gonzalez, Ion Stoica, and
  Eric~P. Xing.
\newblock Vicuna: An open-source chatbot impressing gpt-4 with 90\%* chatgpt
  quality, March 2023.
\newblock URL \url{https://lmsys.org/blog/2023-03-30-vicuna/}.

\bibitem[Sun et~al.(2023)Sun, Shen, Zhou, Zhang, Chen, Cox, Yang, and
  Gan]{sun2023principledriven}
Zhiqing Sun, Yikang Shen, Qinhong Zhou, Hongxin Zhang, Zhenfang Chen, David
  Cox, Yiming Yang, and Chuang Gan.
\newblock Principle-driven self-alignment of language models from scratch with
  minimal human supervision, 2023.

\bibitem[Köpf et~al.(2023)Köpf, Kilcher, von Rütte, Anagnostidis, Tam,
  Stevens, Barhoum, Duc, Stanley, Nagyfi, ES, Suri, Glushkov, Dantuluri,
  Maguire, Schuhmann, Nguyen, and Mattick]{köpf2023openassistant}
Andreas Köpf, Yannic Kilcher, Dimitri von Rütte, Sotiris Anagnostidis,
  Zhi-Rui Tam, Keith Stevens, Abdullah Barhoum, Nguyen~Minh Duc, Oliver
  Stanley, Richárd Nagyfi, Shahul ES, Sameer Suri, David Glushkov, Arnav
  Dantuluri, Andrew Maguire, Christoph Schuhmann, Huu Nguyen, and Alexander
  Mattick.
\newblock Openassistant conversations -- democratizing large language model
  alignment, 2023.

\bibitem[Zhou et~al.(2023)Zhou, Liu, Xu, Iyer, Sun, Mao, Ma, Efrat, Yu, Yu,
  Zhang, Ghosh, Lewis, Zettlemoyer, and Levy]{zhou2023lima}
Chunting Zhou, Pengfei Liu, Puxin Xu, Srini Iyer, Jiao Sun, Yuning Mao, Xuezhe
  Ma, Avia Efrat, Ping Yu, Lili Yu, Susan Zhang, Gargi Ghosh, Mike Lewis, Luke
  Zettlemoyer, and Omer Levy.
\newblock Lima: Less is more for alignment, 2023.

\bibitem[Rasley et~al.(2020)Rasley, Rajbhandari, Ruwase, and
  He]{10.1145/3394486.3406703}
Jeff Rasley, Samyam Rajbhandari, Olatunji Ruwase, and Yuxiong He.
\newblock Deepspeed: System optimizations enable training deep learning models
  with over 100 billion parameters.
\newblock In \emph{Proceedings of the 26th ACM SIGKDD International Conference
  on Knowledge Discovery \&amp; Data Mining}, KDD '20, page 3505–3506, New
  York, NY, USA, 2020. Association for Computing Machinery.
\newblock ISBN 9781450379984.
\newblock \doi{10.1145/3394486.3406703}.
\newblock URL \url{https://doi.org/10.1145/3394486.3406703}.

\bibitem[Chen et~al.(2016)Chen, Xu, Zhang, and Guestrin]{chen2016training}
Tianqi Chen, Bing Xu, Chiyuan Zhang, and Carlos Guestrin.
\newblock Training deep nets with sublinear memory cost, 2016.

\bibitem[Johnson et~al.(2017)Johnson, Schuster, Le, Krikun, Wu, Chen, Thorat,
  Vi{\'e}gas, Wattenberg, Corrado, Hughes, and Dean]{johnson-etal-2017-googles}
Melvin Johnson, Mike Schuster, Quoc~V. Le, Maxim Krikun, Yonghui Wu, Zhifeng
  Chen, Nikhil Thorat, Fernanda Vi{\'e}gas, Martin Wattenberg, Greg Corrado,
  Macduff Hughes, and Jeffrey Dean.
\newblock {G}oogle{'}s multilingual neural machine translation system: Enabling
  zero-shot translation.
\newblock \emph{Transactions of the Association for Computational Linguistics},
  5:\penalty0 339--351, 2017.
\newblock \doi{10.1162/tacl_a_00065}.
\newblock URL \url{https://aclanthology.org/Q17-1024}.

\bibitem[Team et~al.(2022)Team, Costa-jussà, Cross, Çelebi, Elbayad,
  Heafield, Heffernan, Kalbassi, Lam, Licht, Maillard, Sun, Wang, Wenzek,
  Youngblood, Akula, Barrault, Gonzalez, Hansanti, Hoffman, Jarrett, Sadagopan,
  Rowe, Spruit, Tran, Andrews, Ayan, Bhosale, Edunov, Fan, Gao, Goswami,
  Guzmán, Koehn, Mourachko, Ropers, Saleem, Schwenk, and
  Wang]{nllbteam2022language}
NLLB Team, Marta~R. Costa-jussà, James Cross, Onur Çelebi, Maha Elbayad,
  Kenneth Heafield, Kevin Heffernan, Elahe Kalbassi, Janice Lam, Daniel Licht,
  Jean Maillard, Anna Sun, Skyler Wang, Guillaume Wenzek, Al~Youngblood, Bapi
  Akula, Loic Barrault, Gabriel~Mejia Gonzalez, Prangthip Hansanti, John
  Hoffman, Semarley Jarrett, Kaushik~Ram Sadagopan, Dirk Rowe, Shannon Spruit,
  Chau Tran, Pierre Andrews, Necip~Fazil Ayan, Shruti Bhosale, Sergey Edunov,
  Angela Fan, Cynthia Gao, Vedanuj Goswami, Francisco Guzmán, Philipp Koehn,
  Alexandre Mourachko, Christophe Ropers, Safiyyah Saleem, Holger Schwenk, and
  Jeff Wang.
\newblock No language left behind: Scaling human-centered machine translation,
  2022.

\bibitem[Jiao et~al.(2023)Jiao, tse Huang, Wang, Wang, Shi, and
  Tu]{jiao2023parrot}
Wenxiang Jiao, Jen tse Huang, Wenxuan Wang, Xing Wang, Shuming Shi, and
  Zhaopeng Tu.
\newblock Parrot: Translating during chat using large language models, 2023.

\bibitem[Post(2018)]{post-2018-call}
Matt Post.
\newblock A call for clarity in reporting {BLEU} scores.
\newblock In \emph{Proceedings of the Third Conference on Machine Translation:
  Research Papers}, pages 186--191, Brussels, Belgium, October 2018.
  Association for Computational Linguistics.
\newblock \doi{10.18653/v1/W18-6319}.
\newblock URL \url{https://aclanthology.org/W18-6319}.

\bibitem[Rei et~al.(2022)Rei, C.~de Souza, Alves, Zerva, Farinha, Glushkova,
  Lavie, Coheur, and Martins]{rei-etal-2022-comet}
Ricardo Rei, Jos{\'e}~G. C.~de Souza, Duarte Alves, Chrysoula Zerva, Ana~C
  Farinha, Taisiya Glushkova, Alon Lavie, Luisa Coheur, and Andr{\'e} F.~T.
  Martins.
\newblock {COMET}-22: Unbabel-{IST} 2022 submission for the metrics shared
  task.
\newblock In \emph{Proceedings of the Seventh Conference on Machine Translation
  (WMT)}, pages 578--585, Abu Dhabi, United Arab Emirates (Hybrid), December
  2022. Association for Computational Linguistics.
\newblock URL \url{https://aclanthology.org/2022.wmt-1.52}.

\bibitem[Zhu et~al.(2023)Zhu, Liu, Dong, Xu, Huang, Kong, Chen, and
  Li]{zhu2023multilingual}
Wenhao Zhu, Hongyi Liu, Qingxiu Dong, Jingjing Xu, Shujian Huang, Lingpeng
  Kong, Jiajun Chen, and Lei Li.
\newblock Multilingual machine translation with large language models:
  Empirical results and analysis, 2023.

\bibitem[Hu et~al.(2019)Hu, Khayrallah, Culkin, Xia, Chen, Post, and
  Van~Durme]{hu-etal-2019-improved}
J.~Edward Hu, Huda Khayrallah, Ryan Culkin, Patrick Xia, Tongfei Chen, Matt
  Post, and Benjamin Van~Durme.
\newblock Improved lexically constrained decoding for translation and
  monolingual rewriting.
\newblock In \emph{Proceedings of the 2019 Conference of the North {A}merican
  Chapter of the Association for Computational Linguistics: Human Language
  Technologies, Volume 1 (Long and Short Papers)}, pages 839--850, Minneapolis,
  Minnesota, June 2019. Association for Computational Linguistics.
\newblock \doi{10.18653/v1/N19-1090}.
\newblock URL \url{https://www.aclweb.org/anthology/N19-1090}.

\bibitem[Chen et~al.(2020)Chen, Chen, Wang, and Li]{ijcai2020p0496}
Guanhua Chen, Yun Chen, Yong Wang, and Victor~O.K. Li.
\newblock Lexical-constraint-aware neural machine translation via data
  augmentation.
\newblock In Christian Bessiere, editor, \emph{Proceedings of the Twenty-Ninth
  International Joint Conference on Artificial Intelligence, {IJCAI-20}}, pages
  3587--3593. International Joint Conferences on Artificial Intelligence
  Organization, 7 2020.
\newblock \doi{10.24963/ijcai.2020/496}.
\newblock URL \url{https://doi.org/10.24963/ijcai.2020/496}.
\newblock Main track.

\bibitem[Jon et~al.(2021)Jon, Aires, Varis, and Bojar]{jon-etal-2021-end}
Josef Jon, Jo{\~a}o~Paulo Aires, Dusan Varis, and Ond{\v{r}}ej Bojar.
\newblock End-to-end lexically constrained machine translation for
  morphologically rich languages.
\newblock In \emph{Proceedings of the 59th Annual Meeting of the Association
  for Computational Linguistics and the 11th International Joint Conference on
  Natural Language Processing (Volume 1: Long Papers)}, pages 4019--4033,
  Online, August 2021. Association for Computational Linguistics.
\newblock \doi{10.18653/v1/2021.acl-long.311}.
\newblock URL \url{https://aclanthology.org/2021.acl-long.311}.

\bibitem[Liu et~al.(2023)Liu, Iter, Xu, Wang, Xu, and Zhu]{liu2023geval}
Yang Liu, Dan Iter, Yichong Xu, Shuohang Wang, Ruochen Xu, and Chenguang Zhu.
\newblock G-eval: Nlg evaluation using gpt-4 with better human alignment, 2023.

\bibitem[Zhong et~al.(2023)Zhong, Cui, Guo, Liang, Lu, Wang, Saied, Chen, and
  Duan]{zhong2023agieval}
Wanjun Zhong, Ruixiang Cui, Yiduo Guo, Yaobo Liang, Shuai Lu, Yanlin Wang, Amin
  Saied, Weizhu Chen, and Nan Duan.
\newblock Agieval: A human-centric benchmark for evaluating foundation models,
  2023.

\bibitem[Wang et~al.(2023)Wang, Li, Chen, Zhu, Lin, Cao, Liu, Liu, and
  Sui]{wang2023large}
Peiyi Wang, Lei Li, Liang Chen, Dawei Zhu, Binghuai Lin, Yunbo Cao, Qi~Liu,
  Tianyu Liu, and Zhifang Sui.
\newblock Large language models are not fair evaluators, 2023.

\end{thebibliography}

\newpage
\appendix

\section{Translation Benchmarks}
\label{app:benchmark}

We evaluate the translation quality of LLMs on WMT22 benchmarks, consisting of Chinese$\Leftrightarrow$English and Germain$\Leftrightarrow$English multi-domain benchmarks and multilingual benchmarks. Note that we evaluate LLMs on the full test set of WMT22 benchmarks. Tables~\ref{table:benchmark_zhen_deen} and \ref{table:benchmark_mmt} report the numerical results of LLMs on these two benchmarks respectively. We also release their translation results as open source\footnote{Translation results of LLMs on WMT22 benchmark: \url{https://github.com/ictnlp/BayLing/tree/main/exp/translation_benchmark}}, which can serve as baselines in future translation research.

\begin{table}[h]
\caption{Results on WMT22 Chinese$\Leftrightarrow$English and Germain$\Leftrightarrow$English translation tasks}
\label{table:benchmark_zhen_deen}
\scriptsize
\begin{tabular} {lC{0.35cm}C{0.35cm}C{0.35cm}C{0.35cm}C{0.35cm}C{0.35cm}C{0.35cm}C{0.35cm}C{0.35cm}C{0.35cm}C{0.35cm}C{0.35cm}C{0.35cm}C{0.35cm}C{0.35cm}} \toprule
\multirow{3}{*}{\textbf{Systems}}&\multicolumn{12}{c}{\textbf{Multi-domain}}                                                                                                                                                       &\multicolumn{3}{c}{\multirow{2}{*}{\textbf{ALL}}} \\
                                 &\multicolumn{3}{c}{\textbf{Conversation}}     &\multicolumn{3}{c}{\textbf{News}}             &\multicolumn{3}{c}{\textbf{Social}}           &\multicolumn{3}{c}{\textbf{Ecommerce}}        &\multicolumn{3}{c}{}                              \\
                                 &\textbf{$\!\!\!\!$COMET}&\textbf{BLEU}&\textbf{chrF}&\textbf{$\!\!\!\!$COMET}&\textbf{BLEU}&\textbf{chrF}&\textbf{$\!\!\!\!$COMET}&\textbf{BLEU}&\textbf{chrF}&\textbf{$\!\!\!\!$COMET}&\textbf{BLEU}&\textbf{chrF}&\textbf{$\!\!\!\!$COMET} &\textbf{BLEU} &\textbf{chrF}  \\ \midrule
\multicolumn{16}{c}{\textit{\textbf{Chinese-to-English}}}                                                                                                                                                                                                                                                              \\\midrule
\textbf{GPT-4}      &83.87            &30.95             &53.84        &83.21            &27.54             &60.89        &83.41            &31.31             &61.72        &81.04            &20.78             &53.59        &82.79            &27.20             &58.95         \\
\textbf{GPT-3.5-turbo}   &84.03            &29.51             &53.13        &83.08            &26.68             &60.21        &83.43            &30.17             &60.22        &80.52            &19.36             &52.38        &82.64            &26.13             &57.90         \\
\textbf{Google Translate}&81.22            &23.57             &48.13        &80.70            &27.96             &60.06        &82.25            &35.47             &63.39        &79.25            &22.21             &53.24        &80.81            &28.63             &58.72         \\
\textbf{BLOOMZ-7B1-MT}   &76.47            &18.46             &39.48        &76.48            &15.26             &44.94        &74.98            &16.60             &42.95        &69.86            &11.69             &34.76        &74.25            &14.98             &41.59         \\
\textbf{NLLB-3.3B}       &78.08            &22.64             &42.30        &78.66            &21.76             &52.80        &78.53            &25.34             &53.50        &72.89            &13.17             &42.67        &76.92            &21.07             &49.91         \\
\textbf{ParroT-7B}       &73.54            &15.31             &37.13        &77.53            &20.39             &52.07        &75.74            &17.44             &48.17        &71.72            &10.06             &40.22        &74.70            &16.35             &47.19         \\
\textbf{Alpaca-7B}       &73.99            &18.89             &37.98        &74.52            &10.84             &40.65        &73.84            &13.36             &40.79        &71.34            &9.48              &36.02        &73.36            &11.80             &39.40         \\
\textbf{ChatGLM-6B}      &78.78            &21.09             &43.44        &75.01            &15.40             &46.25        &76.56            &18.70             &47.52        &73.21            &12.12             &39.67        &75.63            &16.08             &44.84         \\
\textbf{Vicuna-13B}      &69.58            &13.80             &40.83        &76.38            &16.82             &50.08        &75.98            &18.91             &50.52        &70.37            &11.17             &43.02        &73.35            &15.63             &47.92         \\ \arrayrulecolor{gray}\midrule \arrayrulecolor{black}
\textbf{BayLing-7B}      &76.98            &20.46             &41.19        &79.45            &21.54             &53.79        &78.60            &23.36             &51.64        &74.80            &14.66             &43.37        &77.48            &20.31             &49.94         \\
\textbf{BayLing-13B}     &77.87            &22.96             &43.09        &79.33            &20.05             &51.75        &78.94            &24.41             &53.10        &74.85            &14.26             &42.65        &77.72            &20.12             &49.44         \\\midrule
\multicolumn{16}{c}{\textit{\textbf{English-to-Chinese}}}                                                                                                                                                                                                                                                              \\\midrule
\textbf{GPT-4}      &89.01            &42.52             &41.63        &87.93            &48.87             &44.81        &84.56            &37.86             &34.86        &88.52            &43.14             &40.09        &87.49            &43.98             &41.00         \\
\textbf{GPT-3.5-turbo}   &88.63            &43.43             &42.03        &87.40            &49.41             &43.09        &83.54            &38.92             &34.71        &87.76            &43.81             &39.94        &86.81            &44.99             &40.32         \\
\textbf{Google Translate}&89.55            &47.33             &45.43        &87.75            &55.34             &49.55        &83.37            &42.35             &38.01        &88.78            &47.94             &44.90        &87.34            &49.89             &45.28         \\
\textbf{BLOOMZ-7B1-MT}   &79.98            &26.00             &24.40        &80.06            &24.93             &23.59        &74.89            &20.58             &19.49        &80.90            &23.84             &22.31        &78.96            &23.86             &22.49         \\
\textbf{NLLB-3.3B}       &85.65            &31.52             &29.29        &80.80            &35.14             &32.97        &76.59            &26.30             &24.15        &83.37            &33.51             &29.57        &81.56            &32.52             &29.66         \\
\textbf{ParroT-7B}       &75.76            &20.59             &22.60        &76.59            &26.91             &26.57        &71.05            &19.28             &19.14        &75.47            &19.99             &21.09        &74.71            &22.56             &23.01         \\
\textbf{Alpaca-7B}       &61.94            &13.32             &14.72        &40.53            &4.92              &7.17         &46.20            &4.84              &7.30         &58.64            &11.52             &14.05        &51.75            &7.80              &10.00         \\
\textbf{ChatGLM-6B}      &84.47            &32.62             &31.56        &78.92            &29.49             &27.17        &77.10            &25.02             &23.01        &83.15            &31.95             &29.72        &80.88            &30.30             &27.54         \\
\textbf{Vicuna-13B}      &81.90            &29.62             &29.18        &76.80            &24.82             &25.23        &74.80            &21.46             &22.01        &82.86            &28.71             &28.84        &79.09            &25.71             &25.97         \\\arrayrulecolor{gray}\midrule \arrayrulecolor{black}
\textbf{BayLing-7B}      &87.06            &40.06             &37.26        &85.15            &41.69             &37.33        &80.09            &30.23             &27.14        &85.52            &36.91             &34.80        &84.43            &38.19             &34.47         \\
\textbf{BayLing-13B}     &87.35            &40.14             &36.78        &85.12            &41.26             &37.20        &80.60            &30.59             &27.57        &85.52            &37.58             &33.80        &84.62            &37.92             &34.19         \\\midrule
\multicolumn{16}{c}{\textit{\textbf{German-to-English}}}                                                                                                                                                                                                                                                               \\\midrule
\textbf{GPT-4}      &86.08            &36.81             &56.29        &85.70            &31.55             &59.70        &84.75            &32.12             &56.97        &86.01            &35.69             &62.03        &85.62            &33.87             &59.29         \\
\textbf{GPT-3.5-turbo}   &85.86            &35.52             &55.45        &85.40            &30.83             &59.31        &84.74            &30.69             &55.67        &85.95            &35.30             &61.76        &85.47            &32.94             &58.65         \\
\textbf{Google Translate}&85.65            &34.53             &54.31        &84.30            &31.09             &58.48        &83.61            &31.57             &56.12        &85.56            &35.04             &60.91        &84.75            &33.21             &58.09         \\
\textbf{BLOOMZ-7B1-MT}   &73.67            &22.39             &40.18        &70.10            &17.74             &43.94        &69.28            &16.67             &39.66        &70.79            &19.03             &43.82        &70.89            &18.61             &42.32         \\
\textbf{NLLB-3.3B}       &83.62            &30.84             &51.19        &84.15            &28.13             &55.93        &82.14            &28.14             &52.52        &83.80            &31.35             &57.50        &83.42            &29.54             &54.93         \\
\textbf{ParroT-7B}       &79.65            &18.10             &46.64        &81.89            &19.81             &52.28        &79.04            &16.95             &46.31        &81.34            &19.74             &51.03        &80.49            &18.70             &49.58         \\
\textbf{Alpaca-7B}       &82.06            &27.02             &47.99        &81.74            &23.71             &52.12        &79.80            &22.75             &47.15        &81.99            &25.46             &52.31        &81.37            &24.52             &50.37         \\
\textbf{ChatGLM-6B}      &74.16            &21.77             &40.39        &70.66            &17.09             &43.54        &71.18            &17.17             &40.73        &68.71            &18.15             &42.58        &71.12            &18.26             &42.10         \\
\textbf{Vicuna-13B}      &80.25            &25.30             &50.86        &82.11            &26.84             &56.55        &78.33            &24.57             &51.56        &82.50            &29.08             &57.25        &80.80            &26.62             &54.71         \\\arrayrulecolor{gray}\midrule \arrayrulecolor{black}
\textbf{BayLing-7B}      &83.26            &31.26             &51.27        &83.96            &26.80             &54.92        &82.24            &27.44             &51.38        &83.33            &28.44             &55.20        &83.19            &28.16             &53.62         \\
\textbf{BayLing-13B}     &83.54            &31.16             &50.42        &83.66            &25.93             &53.97        &81.73            &25.69             &50.38        &83.24            &28.06             &54.88        &83.02            &27.34             &52.87         \\\midrule
\multicolumn{16}{c}{\textit{\textbf{English-to-German}}}                                                                                                                                                                                                                                                               \\\midrule
\textbf{GPT-4}      &88.68            &36.78             &64.30        &87.60            &34.36             &64.99        &84.93            &34.37             &60.03        &88.58            &36.86             &64.81        &87.44            &35.38             &63.76         \\
\textbf{GPT-3.5-turbo}   &87.76            &35.10             &63.56        &87.01            &33.85             &64.46        &84.57            &32.37             &58.26        &88.38            &35.48             &63.64        &86.93            &34.12             &62.76         \\
\textbf{Google Translate}&89.03            &42.61             &67.75        &86.97            &37.77             &65.63        &83.72            &35.51             &59.72        &88.65            &38.54             &65.31        &87.08            &38.27             &64.60         \\
\textbf{BLOOMZ-7B1-MT}   &57.60            &10.78             &32.46        &47.85            &7.14              &34.22        &47.84            &7.82              &32.60        &55.43            &7.36              &33.95        &52.14            &7.98              &33.51         \\
\textbf{NLLB-3.3B}       &87.52            &35.32             &62.51        &86.35            &33.06             &63.31        &83.58            &33.36             &57.90        &87.49            &34.87             &62.38        &86.23            &33.98             &61.77         \\
\textbf{ParroT-7B}       &73.83            &11.86             &42.99        &78.87            &16.56             &52.58        &71.36            &8.76              &41.01        &76.20            &11.32             &44.58        &75.09            &12.09             &46.27         \\
\textbf{Alpaca-7B}       &78.47            &16.66             &42.10        &79.26            &19.40             &49.86        &74.92            &17.50             &43.27        &80.10            &18.27             &46.69        &78.20            &18.22             &46.37         \\
\textbf{ChatGLM-6B}      &48.59            &4.33              &18.61        &48.44            &1.88              &23.49        &43.35            &1.36              &19.12        &46.08            &2.54              &21.58        &46.58            &2.33              &21.26         \\
\textbf{Vicuna-13B}      &79.85            &17.26             &47.83        &82.02            &23.60             &55.28        &76.93            &21.52             &49.71        &82.52            &21.93             &52.68        &80.35            &21.55             &52.19         \\\arrayrulecolor{gray}\midrule \arrayrulecolor{black}
\textbf{BayLing-7B}      &81.74            &25.31             &50.57        &84.26            &28.23             &58.23        &80.02            &24.77             &50.55        &82.65            &23.02             &51.62        &82.18            &25.66             &53.61         \\
\textbf{BayLing-13B}     &83.43            &24.67             &50.45        &84.34            &28.24             &58.11        &79.80            &25.37             &51.21        &83.19            &22.67             &52.17        &82.69            &25.62             &53.84        \\\bottomrule
\end{tabular}
\end{table}

\begin{table}[t]
\caption{Results on WMT22 multilingual translation tasks}
\label{table:benchmark_mmt}
\scriptsize
\begin{tabular}{lC{0.55cm}C{0.55cm}C{0.55cm}C{0.55cm}C{0.55cm}C{0.55cm}C{0.55cm}C{0.55cm}C{0.55cm}C{0.55cm}C{0.55cm}C{0.55cm}} \toprule
\multicolumn{13}{c}{\textit{\textbf{X-to-English}}}                                                                                                                                                                                                            \\\midrule 
\multicolumn{1}{c}{\multirow{2}{*}{\textbf{Systems}}}&\multicolumn{3}{c}{\textbf{Czech-to-English}} &\multicolumn{3}{c}{\textbf{Japanese-to-English}}&\multicolumn{3}{c}{\textbf{Russian-to-English}}&\multicolumn{3}{c}{\textbf{Ukrainian-to-English}} \\
\multicolumn{1}{c}{}                                 &\textbf{$\!\!\!\!$COMET}&\textbf{$\!$BLEU}&\textbf{chrF}&\textbf{$\!\!\!\!$COMET}&\textbf{$\!$BLEU} &\textbf{chrF}&\textbf{$\!\!\!\!$COMET}&\textbf{$\!$BLEU}&\textbf{chrF}&\textbf{$\!\!\!\!$COMET}&\textbf{$\!$BLEU} &\textbf{chrF}  \\
\textbf{GPT-4}	&87.43 	&48.67 	&71.42 	&83.20 	&24.57 	&51.41 	&86.18 	&43.51 	&68.79 	&85.67 	&40.47 	&65.71 \\
\textbf{GPT-3.5-turbo}                               &86.75         &45.99        &69.65        &82.39          &22.14         &50.24        &85.95          &41.79        &67.83        &85.32          &39.00         &64.26          \\
\textbf{Google Translate}                            &86.95         &49.26        &72.44        &81.69          &23.17         &48.83        &84.81          &43.54        &67.51        &85.55          &41.60         &66.94          \\
\textbf{NLLB-3.3B}                                   &85.92         &49.06        &69.86        &74.87          &15.00         &40.21        &83.95          &40.08        &64.77        &83.01          &36.19         &60.05          \\
\textbf{Alpaca-7B}                                   &78.00         &27.31        &52.74        &70.46          &8.63          &31.97        &80.68          &30.49        &56.64        &79.94          &28.57         &53.20          \\
\textbf{ChatGLM-6B}                                  &57.30         &7.32         &31.63        &66.89          &7.50          &30.80        &57.54          &9.36         &33.05        &55.60          &6.83          &27.00          \\
\textbf{Vicuna-13B}                                  &79.54         &30.74        &59.82        &69.00          &9.31          &37.55        &79.88          &29.77        &61.04        &79.60          &27.70         &57.68          \\\arrayrulecolor{gray}\midrule \arrayrulecolor{black}
\textbf{BayLing-7B}                                  &82.03         &35.98        &59.76        &72.16          &11.63         &34.71        &82.48          &34.74        &60.35        &81.38          &33.07         &56.97          \\
\textbf{BayLing-13B}                                 &81.65         &33.87        &58.93        &72.14          &12.23         &34.95        &82.07          &33.95        &59.80        &81.41          &32.67         &56.93          \\\midrule    
\multicolumn{13}{c}{\textit{\textbf{English-to-X}}}                                                                                                                                                                                                             \\\midrule    
\multirow{2}{*}{\textbf{Systems}}                    &\multicolumn{3}{c}{\textbf{English-to-Czech}} &\multicolumn{3}{c}{\textbf{English-to-Japanese}}&\multicolumn{3}{c}{\textbf{English-to-Russian}}&\multicolumn{3}{c}{\textbf{English-to-Ukrainian}} \\\multicolumn{1}{c}{}                                 &\textbf{$\!\!\!\!$COMET}&\textbf{$\!$BLEU}&\textbf{chrF}&\textbf{$\!\!\!\!$COMET}&\textbf{$\!$BLEU} &\textbf{chrF}&\textbf{$\!\!\!\!$COMET}&\textbf{$\!$BLEU}&\textbf{chrF}&\textbf{$\!\!\!\!$COMET}&\textbf{$\!$BLEU} &\textbf{chrF}  \\
\textbf{GPT-4}	&90.77 	&34.53 	&61.24 	&89.87 	&24.71 	&35.63 	&88.87 	&30.45 	&57.44 	&88.46 	&26.71 	&55.96 \\
\textbf{GPT-3.5-turbo}                               &90.05         &32.71        &59.68        &83.26          &22.22         &33.24        &87.52          &29.59        &56.13        &87.43          &25.87         &54.78          \\
\textbf{Google Translate}                            &91.28         &48.10        &68.85        &88.64          &26.50         &35.49        &88.91          &35.04        &59.76        &88.63          &32.05         &59.09          \\
\textbf{NLLB-3.3B}                                   &89.90         &36.30        &61.29        &84.73          &15.66         &25.97        &87.51          &30.11        &55.94        &86.64          &25.57         &54.02          \\
\textbf{Alpaca-7B}                                   &73.51         &13.13        &37.26        &44.98          &1.43          &4.54         &74.87          &14.10        &36.48        &73.34          &7.62          &28.00          \\
\textbf{ChatGLM-6B}                                  &40.50         &1.91         &15.23        &58.66          &2.42          &5.34         &45.48          &1.42         &1.94         &41.44          &1.41          &1.89           \\
\textbf{Vicuna-13B}                                  &76.85         &15.35        &42.92        &79.76          &11.92         &21.31        &78.57          &17.67        &43.58        &72.84          &12.07         &37.23          \\\arrayrulecolor{gray}\midrule \arrayrulecolor{black}
\textbf{BayLing-7B}                                  &76.85         &15.64        &41.64        &71.23          &4.51          &9.56         &74.72          &14.85        &35.54        &76.01          &11.66         &34.32          \\
\textbf{BayLing-13B}                                 &78.22         &16.43        &42.37        &71.39          &6.05          &12.45        &71.01          &12.77        &29.52        &66.83          &8.32          &22.78          \\\midrule    
\multicolumn{13}{c}{\textit{\textbf{X-to-X}}}                                                                                                                                                                                                                   \\\midrule    
\multirow{2}{*}{\textbf{Systems}}                    &\multicolumn{3}{c}{\textbf{German-to-French}}          &\multicolumn{3}{c}{\textbf{French-to-German}}            &               &             &             &               &              &               \\
                                                     &\textbf{$\!\!\!\!$COMET}         &\textbf{$\!$BLEU}         &\textbf{chrF}         &\textbf{$\!\!\!\!$COMET}          &\textbf{$\!$BLEU}          &\textbf{chrF}         &               &             &             &               &              &               \\
\textbf{GPT-4}	&84.20 	&41.12 	&64.83 	&87.49 	&47.82 	&71.13         &               &             &             &               &              &\\
\textbf{GPT-3.5-turbo}                               &83.75         &35.27        &62.08        &86.74          &45.74         &68.97        &               &             &             &               &              &               \\
\textbf{Google Translate}                            &84.71         &41.00        &71.15        &85.83          &46.59         &68.56        &               &             &             &               &              &               \\
\textbf{NLLB-3.3B}                                   &81.91         &38.31        &61.06        &85.05          &43.24         &65.76        &               &             &             &               &              &               \\
\textbf{Alpaca-7B}                                   &71.94         &18.61        &43.12        &73.87          &20.43         &44.74        &               &             &             &               &              &               \\
\textbf{ChatGLM-6B}                                  &50.01         &2.85         &19.59        &50.76          &3.42          &20.37        &               &             &             &               &              &               \\
\textbf{Vicuna-13B}                                  &66.71         &12.82        &36.69        &70.83          &19.98         &45.63        &               &             &             &               &              &               \\\arrayrulecolor{gray}\midrule \arrayrulecolor{black}
\textbf{BayLing-7B}                                  &77.80         &26.55        &51.44        &80.28          &29.64         &55.26        &               &             &             &               &              &               \\
\textbf{BayLing-13B}                                 &77.78         &24.73        &50.51        &80.27          &28.91         &54.80        &               &             &             &               &              &       \\\bottomrule       
\end{tabular}
\end{table}

\section{BayLing-80 Test Set}
\label{app:BayLing-80}

\begin{figure}
    \centering
    \includegraphics[width=0.8\textwidth]{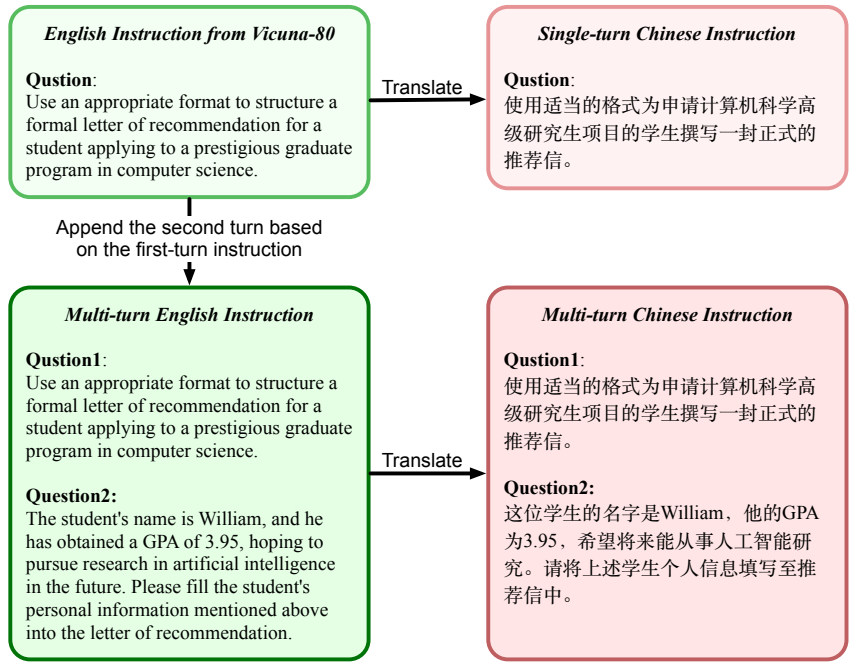}
    \caption{Construction of \textbf{BayLing-80}, a test set comprising Chinese and English multi-turn instructions for general tasks}
    \label{fig:BayLing-80}
\end{figure}

To assess the multilingual and multi-turn interactive capabilities of LLMs, we present BayLing-80, a multi-turn instruction test set augmented based on Vicuna-80 \cite{vicuna2023}. BayLing-80 comprises four components: (1) the original 80 single-turn English instructions in Vicuna-80 test set, (2) the corresponding translated single-turn Chinese instructions, (3) the expansion of the single-turn instructions into multi-turn English instructions, and (4) multi-turn Chinese instructions.

Specifically, we artificially construct the second turn of instructions based on the single turn of instructions. The second turn of instructions is semantically strongly related to the first turn of instructions, while omitting related references and implying some implicit logic to examine the context understanding ability of LLMs in multi-turn interactions. Figure~\ref{fig:BayLing-80} shows a test example in BayLing-80, where the second instruction is a supplement and additional constraint to the first instruction, to examine whether LLMs can adjust their previous responses to accommodate the user’s additional requirements.

\section{Templates of GPT-4 Evaluation}
\label{app:template}

Figures~\ref{fig:gpt4_template_single} and \ref{fig:gpt4_template_multi} show the templates for GPT-4 evaluation on single-turn instruction and multi-turn instruction respectively. We fill questions and the corresponding responses of two systems in the template, and then let GPT-4 score the responses of two systems. For a more accurate evaluation, following \citet{wang2023large}, we let GPT-4 generate the evaluation evidence first and then score the two systems.

\begin{figure}[h]

\scriptsize
\begin{tcolorbox}[title=The template of GPT-4 Evaluation for Single-turn Interaction]
You are a helpful and precise assistant for checking the quality of the answer.

[Question] 

\textcolor[rgb]{0,0,0.9}{\{$\mathrm{Q}$\}}

[The Start of Assistant 1's Answer] 

\textcolor[rgb]{0,0,0.9}{\{$\mathrm{R}_{1}$\}}

[The End of Assistant 1's Answer]

[The Start of Assistant 2's Answer] 

\textcolor[rgb]{0,0,0.9}{\{$\mathrm{R}_{2}$\}}

[The End of Assistant 2's Answer]

[System]

We would like to request your feedback on the performance of two AI assistants in response to the user question displayed above.

Please rate the helpfulness, relevance, accuracy, and level of detail of their responses. 
Each assistant receives an overall score on a scale of 1 to 10, where a higher score indicates better overall performance.

Please first provide a comprehensive explanation of your evaluation, avoiding any potential bias and ensuring that the order in which the responses were presented does not affect your judgment. Then, output two lines indicating the scores for Assistant 1 and 2, respectively.

Output with the following format:

Evaluation evidence: <evaluation explanation here>

The score of Assistant 1: <score>

The score of Assistant 2: <score>

\end{tcolorbox}
\caption{The template of GPT-4 evaluation for single-turn interaction. \textcolor[rgb]{0,0,0.9}{$\mathrm{Q}$} is the question from user, and \textcolor[rgb]{0,0,0.9}{$\mathrm{R}_{1}$} and \textcolor[rgb]{0,0,0.9}{$\mathrm{R}_{2}$} are the responses of \textbf{system1} and \textbf{system2} to this question, respectively.}
\label{fig:gpt4_template_single}
\end{figure}

\begin{figure}[h]
\scriptsize
\begin{tcolorbox}[title=The template of GPT-4 Evaluation for Multi-turn Interaction]
You are a helpful and precise assistant for checking the quality of the \textcolor[rgb]{0.8,0,0}{multi-turn interaction}.

[Question1] 

\textcolor[rgb]{0,0,0.9}{\{$\mathrm{Q}_{1}$\}}

[Question2] 

\textcolor[rgb]{0,0,0.9}{\{$\mathrm{Q}_{2}$\}}

[The Start of Assistant 1's Answer] 

Question1: \textcolor[rgb]{0,0,0.9}{\{$\mathrm{Q}_{1}$\}}

Answer1: \textcolor[rgb]{0,0,0.9}{\{$\mathrm{R}_{1,1}$\}}

Question2: \textcolor[rgb]{0,0,0.9}{\{$\mathrm{Q}_{2}$\}}

Answer2: \textcolor[rgb]{0,0,0.9}{\{$\mathrm{R}_{1,2}$\}}

[The End of Assistant 1's Answer]

[The Start of Assistant 2's Answer]

Question1: \textcolor[rgb]{0,0,0.9}{\{$\mathrm{Q}_{1}$\}}

Answer1: \textcolor[rgb]{0,0,0.9}{\{$\mathrm{R}_{2,1}$\}}

Question2: \textcolor[rgb]{0,0,0.9}{\{$\mathrm{Q}_{2}$\}}

Answer2: \textcolor[rgb]{0,0,0.9}{\{$\mathrm{R}_{2,2}$\}}

[The End of Assistant 2's Answer]

[System]

We would like to request your feedback on the performance of two AI assistants in \textcolor[rgb]{0.8,0,0}{a multi-turn interaction with the user's question1 and question2 displayed above. Please rate the helpfulness, relevance, accuracy, consistency, contextual understanding, level of details of their multi-turn interaction.} 

Please rate the helpfulness, relevance, accuracy, and level of detail of their responses. 
Each assistant receives an overall score on a scale of 1 to 10, where a higher score indicates better overall performance.

Please first provide a comprehensive explanation of your evaluation, avoiding any potential bias and ensuring that the order in which the responses were presented does not affect your judgment. Then, output two lines indicating the scores for Assistant 1 and 2, respectively.

Output with the following format:

Evaluation evidence: <evaluation explanation here>

The score of Assistant 1: <score>

The score of Assistant 2: <score>

\end{tcolorbox}
\caption{The template of GPT-4 evaluation for multi-turn interaction. \textcolor[rgb]{0,0,0.9}{$\mathrm{Q}_{1}$} and \textcolor[rgb]{0,0,0.9}{$\mathrm{Q}_{2}$} are two-turn questions from the user. \textcolor[rgb]{0,0,0.9}{$\mathrm{R}_{1,1}$} and \textcolor[rgb]{0,0,0.9}{$\mathrm{R}_{1,2}$} are the responses of \textbf{system1} to \textcolor[rgb]{0,0,0.9}{$\mathrm{Q}_{1}$} and \textcolor[rgb]{0,0,0.9}{$\mathrm{Q}_{2}$}, respectively. \textcolor[rgb]{0,0,0.9}{$\mathrm{R}_{2,1}$} and \textcolor[rgb]{0,0,0.9}{$\mathrm{R}_{2,2}$} are the responses of \textbf{system2} to \textcolor[rgb]{0,0,0.9}{$\mathrm{Q}_{1}$} and \textcolor[rgb]{0,0,0.9}{$\mathrm{Q}_{2}$}, respectively.}
\label{fig:gpt4_template_multi}
\end{figure}
\newpage
\section{Cases on General Tasks}
In Tables~\ref{table:case15}, \ref{table:case2}, \ref{table:case73} and \ref{table:case3}, we present several examples of responses from BayLing and other LLMs on the BayLing-80 test set. Additionally, we also provide the reviews and scores assigned by GPT-4. More cases on the BayLing-80 test set can be found here\footnote{\url{https://github.com/ictnlp/BayLing/tree/main/exp/general_tasks}}.

\begin{CJK*}{UTF8}{gbsn}

\begin{table}[t]
\caption{Case\#15 of BayLing-80 in English}\label{table:case15}
\fontsize{4.5pt}{3pt}\selectfont
\begin{tabular}{L{0.225\textwidth}L{0.225\textwidth}L{0.225\textwidth}L{0.225\textwidth}} \toprule
\multicolumn{1}{c}{\textbf{BayLing-13B}}                                                                                                                                                                                                                                                           & \multicolumn{1}{c}{\textbf{GPT-3.5-turbo}}                                                                                                                                                                                                                                                              & \multicolumn{1}{c}{\textbf{Vicuna-13B}}                                                                                                                                                                                                                                  & \multicolumn{1}{c}{\textbf{ChatGLM-6B}}                                                                                                                                                                                                                                                                                                                                                                \\\midrule 
\multicolumn{4}{C{0.98\textwidth}}{\textcolor[rgb]{0,0,0.8}{\textbf{Question1}: Describe a scenario where artificial intelligence could be used to improve the quality and efficiency of healthcare delivery.}}                                                                                                                                                                                             \\\midrule

\justifying One scenario where artificial intelligence could be used to improve the quality and efficiency of healthcare delivery is through the implementation of a virtual personal health assistant. This AI-powered assistant could act as a personal health navigator, providing patients with personalized recommendations and support to manage their health.\newline \newline For example, the AI assistant could use data from a patient's medical records, as well as from wearable devices such as fitness trackers or smartwatches, to monitor their health and identify potential health risks or warning signs. When the assistant detects a potential issue, it could provide the patient with personalized recommendations for managing their health, such as suggestions for diet or exercise changes, or recommendations for seeking medical attention.\newline \newline In addition to managing health risks, the AI assistant could also help patients more effectively manage chronic conditions such as diabetes or heart disease. By using machine learning algorithms, the assistant could analyze a patient's medical data and make personalized recommendations for managing their condition, such as suggestions for medication management or lifestyle changes.\newline \newline The virtual personal health assistant could also be integrated with electronic health records systems, allowing it to automatically update a patient's medical records with their progress and compliance with recommended health management strategies. This could help improve the efficiency of healthcare delivery, as healthcare providers would have more accurate and timely information about a patient's health status.\newline \newline Overall, the use of artificial intelligence in the form of a virtual personal health assistant could improve the quality and efficiency of healthcare delivery by providing patients with personalized support and guidance for managing their health, while also helping healthcare providers more effectively monitor and manage their patients' health.

&   \justifying   One scenario where artificial intelligence could be used to improve the quality and efficiency of healthcare delivery is in the field of diagnostics. Currently, doctors often rely on their own knowledge and experience to diagnose patients, which can lead to errors or delays in treatment. By using AI algorithms to analyze patient data, such as medical images, lab results, and clinical histories, doctors could receive more accurate and timely diagnoses.\newline \newline For example, an AI system could be trained to identify patterns in medical images that are associated with specific diseases or conditions, allowing doctors to quickly and accurately diagnose patients. This could be particularly useful in fields like radiology, where large volumes of medical images must be analyzed regularly.\newline \newline AI could also be used to streamline administrative tasks in healthcare, such as scheduling appointments, managing patient records, and processing insurance claims. By automating these processes, providers could focus more on patient care, reducing wait times and improving overall efficiency.\newline \newline Overall, the use of artificial intelligence in healthcare has the potential to revolutionize the way we deliver and receive medical care, leading to better outcomes for patients and providers alike.

&    \justifying  Artificial intelligence (AI) has the potential to revolutionize the healthcare industry by improving the quality and efficiency of healthcare delivery. One scenario where AI could be used to achieve this is in the area of patient diagnosis and treatment.\newline \newline Imagine a scenario where a patient visits a doctor's office with symptoms that are difficult to diagnose. The doctor can input the patient's medical history, symptoms, and other relevant information into an AI system that is trained to diagnose and treat various medical conditions. The AI system can analyze the data and provide a list of possible diagnoses, along with the likelihood of each condition. The doctor can then use this information to make a more informed diagnosis and provide treatment accordingly.\newline \newline AI can also be used to improve the efficiency of healthcare delivery. For example, AI can be used to analyze a patient's medical history and predict the likelihood of certain health problems, such as diabetes or heart disease. This can help doctors to identify patients who are at high risk of developing these conditions and take preventative measures.\newline \newline AI can also be used to automate routine tasks, such as scheduling appointments or processing insurance claims. This can free up healthcare professionals to focus on more complex and critical tasks, such as providing patient care.\newline \newline Overall, the use of AI in healthcare has the potential to improve the quality and efficiency of healthcare delivery by providing faster and more accurate diagnoses and treatments, identifying high-risk patients, and automating routine tasks.

&  \justifying   One scenario where artificial intelligence could be used to improve the quality and efficiency of healthcare delivery is in the area of patient navigation. AI algorithms could be trained to help healthcare providers navigate complex patient care processes, such as scheduling appointments, managing medications, and providing care coordination. This could help to reduce errors and improve the overall efficiency of healthcare delivery. Additionally, AI-powered tools could be used to analyze large amounts of healthcare data and identify patterns and trends that could inform decision-making and improve patient outcomes. \cr

                       \\\midrule

\multicolumn{4}{C{0.98\textwidth}}{\textcolor[rgb]{0,0,0.8}{\textbf{Question2}: What are the specific ethical considerations that should be addressed in this scenario?}}                                                                                                                                                                  \\\midrule 
\justifying  In the scenario where artificial intelligence is used to improve the quality and efficiency of healthcare delivery through a virtual personal health assistant, there are several ethical considerations that should be addressed. These include:\newline \newline 1. Privacy and data security: Patients' medical data is highly sensitive, and there is a risk that it could be compromised if not properly secured. Ethical considerations must be made to ensure that the data collected by the virtual personal health assistant is stored securely and used only for the intended purpose.\newline 2. Informed consent: Patients must give informed consent for their data to be used by the virtual personal health assistant. Ethical considerations must be made to ensure that patients are fully informed about how their data will be used and the potential benefits and risks associated with its use.\newline 3. Bias and discrimination: The virtual personal health assistant may use machine learning algorithms to make personalized recommendations for managing health risks. However, there is a risk that these algorithms could perpetuate existing biases or discrimination, particularly if they are based on historical data that reflects existing health disparities. Ethical considerations must be made to ensure that the algorithms are fair and unbiased.\newline 4. Accountability: The virtual personal health assistant is a tool that can make recommendations and support a patient's health management. However, it is not a substitute for human healthcare providers. Ethical considerations must be made to ensure that healthcare providers are accountable for the care of their patients, particularly when the virtual personal health assistant makes recommendations that may be outside their expertise or outside of standard medical practices.

&    \justifying   While the use of artificial intelligence in healthcare has many potential benefits, there are also ethical considerations that must be addressed to ensure that patient privacy and autonomy are protected. Some specific ethical considerations that should be addressed in this scenario include:\newline \newline 1. Data privacy: AI systems rely on large amounts of patient data to function effectively, which raises concerns about how this data is collected, stored, and shared. Providers must ensure that patient data is protected from unauthorized access or use.\newline \newline 2. Bias: AI algorithms can be biased if they are trained on incomplete or biased datasets. This could lead to inaccurate diagnoses or treatment recommendations for certain patient populations. Providers must ensure that their AI systems are trained on diverse and representative datasets.\newline \newline 3. Informed consent: Patients have the right to understand how their data is being used and to give informed consent before it is used for research or other purposes. Providers must ensure that patients are fully informed about how their data will be used in AI systems and have the opportunity to opt-out if they wish.\newline \newline 4. Accountability: Providers must be accountable for the decisions made by their AI systems. They must ensure that these systems are transparent and that patients can appeal decisions made by AI systems if necessary.\newline \newline Overall, providers must ensure that the use of AI in healthcare is transparent, accountable, and respects patient autonomy and privacy. By addressing these ethical considerations, providers can ensure that AI enhances the quality and efficiency of healthcare delivery without compromising patient rights.

&   \justifying   The use of AI in healthcare raises several ethical considerations. Here are some specific ethical considerations that should be addressed in the scenario where AI is used to improve the quality and efficiency of healthcare delivery:\newline \newline 1. Bias and discrimination: AI systems can perpetuate biases and discrimination if they are trained on biased data. For example, if an AI system used to diagnose medical conditions is trained on data that is not representative of diverse populations, it may be less accurate for certain groups of patients.\newline 2. Privacy and confidentiality: AI systems may require access to sensitive patient data, such as medical records or genetic information. Healthcare providers must ensure that this data is kept confidential and secure, and that patients have control over their own data.\newline 3. Informed consent: Patients should be informed about how their data is being used and have the ability to consent to the use of AI in their care. Healthcare providers must also ensure that patients understand the potential benefits and risks of AI-assisted diagnosis and treatment.\newline 4. Transparency and explainability: AI systems used in healthcare should be transparent and explainable, so that healthcare providers and patients can understand how the system arrived at its diagnosis or treatment recommendations.\newline 5. Accountability: Healthcare providers must be accountable for the performance of AI systems used in patient care. This includes ensuring that the systems are accurate and reliable, and that they do not cause harm to patients.\newline 6. Equity and access: The use of AI in healthcare should not widen the gap between those who have access to advanced medical care and those who do not. Healthcare providers must ensure that AI-assisted diagnosis and treatment is available to all patients, regardless of their socioeconomic status or other factors.\newline \newline Overall, the ethical use of AI in healthcare requires careful consideration of these and other factors to ensure that patients receive the best possible care while protecting their privacy and rights.

&    \justifying  There are several ethical considerations that should be addressed when using artificial intelligence in the healthcare setting. Some of the key issues include:\newline \newline 1. Privacy: AI-powered systems could potentially collect and store sensitive healthcare data, such as patient information and medical history. It is important to ensure that this data is protected and used in a way that respects the privacy of patients.\newline 2. Bias: AI algorithms are only as good as the data they are trained on, and they can be influenced by biases and errors that were present in the data. It is important to carefully consider and mitigate these biases when developing and using AI-powered systems.\newline 3. Accountability: It is important to ensure that the AI systems being used are accountable for any errors or biases they may make. This could involve measures such as regular testing and monitoring, as well as ethical guidelines and standards for the use of AI in healthcare.\newline 4. 信息安全： AI-powered systems could potentially contain sensitive information that is intended for public access or that could be used to harm individuals. It is important to ensure that the systems are secure and that any sensitive information is protected from unauthorized access or misuse.\newline 5. Ethical considerations in the development and use of AI-powered systems: It is important to consider ethical considerations in the development and use of AI-powered systems, including issues such as fairness, accountability, and transparency. \cr

 \\\midrule 
\multicolumn{4}{C{0.98\textwidth}}{\textcolor[rgb]{0.9,0,0}{\textbf{GPT-4 Evaluation: BayLing-13B v.s. GPT-3.5-turbo}}}                                                                                                                                                                                     \\
\multicolumn{4}{L{0.98\textwidth}}{ \justifying Evaluation evidence: Both Assistant 1 and 2 provided relevant and informative answers to the question, addressing both the scenario where AI can be used in healthcare and the ethical considerations. Assistant 1 described a scenario involving a virtual personal health assistant providing personalized support and management for patients, while Assistant 2 focused on AI in diagnostics and streamlining administrative tasks in healthcare. Both assistants provided detailed explanations and ethical considerations. However, Assistant 1 provided a slightly more comprehensive response with a clear breakdown of the ethical considerations, while Assistant 2 had a slightly less detailed discussion of the ethical considerations.\newline \newline Score of the Assistant 1: 9\newline Score of the Assistant 2: 8

}                                                                                                                                                 \\\midrule 
\multicolumn{4}{C{0.98\textwidth}}{\textcolor[rgb]{0.9,0,0}{\textbf{GPT-4 Evaluation: BayLing-13B v.s. Vicuna-13B}}}                                                                                                                                                                                           \\
\multicolumn{4}{L{0.98\textwidth}}{ \justifying Evaluation evidence: Both Assistant 1 and Assistant 2 provided helpful and relevant answers, elaborating on different scenarios involving artificial intelligence in healthcare. Assistant 1 focused on the virtual personal health assistant while Assistant 2 discussed AI in patient diagnosis and treatment. Both assistants also addressed the ethical considerations in their respective scenarios, with some similar points such as bias and discrimination, privacy, informed consent, and accountability. Assistant 1 provided a slightly more detailed response, fully explaining the virtual personal health assistant's potential benefits and ethical considerations.\newline \newline Score of the Assistant 1: 9.5\newline Score of the Assistant 2: 8.5

}                                                                                                   \\\midrule 
\multicolumn{4}{C{0.98\textwidth}}{\textcolor[rgb]{0.9,0,0}{\textbf{GPT-4 Evaluation: BayLing-13B v.s. ChatGLM-6B}}   }         

\\
\multicolumn{4}{L{0.98\textwidth}}{ \justifying Evaluation evidence: Both Assistant 1 and Assistant 2 provided relevant and accurate information on the application of artificial intelligence in healthcare delivery. Assistant 1 provided a more comprehensive and structured response by detailing a specific scenario involving virtual personal health assistants and addressing the ethical considerations in a clear and organized manner. Assistant 2's response was more concise but still relevant, focusing on patient navigation as a potential use case for AI in healthcare. However, Assistant 2's response was less detailed and somewhat repetitive when discussing ethical considerations.\newline \newline Score of the Assistant 1: 9\newline Score of the Assistant 2: 7

}             \\\bottomrule                                                                                                                                             
\end{tabular}
\end{table}
\end{CJK*}

\begin{CJK*}{UTF8}{gbsn}

\begin{table}[t]
\caption{Case\#2 of BayLing-80 in English}\label{table:case2}
\fontsize{4.5pt}{3pt}\selectfont
\begin{tabular}{L{0.225\textwidth}L{0.225\textwidth}L{0.225\textwidth}L{0.225\textwidth}} \toprule
\multicolumn{1}{c}{\textbf{BayLing-13B}}                                                                                                                                                                                                                                                           & \multicolumn{1}{c}{\textbf{GPT-3.5-turbo}}                                                                                                                                                                                                                                                              & \multicolumn{1}{c}{\textbf{Vicuna-13B}}                                                                                                                                                                                                                                  & \multicolumn{1}{c}{\textbf{ChatGLM-6B}}                                                                                                                                                                                                                                                                                                                                                                \\\midrule 
\multicolumn{4}{C{0.98\textwidth}}{\textcolor[rgb]{0,0,0.8}{\textbf{Question1}: What are the most effective strategies for conflict resolution in the workplace?}}                                                                                                                                                                                             \\\midrule

\justifying Conflict resolution in the workplace involves finding a solution to a problem or dispute between two or more parties. Effective strategies for conflict resolution can help to prevent negative effects on productivity, teamwork, and overall workplace culture. Some strategies for conflict resolution include:\newline \newline 1. Active Listening: Encourage all parties involved in the conflict to express their viewpoints clearly and without interruption. Listening attentively and asking appropriate follow-up questions can help to clarify misunderstandings and reveal common ground.\newline \newline 2. Collaboration: Promote an atmosphere of cooperation by encouraging the parties to work together towards a resolution. Collaboration can help to build trust and understanding, and can lead to more creative and mutually acceptable solutions.\newline \newline 3. Compromise: Encourage the parties involved in the conflict to compromise and find a solution that is acceptable to everyone. Collaboration and compromise can help to prevent the escalation of conflict and can foster a positive working relationship.\newline \newline 4. Mediation: If the conflict is particularly difficult or emotional, consider involving a neutral third-party mediator to facilitate communication and promote a resolution.\newline \newline 5. Conflict Prevention: Conflict prevention involves addressing potential sources of conflict before they escalate. This can include communication training, team building activities, and conflict management training. Preventing conflict can help to maintain a positive and productive workplace environment.

&   \justifying   1. Communication: Effective communication is key to resolving conflicts in the workplace. Encouraging open and honest dialogue can help identify the root cause of the conflict, and finding a solution that works for all parties involved.\newline \newline 2. Active Listening: This involves listening to understand and not just to respond. It requires giving full attention to the other person's perspective and acknowledging their feelings and concerns.\newline \newline 3. Collaboration: Encouraging collaboration among team members can help find a solution that benefits everyone. Teams should work together to identify solutions that meet everyone's needs.\newline \newline 4. Compromise: Finding a middle ground can be an effective way to resolve a conflict. Both parties may need to make concessions or adjustments for a resolution to be reached.\newline \newline 5. Mediation: A neutral third party can be brought in to help facilitate the conversation and guide the parties towards a resolution.\newline \newline 6. Training: Providing training on conflict resolution skills can equip employees with the tools they need to effectively manage conflicts when they arise.\newline \newline 7. Clear Policies and Procedures: Establishing clear policies and procedures for conflict resolution can help prevent conflicts from escalating and provide a framework for addressing them when they do occur.

&    \justifying  Conflict resolution in the workplace is an important aspect of maintaining a productive and positive work environment. Here are some strategies that can be effective in resolving conflicts in the workplace:\newline \newline 1. Active Listening: One of the most important strategies for conflict resolution in the workplace is active listening. This involves listening carefully to the concerns of all parties involved in the conflict, and making an effort to understand their perspectives.\newline 2. Clear Communication: Clear and direct communication is essential for resolving conflicts in the workplace. It's important to express your own concerns and needs clearly, while also listening to the other person's point of view.\newline 3. Problem-Solving: Conflict resolution often involves finding a solution to the problem that is causing the conflict. Encourage both parties to work together to find a solution that works for everyone.\newline 4. Mediation: In some cases, it may be helpful to bring in a mediator to help resolve the conflict. A mediator can help both parties communicate and find a solution that works for everyone.\newline 5. Focus on Interests, Not Positions: When resolving conflicts in the workplace, it's important to focus on the interests of both parties, not just their positions. This means looking at the underlying needs and concerns that are driving the conflict.\newline 6. Follow Up: After the conflict has been resolved, it's important to follow up to ensure that the solution is working and that everyone is satisfied. This can help prevent future conflicts and build trust among team members.\newline \newline Remember, effective conflict resolution in the workplace requires patience, empathy, and a willingness to work together to find a solution that benefits everyone.

&  \justifying   There are several strategies that can be effective for conflict resolution in the workplace:\newline \newline 1. Active listening: This involves fully engaging in the other person's voice and emotions, allowing them to express themselves clearly, and responding with empathy and understanding.\newline 2. Conflict resolution training: Providing employees with the training and resources they need to effectively resolve conflicts is an important step towards reducing conflict in the workplace.\newline 3. Conflict-resolution models: Using established conflict-resolution models, such as the Model for Resolution of Conflicts in the Workplace, can help employees develop a healthy and effective way of resolving conflicts.\newline 4. Communication: Clear and effective communication is key to resolving conflicts in the workplace. This includes using written communication, such as emails or reports, as well as verbal communication, such as meetings.\newline 5. Respect: Ensuring that all employees are treated with respect and dignity can help to prevent conflicts in the workplace. This includes treating everyone equally and avoiding negative or disrespectful language.\newline 6. Conflict resolution policies: Establishing clear conflict resolution policies and procedures can help to ensure that conflicts are properly resolved and that employees are held accountable for their actions.
 \cr

                       \\\midrule

\multicolumn{4}{C{0.98\textwidth}}{\textcolor[rgb]{0,0,0.8}{\textbf{Question2}: Given 3 strategies, sorted by effectiveness.}}                                                                                                                                                                  \\\midrule 
\justifying  The effectiveness of conflict resolution strategies can vary depending on the specific situation and the personalities involved. However, some strategies are generally considered more effective than others. Here are three strategies for conflict resolution, sorted by effectiveness:\newline \newline 1. Collaboration: Encouraging collaboration can be an effective conflict resolution strategy because it fosters an atmosphere of cooperation and communication. Collaboration can help to build trust and understanding, and can lead to more creative and mutually acceptable solutions.\newline 2. Compromise: Encouraging parties to compromise can be an effective conflict resolution strategy because it allows for a win-win solution. Collaboration and compromise can help to prevent the escalation of conflict and can foster a positive working relationship.\newline 3. Mediation: Mediation can be an effective conflict resolution strategy if the conflict is particularly difficult or emotional. A neutral third-party mediator can facilitate communication and promote a resolution. Mediation can help to maintain a positive and productive workplace environment.

&    \justifying   It's difficult to rank strategies by effectiveness as it depends on the specific situation and individuals involved. However, here is one possible ranking based on general effectiveness:\newline \newline 1. Communication: Effective communication is often cited as the most important strategy for resolving conflicts in the workplace. It involves actively listening and expressing oneself clearly and respectfully. By encouraging open dialogue and understanding, parties can work together to find a mutually beneficial solution.\newline \newline 2. Collaboration: Encouraging collaboration among team members is also an effective strategy for resolving conflicts. By working together towards a common goal and finding a solution that benefits everyone, conflicts can be resolved in a way that maintains positive relationships and fosters teamwork.\newline \newline 3. Mediation: While mediation can be effective in some situations, it may not always be necessary or appropriate. If parties are willing and able to communicate and work collaboratively, they may be able to resolve the conflict without the need for a third party mediator. However, if communication has broken down or there is a power imbalance between parties, mediation may be necessary to facilitate a resolution.

&   \justifying   It's difficult to rank conflict resolution strategies in terms of effectiveness, as different strategies may work better in different situations and with different personalities. However, here are three strategies for conflict resolution in the workplace, sorted by general effectiveness:\newline \newline 1. Active Listening: Active listening is a highly effective conflict resolution strategy because it helps to build trust and understanding between parties. By actively listening to the concerns of others and making an effort to understand their perspectives, you can help to diffuse the conflict and create an environment that is conducive to collaboration and problem-solving.\newline 2. Problem-Solving: Problem-solving is another highly effective conflict resolution strategy because it focuses on finding a solution to the underlying issue that is causing the conflict. By working together to find a solution, both parties can feel heard and respected, and the conflict can be resolved in a way that benefits everyone.\newline 3. Mediation: Mediation can be an effective conflict resolution strategy in some cases, particularly when there are strong emotions or entrenched positions on both sides. A mediator can help facilitate communication and guide both parties towards a resolution that works for everyone.\newline \newline Remember, there is no one-size-fits-all solution to conflict resolution, and what works in one situation may not work in another. It's important to approach conflict resolution with an open mind and a willingness to try different strategies until a resolution is reached.

&    \justifying  Here are the three strategies for conflict resolution in the workplace sorted by their effectiveness:\newline \newline 1. Active listening: This strategy is highly effective in resolving conflicts because it allows the person being heard to express themselves clearly and honestly, which can help to clarify the situation and find a mutually acceptable resolution.\newline 2. Conflict-resolution models: Using established conflict-resolution models, such as the Model for Resolution of Conflicts in the Workplace, can be very effective in training employees on how to resolve conflicts in a healthy and effective way. These models provide a framework for resolving conflicts and can be customized to fit the unique needs of each workplace.\newline 3. Communication: Clear and effective communication is crucial for resolving conflicts in the workplace. This includes using written communication, such as emails or reports, as well as verbal communication, such as meetings. Effective communication can help to ensure that all parties involved are aware of the situation and can provide feedback and suggestions for resolution.
 \cr

 \\\midrule 
\multicolumn{4}{C{0.98\textwidth}}{\textcolor[rgb]{0.9,0,0}{\textbf{GPT-4 Evaluation: BayLing-13B v.s. GPT-3.5-turbo}}}                                                                                                                                                                                     \\
\multicolumn{4}{L{0.98\textwidth}}{ \justifying Evaluation evidence: Both assistants provided relevant and detailed information on strategies for conflict resolution in the workplace. Assistant 1 offered five strategies, while Assistant 2 provided seven. However, Assistant 2's response appears more organized, distinguishing general strategies first and then ranking the top three in order of effectiveness. Assistant 1's response includes an extra strategy (Conflict Prevention) that doesn't directly address the question about resolving conflicts but focuses on preventing them instead.\newline \newline Score of the Assistant 1: 8\newline Score of the Assistant 2: 9

}                                                                                                                                                 \\\midrule 
\multicolumn{4}{C{0.98\textwidth}}{\textcolor[rgb]{0.9,0,0}{\textbf{GPT-4 Evaluation: BayLing-13B v.s. Vicuna-13B}}}                                                                                                                                                                                           \\
\multicolumn{4}{L{0.98\textwidth}}{ \justifying Evaluation evidence: Both Assistant 1 and Assistant 2 provided comprehensive answers with various strategies for conflict resolution in the workplace. However, Assistant 1 seemed to better address the request of providing a list of three strategies sorted by effectiveness. Assistant 2 did provide a list of three strategies sorted by effectiveness but mentioned that it is difficult to rank the strategies in terms of effectiveness, which might cause confusion for the reader. Both answers included extra strategies to provide a broader insight into the topic. Assistant 1's answer structure was slightly easier to follow.\newline \newline Score of the Assistant 1: 9\newline Score of the Assistant 2: 8

}                                                                                                   \\\midrule 
\multicolumn{4}{C{0.98\textwidth}}{\textcolor[rgb]{0.9,0,0}{\textbf{GPT-4 Evaluation: BayLing-13B v.s. ChatGLM-6B}}   }         

\\
\multicolumn{4}{L{0.98\textwidth}}{ \justifying Evaluation evidence: Both assistants provided relevant information and strategies, but Assistant 1 organized the response more effectively by focusing on conflict resolution strategies and sorting them by effectiveness, as requested by the user. Assistant 2's answer did not directly address the effectiveness sorting aspect of the user's question but provided useful strategies. Assistant 1's answer demonstrated better understanding of the user's question and offered a more precise response.\newline \newline Score of the Assistant 1: 9\newline Score of the Assistant 2: 7

}             \\\bottomrule                                                                                                                                             
\end{tabular}
\end{table}
\end{CJK*}

\begin{CJK*}{UTF8}{gbsn}

\begin{table}[]
\caption{Case\#73 of BayLing-80 in Chinese}\label{table:case73}
\fontsize{4.5pt}{3pt}\selectfont
\begin{tabular}{L{0.225\textwidth}L{0.225\textwidth}L{0.225\textwidth}L{0.225\textwidth}} \toprule
\multicolumn{1}{c}{\textbf{BayLing-13B}}                                                                                                                                                                                                                                                           & \multicolumn{1}{c}{\textbf{GPT-3.5-turbo}}                                                                                                                                                                                                                                                              & \multicolumn{1}{c}{\textbf{Vicuna-13B}}                                                                                                                                                                                                                                  & \multicolumn{1}{c}{\textbf{ChatGLM-6B}}                                                                                                                                                                                                                                                                                                                                                                \\\midrule 
\multicolumn{4}{C{0.98\textwidth}}{\textcolor[rgb]{0,0,0.8}{\textbf{Question1}: 使用适当的格式为申请计算机科学高级研究生项目的学生撰写一封正式的推荐信。}}                                                                                                                                                                                             \\\midrule 
亲爱的招生委员会成员：\newline \newline 我们很高兴为您推荐此学生申请计算机科学高级研究生项目。\newline \newline {[}学生姓名{]} 是一位在我们的计算机科学专业中表现出色的学生。他/她在数据结构、算法、操作系统、编程语言等方面都有着深入的理解和扎实的技能。他/她在团队中扮演着领导者的角色，带领团队完成了多项团队项目，并多次获得团队的表现奖。\newline \newline {[}学生姓名{]} 还积极参加校内外的科研活动，并取得了优异的成绩。在校内，他/她曾获得过多项奖项，包括最佳论文奖、最佳计算机科学学生奖等；在校外，他/她多次获得国家级、国际赛的奖项，包括全国大学生科研创新实践中奖、国际大学生计算机科学竞赛等。\newline \newline 总之，{[}学生姓名{]} 是一位具有极高潜力和优秀表现的学生，我们强烈推荐他/她申请计算机科学高级研究生项目。我们肯定，他/她一定会在您的学校中成为一名优秀的学生和研究者。\newline \newline 此致，\newline \newline {[}你的名字{]}\newline {[}你的职位{]}\newline {[}你的学校{]} & {[}您的姓名{]}\newline {[}您的职位或头衔{]}\newline {[}您所在的机构或公司{]}\newline {[}您的地址{]}\newline {[}日期{]}\newline \newline 致：计算机科学高级研究生项目招生委员会\newline \newline 尊敬的招生委员会：\newline \newline 我写此信是为了强烈推荐 {[}申请人姓名{]} 先生/女士参加贵校计算机科学高级研究生项目。作为 {[}申请人姓名{]} 的 {[}教授/导师/上司/同事等{]}，我对他/她的表现非常印象深刻，深信他/她已准备好接受更高层次的学术挑战。\newline \newline 在过去的 {[}时间段{]} 中，我与 {[}申请人姓名{]} 一起工作，他/她在 {[}项目/课程/研究领域{]} 上表现出色，取得了令人满意的成绩。他/她在 {[}技能/知识领域{]} 上有着扎实的基础，并且具有很强的学习和创新能力。他/她在 {[}研究/实验室/项目{]} 中的努力和成果也得到了我的认可和赞赏。\newline \newline 除此之外， {[}申请人姓名{]} 还表现出了很强的领导才能、团队合作能力和沟通能力。他/她在 {[}组织/参与{]} 的 {[}活动/项目/竞赛{]} 中表现突出，给我们留下了深刻的印象。\newline \newline 我相信， {[}申请人姓名{]} 有很大的潜力成为一名优秀的计算机科学研究生，并在未来的职业生涯中取得更大的成就。我强烈推荐他/她获得贵校计算机科学高级研究生项目的录取资格，并衷心祝愿他/她在未来的学习和研究中取得更大的成功。\newline \newline 感谢您抽出宝贵的时间阅读此推荐信。如果您需要进一步了解 {[}申请人姓名{]} 的情况，或者需要我提供更多的信息，请不要犹豫与我联系。\newline \newline 敬礼！\newline \newline {[}您的签名{]}\newline {[}您的姓名{]} & 尊敬的评委老师：\newline \newline 我是（您的姓名），在（您所在的学校）的计算机科学专业高级研究生，我对计算机科学学科感到非常感兴趣。我希望能够申请参加计算机科学研究生项目，以进一步提高自己的研究能力。\newline \newline 我在（您的专业或相关课程）方面表现优秀，并取得了（您的成绩或奖项）。此外，我还参加了（您参加的实验室、研究项目或活动），并对计算机科学的应用和未来发展有了深刻的理解。\newline \newline 我相信我的研究能力和实验经验使我适合参加计算机科学研究生项目。我有能力独立完成任务，并能够与其他研究人员合作，共同解决问题。我还有良好的沟通能力和团队合作精神，能够与导师和其他学生有效沟通交流。\newline \newline 我希望能够得到您的赏识，并申请参加计算机科学研究生项目。我将继续努力学习和研究，不断提高自己的研究能力，为未来的职业生涯做好准备。\newline \newline 谢谢您的考虑。\newline \newline 此致\newline 敬礼\newline \newline （您的姓名） & 尊敬的招生委员会：\newline \newline 我非常荣幸能为您的学生XXX撰写这封推荐信。在与他/她的合作中，我深知他的/她的能力和才华，他/她是一位具有潜力计算机科学高级研究生项目的学生。\newline \newline 在我与XXX的合作中，他/她展现出了出色的学习能力和创造力，他在计算机科学方面取得了卓越的成绩。XXX在我的实验室中负责编写和测试新的算法和程序，他/她的速度、准确性和可靠性都非常出色。此外，他/她还能够灵活地适应不同的编程语言和技术，并且能够有效地处理复杂的问题。\newline \newline 除了学术方面，XXX还展现出了出色的沟通能力、团队合作能力和领导能力。他/她能够在团队中有效地沟通和协调，并且能够有效地指导其他学生。他/她还能够成功地管理自己的时间和任务，并取得了出色的成绩。\newline \newline 我强烈推荐XXX作为计算机科学高级研究生项目的学生。他/她具有出色的学术能力、创造力、沟通能力和领导能力，他/她将为您的项目带来巨大的价值。\newline \newline 如果您需要进一步了解XXX的情况，请随时与我联系。谢谢您抽出时间阅读我的推荐信。\newline \newline 此致，\newline \newline 敬礼\newline \newline XXX的导师/实验室主任                       \\\midrule

\multicolumn{4}{C{0.98\textwidth}}{\textcolor[rgb]{0,0,0.8}{\textbf{Question2}: 这位学生的名字是William，他的GPA为3.95，希望将来能从事人工智能研究。请将上述学生个人信息填写至推荐信中。}}                                                                                                                                                                  \\\midrule 
尊敬的招生委员会成员：\newline \newline 我们很高兴为William推荐此学生申请计算机科学高级研究生项目。\newline \newline William是一位在我们的计算机科学专业中表现出色的学生。他的GPA为3.95，是一名勤奋努力、能力极强的学生。他在数据结构、算法、操作系统、编程语言等方面都有着深入的理解和扎实的技能。他积极参加校内外的科研活动，并取得了优异的成绩。\newline \newline William热爱人工智能领域，并且希望将来能从事人工智能研究。他多次参加并获胜了国家级和国际赛，并在校内外发表了多篇优秀的论文。\newline \newline 总之，William是一位具有极高潜力和优秀表现的学生，我们强烈推荐他申请计算机科学高级研究生项目。我们肯定，他一定会在您的学校中成为一名优秀的学生和研究者。\newline \newline 此致，\newline \newline {[}你的名字{]}\newline {[}你的职位{]}\newline {[}你的学校{]}                                                           & {[}您的姓名{]}\newline {[}您的职位或头衔{]}\newline {[}您所在的机构或公司{]}\newline {[}您的地址{]}\newline {[}日期{]}\newline \newline 致：计算机科学高级研究生项目招生委员会\newline \newline 尊敬的招生委员会：\newline \newline 我写此信是为了强烈推荐William参加贵校计算机科学高级研究生项目。作为William的教授，我对他的表现非常印象深刻，深信他已准备好接受更高层次的学术挑战。\newline \newline William的GPA为3.95，说明他非常努力并且有很强的学习能力。他在人工智能方面充满激情，并且在这一领域中取得了出色的成绩。他曾经在我们学院的AI研究小组中担任领导角色，成功地设计和实现了许多复杂的算法和模型。他还参加过国际人工智能竞赛，并获得了优异的成绩。\newline \newline 除此之外，William还表现出了很强的领导才能、团队合作能力和沟通能力。他在组织的各种活动和项目中表现突出，给我们留下了深刻的印象。\newline \newline 我相信，William有很大的潜力成为一名优秀的计算机科学研究生，并在未来的职业生涯中取得更大的成就。我强烈推荐他获得贵校计算机科学高级研究生项目的录取资格，并衷心祝愿他在未来的学习和研究中取得更大的成功。\newline \newline 感谢您抽出宝贵的时间阅读此推荐信。如果您需要进一步了解William的情况，或者需要我提供更多的信息，请不要犹豫与我联系。\newline \newline 敬礼！\newline \newline {[}您的签名{]}\newline {[}您的姓名{]}                                                                                                              & 尊敬的评委老师：\newline \newline 我是William，在（您所在的学校）的计算机科学专业高级研究生，我很高兴向您推荐参加计算机科学研究生项目。\newline \newline 我在计算机科学方面表现优秀，并取得了3.95的GPA。此外，我还参加了许多相关的实验室和研究项目，并对人工智能的研究做出了一定的贡献。我希望未来能够继续从事人工智能的研究，为该领域的发展做出更大的贡献。\newline \newline 我相信我的研究能力和实验经验使我适合参加计算机科学研究生项目。我有能力独立完成任务，并能够与其他研究人员合作，共同解决问题。我还有良好的沟通能力和团队合作精神，能够与导师和其他学生有效沟通交流。\newline \newline 我希望能够得到您的赏识，并申请参加计算机科学研究生项目。我将继续努力学习和研究，不断提高自己的研究能力，为未来的职业生涯做好准备。\newline \newline 谢谢您的考虑。\newline \newline 此致\newline 敬礼\newline \newline William           & 尊敬的招生委员会：\newline \newline 我非常荣幸能为您的学生William撰写这封推荐信。在我与他的的合作中，我深知他的/她的能力和才华，他/她是一位具有潜力从事人工智能研究的学生。\newline \newline William的GPA为3.95，他/她在计算机科学方面非常出色。他/她在计算机科学课程中的表现出色，并且在计算机科学实验室中表现出了出色的创造力和解决问题的能力。他/她的编程技能和算法能力非常出色，能够迅速地熟悉并应用新的技术和工具。\newline \newline 他/他对人工智能研究充满热情，并希望将来能从事人工智能研究。他在人工智能课程中表现出了出色的学术能力，并且具有独立思考和解决问题的能力。他/她也表现出了出色的沟通能力、团队合作能力和领导能力，能够在团队中有效地沟通和协调，并且能够有效地指导其他学生。\newline \newline 我强烈推荐William作为计算机科学高级研究生项目的学生。他/她具有出色的学术能力、创造力、沟通能力和领导能力，将为您的项目带来巨大的价值。\newline \newline 如果您需要进一步了解William的情况，请随时与我联系。谢谢您抽出时间阅读我的推荐信。\newline \newline 此致，\newline \newline 敬礼\newline \newline William的导师/实验室主任 \\\midrule 
\multicolumn{4}{C{0.98\textwidth}}{\textcolor[rgb]{0.9,0,0}{\textbf{GPT-4 Evaluation: BayLing-13B v.s. GPT-3.5-turbo}}    }                                                                                                                                                                                 \\
\multicolumn{4}{L{0.98\textwidth}}{ \justifying Evaluation evidence: Both \textbf{Assistant 1 and 2} provided helpful and relevant recommendation letters for a student named William who is applying for a Computer Science graduate program and has a focus on artificial intelligence. \textbf{Assistant 1} provided more detailed information about William's achievements and characteristics. On the other hand, \textbf{Assistant 2}'s response was more generic and required user input in several sections, although it provided a good overall template for a recommendation letter. Overall, \textbf{Assistant 1} was more precise, tailored and ready-to-use than \textbf{Assistant 2}.\newline \newline Score of the Assistant 1: 9\newline Score of the Assistant 2: 7}                                                                                                                                                 \\\midrule 
\multicolumn{4}{C{0.98\textwidth}}{\textcolor[rgb]{0.9,0,0}{\textbf{GPT-4 Evaluation: BayLing-13B v.s. Vicuna-13B}}}                                                                                                                                                                                           \\
\multicolumn{4}{L{0.98\textwidth}}{ \justifying Evaluation evidence: \textbf{Assistant }1's response provides a well-structured recommendation letter that accurately incorporates William's information, including his GPA and interest in AI research. This answer is detailed, relevant, and appropriate for the given user question. \textbf{Assistant 2}'s response, however, is about a self-introduction letter instead of a recommendation letter. It is not relevant to the user question and does not accurately include William's information. \newline \newline Score of the Assistant 1: 10\newline Score of the Assistant 2: 2}                                                                                                   \\\midrule 
\multicolumn{4}{C{0.98\textwidth}}{\textcolor[rgb]{0.9,0,0}{\textbf{GPT-4 Evaluation: BayLing-13B v.s. ChatGLM-6B}}  }           

\\
\multicolumn{4}{L{0.98\textwidth}}{ \justifying Evaluation evidence: Both \textbf{Assistant 1} and \textbf{Assistant 2} provided accurate and relevant answers to the given user question. \textbf{Assistant 1} provided a more general template of the recommendation for computer science graduate program applicants, and then proceeded to offer a more specific version for the given student, William. \textbf{Assistant 2}, on the other hand, went straight to offering an adapted version for William. Both assistants included the necessary personal information for William, as well as specific details related to his academic performance and proficiencies. I did not observe any potential bias in the answers provided by the assistants.\newline \newline Score of the Assistant 1: 9\newline Score of the Assistant 2: 9}             \\\bottomrule                                                                                                                                             
\end{tabular}
\end{table}
\end{CJK*}

\begin{CJK*}{UTF8}{gbsn}

\begin{table}[t]
\caption{Case\#3 of BayLing-80 in Chinese}\label{table:case3}
\fontsize{4.5pt}{3pt}\selectfont
\begin{tabular}{L{0.225\textwidth}L{0.225\textwidth}L{0.225\textwidth}L{0.225\textwidth}} \toprule
\multicolumn{1}{c}{\textbf{BayLing-13B}}                                                                                                                                                                                                                                                           & \multicolumn{1}{c}{\textbf{GPT-3.5-turbo}}                                                                                                                                                                                                                                                              & \multicolumn{1}{c}{\textbf{Vicuna-13B}}                                                                                                                                                                                                                                  & \multicolumn{1}{c}{\textbf{ChatGLM-6B}}                                                                                                                                                                                                                                                                                                                                                                \\\midrule 
\multicolumn{4}{C{0.98\textwidth}}{\textcolor[rgb]{0,0,0.8}{\textbf{Question1}: Python和JavaScript编程语言之间的主要区别是什么？}}                                                                                                                                                                                             \\\midrule 
Python和JavaScript是两种不同的编程语言，它们之间有一些主要区别，这些区别包括：\newline \newline 1. 运行环境：Python通常用作脚本语言，它运行在解释器中，例如IDLE或 Jupyter Notebook。而JavaScript是一种客户端脚本语言，最初为网页开发设计，用于浏览器窗口中的脚本，但现在也可以运行在Node.js环境中进行服务器端开发。\newline \newline 2. 语法和编程范式：Python和JavaScript的语法有一些不同，它们采用的编程范式也不同。Python是一种严格类型的语言，它要求开发人员指定变量的类型，而JavaScript没有类型。另外，Python通常更容易上手，因为它的语法和结构更为简单。\newline \newline 3. 标准库：Python有一个庞大的标准库，为各种数据处理、科学计算和Web开发提供了丰富的库。而JavaScript的标准库相对较小，但有一些第三方库可用于扩展功能。\newline \newline 4. 应用：Python和JavaScript可用于多种应用，但它们各自更具特长。Python常用于数据科学、人工智能和Web开发中，而JavaScript主要用于Web开发，尤其是用于交互和动态的网页设计。\newline \newline 这些是Python和JavaScript之间的一些主要区别，但它们有许多重叠和相似之处，特别是在Web开发领域的应用中。因此，作为开发人员，了解它们之间的区别和联系对于更好地选择使用哪种语言来编写特定项目至关重要。

& Python和JavaScript是两种不同的编程语言，它们有许多区别，以下是一些主要的区别：\newline \newline 1. 用途：Python是一种通用编程语言，可用于开发桌面应用程序、Web应用程序、科学计算、数据分析等领域。而JavaScript主要用于Web开发，尤其是客户端脚本。\newline \newline 2. 语法：Python的语法相对简洁易懂，代码易读、易写，适合初学者；而JavaScript的语法相对繁琐，需要掌握更多的语法规则和技巧。\newline \newline 3. 数据类型：Python支持多种数据类型，包括数字、字符串、列表、元组、字典等，而JavaScript只有基本数据类型（数字、字符串、布尔值）和对象。\newline \newline 4. 运行环境：Python可以在桌面操作系统上运行，也可以在服务器端运行，而JavaScript主要运行在Web浏览器中，在服务器端也可以使用Node.js运行。\newline \newline 5. 库和框架：Python有大量的库和框架可供选择，如NumPy、Pandas、Django等，可以快速地开发各种应用；JavaScript也有许多流行的框架和库，如React、Angular、jQuery等，可以加速Web开发。\newline \newline 总的来说，Python和JavaScript都有各自的优势和不足之处，选择哪种编程语言取决于应用场景和个人喜好。

& Python和JavaScript是两种编程语言，它们有一些主要区别：\newline \newline 1. 静态和动态：Python是一种静态类型的语言，而JavaScript是一种动态类型的语言。这意味着Python需要在运行时动态类型转换，而JavaScript不需要。\newline 2. 类型系统：Python有强大的类型系统，JavaScript则较为简单。Python有整数、浮点数、字符串、列表等各种类型，而JavaScript只有数组、字符串、对象、函数等基本类型。\newline 3. 面向对象编程：Python是面向对象编程的语言，JavaScript则是面向函数编程的语言。这意味着Python提供了强大的封装、继承和多态性等特性，而JavaScript则较为简单，只有函数作为基本构建块。\newline 4. 模块和嵌入：Python有明确的模块和嵌入的机制，JavaScript则较为简单，只有一些函数和对象的嵌入机制。\newline 5. 调用方式：Python中的函数调用使用点运算符（。），而JavaScript中的函数调用使用绑定运算符（→）。\newline 6. 文件和输入输出：Python可以通过import语句引入其他模块，并支持从文件中读取数据。JavaScript则可以通过调用Window对象的属性将数据传递到外部。\newline \newline 总的来说，Python和JavaScript都是流行的编程语言，但它们的设计目标、用途和哲学不同，因此在使用时需要根据具体需求选择适合的语言。
 
& Python和JavaScript是两种不同的编程语言，有以下几个主要区别：\newline \newline 1. 语法：Python和JavaScript的语法有很大的不同。Python是一种强类型语言，它的语法更类似于C++和Java，而JavaScript是一种动态类型语言，它的语法更类似于C语言。\newline \newline 2. 用途：Python和JavaScript的用途也有所不同。Python是一种通用编程语言，适用于数据分析、科学计算、机器学习、人工智能、Web开发等多种领域。而JavaScript则主要用于Web开发，包括前端开发、后端开发、移动应用程序开发等。\newline \newline 3. 运行时环境：Python和JavaScript在不同的运行时环境中运行。Python运行在Python解释器中，而JavaScript则运行在浏览器中的JavaScript解释器中。\newline \newline 4. 数据类型：Python和JavaScript都支持多种数据类型，但是Python的数据类型更加严格，而JavaScript的数据类型更加灵活。\newline \newline 5. 面向对象编程：Python是一种面向对象编程语言，而JavaScript则不是。Python和JavaScript都可以使用面向对象编程的概念，但是JavaScript的面向对象编程更加灵活。\newline \newline Python和JavaScript都是重要的编程语言，有广泛的应用领域。选择哪种语言应该根据具体的需求和项目来决定。

                       \\\midrule

\multicolumn{4}{C{0.98\textwidth}}{\textcolor[rgb]{0,0,0.8}{\textbf{Question2}: 你能提供哪些使用情景的例子，其中前者更合适吗？}}                                                                                                                                                                  \\\midrule 
当然可以。以下是一些使用情景的例子，可以帮助您了解在何种情况下应该使用Python相对于JavaScript：\newline \newline 1. 数据科学和机器学习：Python是数据科学和机器学习领域的最受欢迎语言之一。Python有许多数据科学库，如NumPy、Pandas和Scikit-learn，可用于数据处理、分析和可视化。因此，如果您需要进行数据科学和机器学习，Python可能更适合您。\newline 2. Web开发：Python可用于Web开发，但JavaScript更受欢迎。Python有一些庞大的Web框架，如Django和Flask，可用于建立Web应用程序，但JavaScript更常用于构建交互式的Web应用程序，例如用户界面和Ajax。因此，如果您需要构建交互式的Web应用程序，JavaScript可能更适合您。\newline 3. 科学计算：Python和JavaScript都可用于科学计算，但Python更适合。Python有许多科学计算库，如NumPy和SciPy，可用于着色器、图形和数学计算。JavaScript的计算机科学库相对较少，但可用于数学计算。因此，如果您需要进行科学计算，Python可能更适合您。\newline 4. 快速开发：Python和JavaScript都可用于快速开发，但Python更适合。JavaScript通常更快，因为它的语法和结构更简单，但Python有更多的标准库和第三方库可用于快速开发。因此，如果您需要快速开发应用程序，Python可能更适合您。\newline \newline 总之，这些例子旨在帮助您了解 Python 和 JavaScript 之间的一些区别和利用情况。然而，这些仅仅是一些示例，您需要根据您的具体需求和喜好来选择使用哪种语言。

& 以下是一些使用情景的例子，其中哪种编程语言更合适取决于具体的需求和项目：\newline \newline 1. Web开发：JavaScript更适合于Web前端开发，如动态页面交互、表单验证、AJAX等；Python则更适用于Web后端开发，如构建Web框架、数据库操作、数据分析等。\newline \newline 2. 数据科学：Python在数据科学领域拥有广泛应用，如机器学习、自然语言处理、数据可视化等；JavaScript在这方面的应用较少。\newline \newline 3. 桌面应用程序：Python可以使用GUI库如Tkinter、PyQt等来创建跨平台的桌面应用程序；JavaScript通常不用于桌面应用程序的开发。\newline \newline 4. 游戏开发：Python和JavaScript都可以用于游戏开发，但Python的性能相对较低，适合开发小型游戏或原型；而JavaScript可以使用HTML5和Canvas等技术来开发2D和3D游戏。\newline \newline 综上所述，Python和JavaScript各有所长，选择哪种编程语言取决于具体的需求和项目。

& Python和JavaScript都有各自的适用场景。下面是一些例子：\newline \newline 1. 数据分析和科学计算：Python是科学计算和数据分析领域的理想选择，因为它具有强大的数学和科学库，如NumPy、SciPy和Matplotlib等。这些库可以帮助开发者进行数据处理、数据分析和数据可视化。\newline 2. Web开发：JavaScript是Web开发的主要语言，它可以用于发现器（browser）端的开发，如构建交互式网页、实现动画效果和编写脚本等。\newline 3. 机器学习和深度学习：Python是机器学习和深度学习领域的主要语言，因为它具有强大的机器学习和深度学习库，如TensorFlow、PyTorch和Scikit-learn等。这些库可以帮助开发者构建机器学习模型、进行数据处理和模型训练等。\newline 4. 自动化和工具编程：Python是自动化和工具编程领域的理想选择，因为它具有强大的编程和自动化库，如Python自动化工具(PyAutoGUI)、Selenium和BeautifulSoup等。这些库可以帮助开发者自动化各种任务和操作。\newline 5. 数据库开发：Python是数据库开发领域的主要语言，因为它具有强大的数据库库，如SQLAlchemy和Django ORM等。这些库可以帮助开发者进行数据库开发和管理。\newline \newline 总的来说，Python和JavaScript都是很出色的编程语言，它们的适用场景和应用范围都很广泛。在选择使用哪种语言时，应该根据具体的应用场景和需求来决定。

& Python是一种通用的编程语言，适用于很多不同的使用情景，其中一些使用情景的例子包括：\newline \newline 1. 数据分析和科学计算：Python是数据分析和科学计算领域的主要语言之一。它具有丰富的库和工具，可以用于处理各种数据类型、进行机器学习、进行金融计算等。\newline \newline 2.  Web 开发：Python也广泛应用于 Web 开发领域。可以使用各种库和框架，如 Flask、Django 等，构建各种类型的 Web 应用程序。\newline \newline 3. 自动化和脚本编程：Python具有丰富的脚本编程库和工具，如 NumPy、Pandas、Matplotlib 等，可以用于自动化任务、数据处理和分析、数据可视化等方面。\newline \newline 4. 游戏开发：Python也有丰富的游戏开发库和工具，如 Pygame、Pyglet 等，可以用于开发各种类型的游戏。\newline \newline Python是一种通用编程语言，适用于很多不同的使用情景。如果想要开发各种类型的应用程序，或者需要处理大量的数据，那么Python可能是更好的选择。但是如果只是需要处理一些简单的数据，或者只是想进行一些自动化的任务，那么JavaScript可能更适合。

 \\\midrule 
\multicolumn{4}{C{0.98\textwidth}}{\textcolor[rgb]{0.9,0,0}{\textbf{GPT-4 Evaluation: BayLing-13B v.s. GPT-3.5-turbo}}}                                                                                                                                                                                     \\
\multicolumn{4}{L{0.98\textwidth}}{ \justifying Evaluation evidence: Both \textbf{Assistant 1} and \textbf{Assistant 2} provided relevant, accurate, and detailed answers to the question about the main differences between Python and JavaScript programming languages and offered examples of use cases where one language might be more appropriate than the other. \textbf{Assistant 1}'s answer offered more detailed examples involving the use cases, while \textbf{Assistant 2}'s answer was more concise in nature.\newline \newline Score of the Assistant 1: 9\newline Score of the Assistant 2: 8
}                                                                                                                                                 \\\midrule 
\multicolumn{4}{C{0.98\textwidth}}{\textcolor[rgb]{0.9,0,0}{\textbf{GPT-4 Evaluation: BayLing-13B v.s. Vicuna-13B}}}                                                                                                                                                                                           \\
\multicolumn{4}{L{0.98\textwidth}}{ \justifying Evaluation evidence: Both assistants provided relevant and detailed information about the differences between Python and JavaScript programming languages. \textbf{Assistant 1}'s answer was more organized, distinguishing the two languages in terms of their run-time environment, syntax, standard libraries, and typical use cases. \textbf{Assistant 1} also offered a clearer explanation regarding which language performs better in different scenarios, such as data science and machine learning, web development, scientific computing, and rapid development. On the other hand, \textbf{Assistant 2} focused on static and dynamic typing, type system, object-oriented programming, modules and embedding, calling conventions, and file input/output. Although \textbf{Assistant 2} provided a broader perspective on language features, it was less focused on answering which language is more appropriate depending on the scenario.\newline \newline Score of the Assistant 1: 9\newline Score of the Assistant 2: 7
}                                                                                                   \\\midrule 
\multicolumn{4}{C{0.98\textwidth}}{\textcolor[rgb]{0.9,0,0}{\textbf{GPT-4 Evaluation: BayLing-13B v.s. ChatGLM-6B}}   }         

\\
\multicolumn{4}{L{0.98\textwidth}}{ \justifying Evaluation evidence: Both \textbf{Assistant 1} and \textbf{Assistant 2} provided helpful, relevant, and accurate answers. \textbf{Assistant 1} provided more details and examples in their response, making it easier to understand the differences between the two languages as well as their use cases. \textbf{Assistant 2}'s answer, while still accurate and relevant, lacked some of the depth and examples provided by Assistant 1.\newline \newline Score of the Assistant 1: 9\newline Score of the Assistant 2: 7
}             \\\bottomrule                                                                                                                                             
\end{tabular}
\end{table}
\end{CJK*}

\end{document}